\documentclass[journal]{vgtc}

\ifpdf
  \pdfoutput=1\relax                   
  \usepackage{graphicx}                
  \DeclareGraphicsExtensions{.pdf,.png,.jpg,.jpeg} 
\else
  \usepackage{graphicx}                
  \DeclareGraphicsExtensions{.eps}     
\fi%

\graphicspath{{figures/}{pictures/}{images/}{./}} 

\usepackage{microtype}                 
\PassOptionsToPackage{warn}{textcomp}  
\usepackage{textcomp}                  
\usepackage{mathptmx}                  
\usepackage{times}                     
\usepackage{float}
\usepackage{cite}                      
\usepackage{tabu}                      
\usepackage{booktabs}                  

\usepackage{amsmath}
\usepackage{amsfonts}
\usepackage{multirow}
\usepackage{xspace}
\usepackage{enumitem}
\usepackage{array}
\usepackage{tabularx}

\DeclareMathOperator*{\argmin}{arg\,min}
\newcommand{\sys}{HyperNP\xspace}

\usepackage[dvipsnames]{xcolor}
\usepackage[normalem]{ulem}
\definecolor{mred}{rgb}{.80,.12,.30}
\definecolor{MRED}{rgb}{.80,.12,.30}
\definecolor{grey}{rgb}{0.5,0.5,0.5}
\definecolor{purple}{rgb}{.75,0,.85}
\definecolor{reb}{rgb}{.047,.317,.314}
\definecolor{monstergreen}{rgb}{.58, .84, 0}
\definecolor{pistachio}{rgb}{0.58, 0.77, 0.45}
\definecolor{eagreen}{rgb}{0.1, 0.5, 0.1}
\usepackage{orcidlink}

\newif\ifnotes
\notestrue

\let\origcite\cite

\renewcommand{\cite}[1]{\ifnotes\mbox{\origcite{#1}}\else \origcite{#1}\fi}

\onlineid{1197}

\vgtccategory{Research}

\vgtcinsertpkg
\usepackage{subcaption}
\usepackage{dblfloatfix}

\title{\sys: Interactive Visual Exploration of\\ Multidimensional Projection Hyperparameters}

\author{
    Gabriel Appleby\,\orcidlink{0000-0003-2436-2121}
    \thanks{Department of Computer Science. Tufts University.}\\
    \and Mateus Espadoto\,\orcidlink{0000-0002-1922-4309}
    \thanks{Institute of Mathematics and Statistics. University of Sao Paulo, Brazil.}
    \thanks{Department of Information and Computing Sciences. University of Utrecht.}\\ 
    \and Rui Chen\,\orcidlink{0000-0001-5938-2195}\footnotemark[1]\\ 
    \and Samuel Goree\,\orcidlink{0000-0002-9650-9172}\thanks{Luddy School of Informatics, Computing, and Engineering. Indiana University Bloomington.}\\ 
    \and Alexandru C Telea\,\orcidlink{0000-0003-0750-0502}\footnotemark[3]\\ 
    \and Erik W Anderson\,\orcidlink{0000-0002-0334-8497}\thanks{Novartis AI Innovation Center. Novartis.}\\ 
    \and Remco Chang\,\orcidlink{0000-0002-6484-6430}\footnotemark[1]\\
}
\date{
Institute of Mathematics and Statistics, University of São Paulo
}

\teaser{
    \centering
    \begin{subfigure}[b]{0.3\linewidth}
        \centering
        \includegraphics[width=\linewidth]{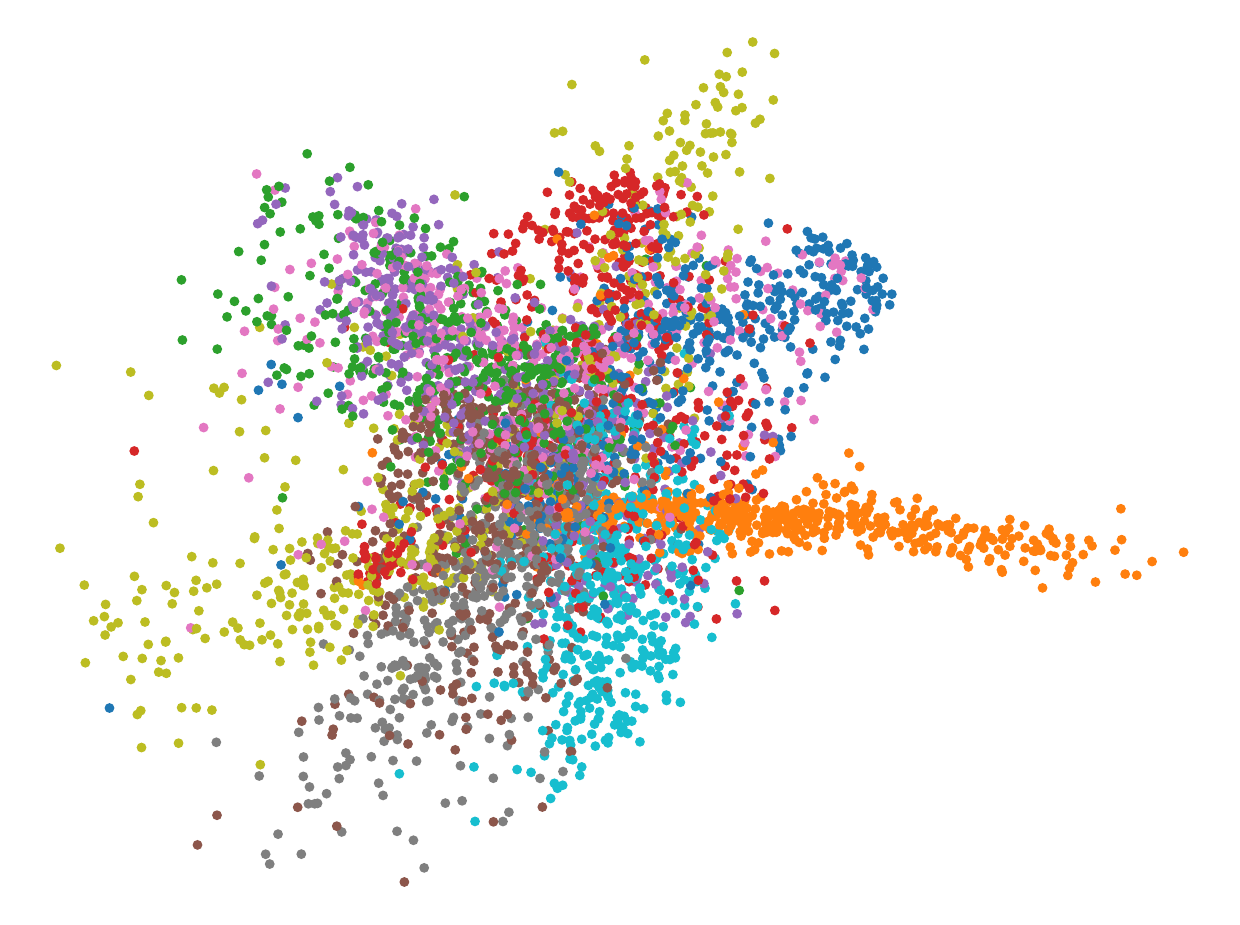}
        \label{fig:teaser:k3}
        \caption{$k=3$}
    \end{subfigure}
    \begin{subfigure}[b]{0.3\linewidth}
        \centering
        \includegraphics[width=\linewidth]{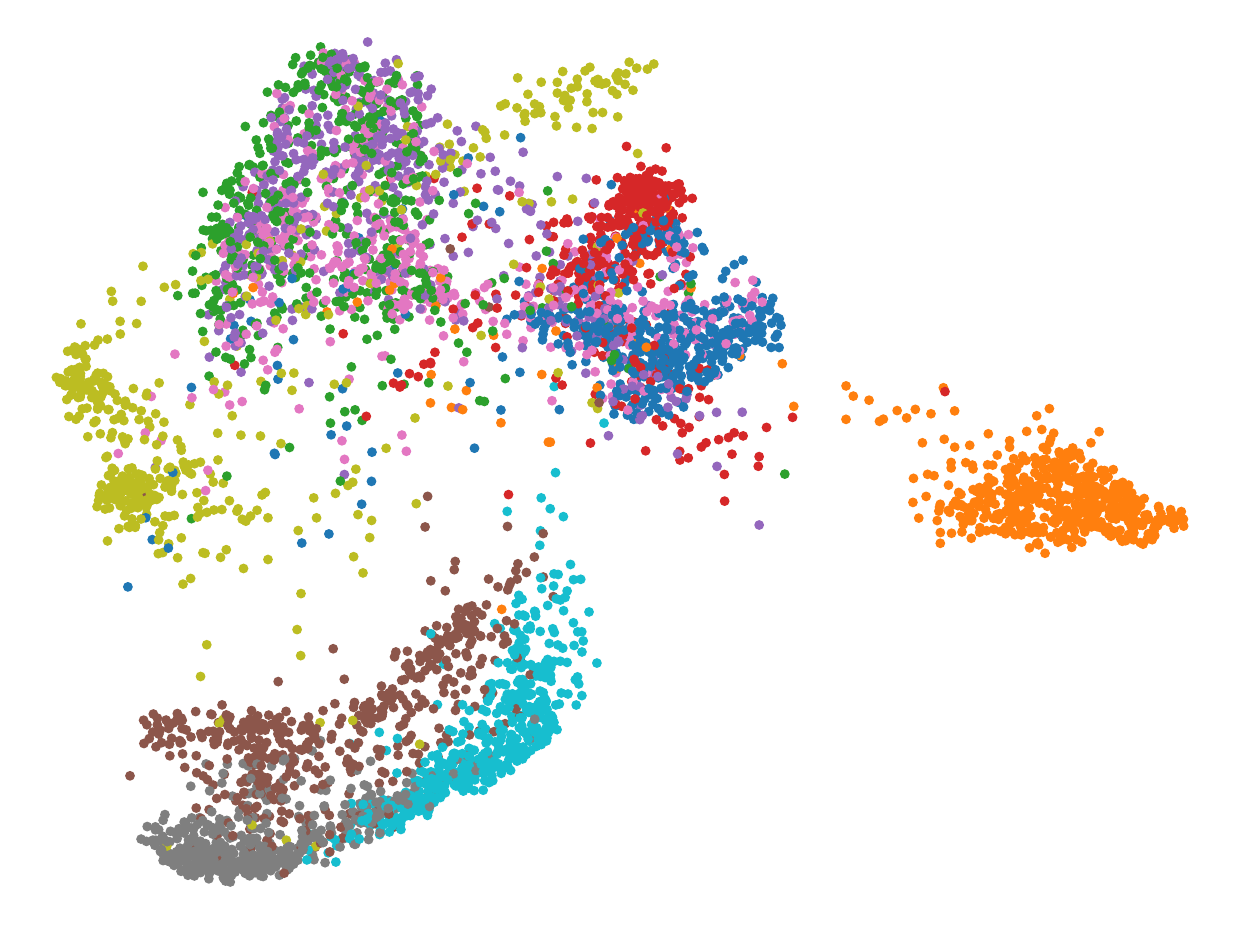}
        \label{fig:teaser:k7}
        \caption{$k=7$}
    \end{subfigure}
    \begin{subfigure}[b]{0.3\linewidth}
        \centering
        \includegraphics[width=\linewidth]{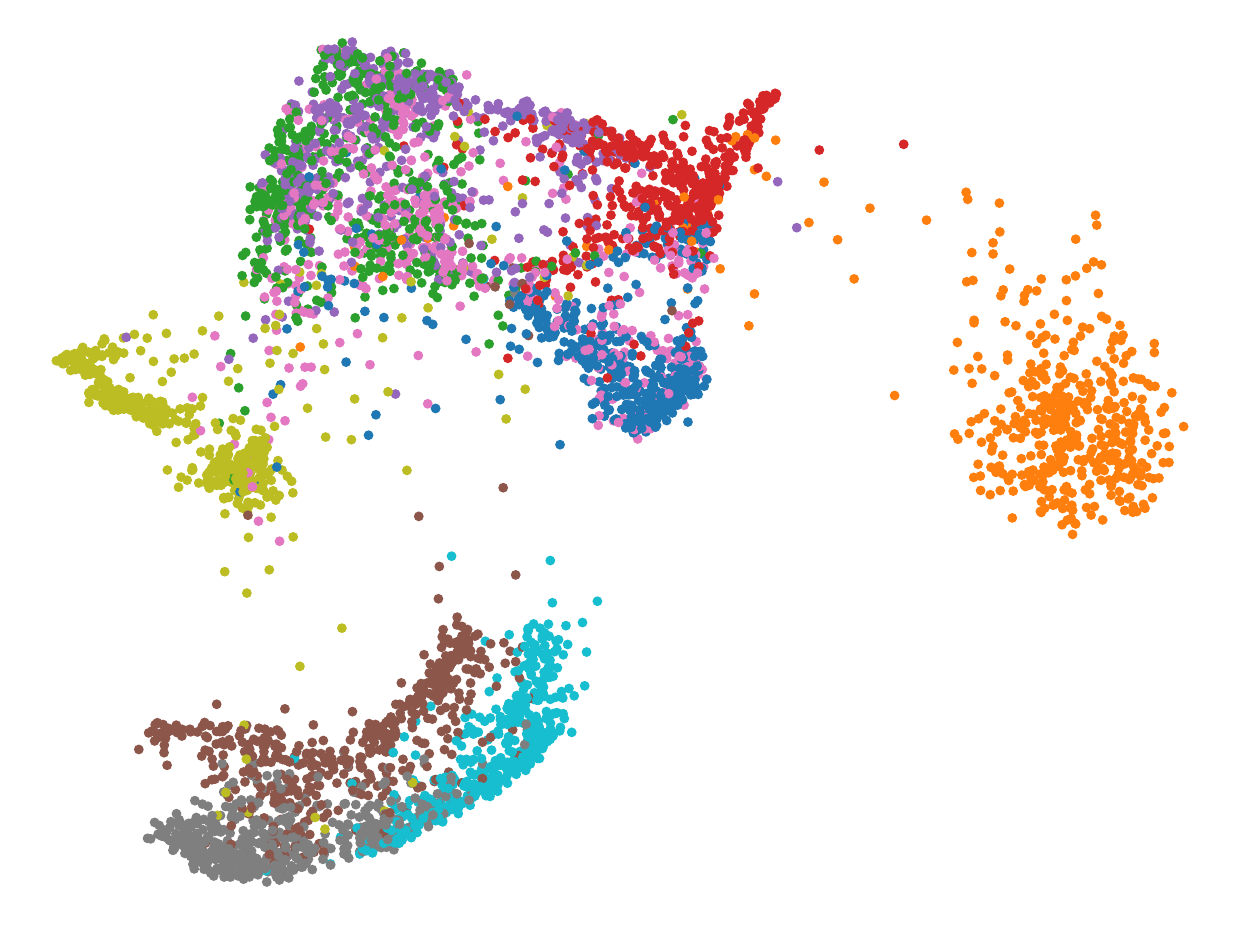}
        \label{fig:teaser:k11}
        \caption{$k=11$}
    \end{subfigure}
    \caption{\sys projecting unseen instances for unseen hyperparameter values. This model was trained on UMAP projections of the FashionMNIST dataset with a gap size of four and a training fraction of $0.2$ (see Sec.~\ref{sec:method} for details). In this case, the hyperparameter used is the number of nearest neighbors $k$. As $k$ increases, UMAP preserves more of the global structures in the data.}
    \label{fig:teaser}
}

\abstract{
Projection algorithms such as t-SNE or UMAP are useful for the visualization of high dimensional data, but depend on hyperparameters which must be tuned carefully.
Unfortunately, iteratively recomputing projections to find the optimal hyperparameter value is computationally intensive and unintuitive due to the stochastic nature of these methods.
In this paper we propose \sys, a scalable method that allows for real-time interactive hyperparameter exploration of projection methods by training neural network approximations.
\sys can be trained on a fraction of the total data instances and hyperparameter configurations and can compute projections for new data and hyperparameters at interactive speeds.
\sys is compact in size and fast to compute, thus allowing it to be embedded in lightweight visualization systems such as web browsers.
We evaluate the performance of the \sys across three datasets in terms of performance and speed.
The results suggest that \sys is accurate, scalable, interactive, and appropriate for use in real-world settings.
}

\CCScatlist{
  \CCScatTwelve{Human-centered computing}{Visu\-al\-iza\-tion}{Visu\-al\-iza\-tion techniques}{Treemaps};
  \CCScatTwelve{Human-centered computing}{Visu\-al\-iza\-tion}{Visualization design and evaluation methods}{}
}

\begin{document}


\firstsection{Introduction}
\maketitle

\label{sec:introduction}
As data-generating devices and computational resources expand, there is a growing need for interactive visual exploration for high-dimensional data\,\cite{liu:2017:viz_high_d_survey}. 
An important class of visualization methods for such data is projection, which maps data from a high-dimensional space to a similarity-preserving low-dimensional representation\,\cite{nonato:2018:projection_for_visual_analytics,espadoto:2019:quantiative_dr_survey}.
Many projection methods exist, including linear and global techniques such as PCA\,\cite{pearson:1901:pca}, and non-linear approaches such as t-SNE\,\cite{maaten:2008:tsne}, UMAP\,\cite{mcinnes:2018:umap}, and Isomap\,\cite{balasubramanian:2002:isomap}.
While widely used to visualize high-dimensional data, projections can be sensitive to hyperparameter choices which the practitioner must tune carefully.
Unfortunately, many projection methods are computationally expensive making real-time hyperparameter changes infeasible.
In order to support a user's exploration of hyperparameters in projections a method is required that enables real-time interaction with a projection algorithm's hyperparameter values.
In this paper, we present \sys, a deep learning technique that approximates projections across hyperparameters.
The method can estimate projections for new data, like existing approximation techniques\,\cite{espadoto:2020:dlmp}.
In addition, \sys computes projections for hyperparameter values \emph{unseen} during training, without the heavy recomputation inherent to the process for most projection methods.
Since this inferencing operation is implemented as a neural network, the compute time required for this step is minimal.
We leverage this fast computation to allow real-time visualization of large data sets and interactive exploration of different hyperparameter values associated with the projection operator.
\sys is computationally lightweight, can be used in a web-based environment, and still enables real-time interactive steering of projections. 
This empowers users to more easily reason about how hyperparameters, common to many projection methods, impact the final visualization.
For example, \sys enables the exploration of the effect of the perplexity parameter in t-SNE as done by Wattenberg \emph{et al.}\,\cite{wattenberg:2016:using-tsne-effectively}
with the key added-value of being able to do this in real time and for large datasets. 
This enables users to follow Wattenberg's proposal -- explore how perplexity controls emphasis on local \emph{vs} global relationships to get the most benefit from the projection, in real time.
Our proposal is applicable to any projection requiring user-tuned input.
We demonstrate \sys's ability to learn hyperparameters by applying it to UMAP, Isomap, and t-SNE. 
This shows that \sys can learn hyperparameter spaces that range over both continuous domains (t-SNE perplexity) as well as discrete integral domains (the $k$-nearest neighbors value for UMAP and Isomap).
Additionally, we demonstrate \sys's ability to project unseen data appropriately without the need for potentially expensive recomputation.
As with all deep learning techniques, \sys has two stages: training and inference.
Training \sys to mimic the outcome of a parameterized projection method has two steps: (1) The hyperparameters are sampled and used to project a subset of the data using the user-selected projection technique $P$. (2) The \sys model is trained with a loss function that captures the differences between its prediction and the ground-truth given by $P$. 
After training, \sys infers the 2D projection of high-dimensional data for any user-selected projection hyperparameter values.
At this stage, the user can interact with the hyperparameters using \sys, as the effect of these parameters can be estimated by the \sys model.
\sys projects data at speeds that allow users to explore different hyperparameter configurations in real time. We validate the accuracy, scalability, and performance of \sys in three sets of experiments.
First, we show the accuracy and flexibility of \sys when approximating t-SNE, UMAP, and Isomap. Our results are visually indistinguishable from those produced by these projection methods, and are within 3\% of the ground-truth in terms of common projection-quality metrics.
Secondly, we show that \sys supports scalable exploration as it can create high-quality projections using a fraction of the training data and hyperparameter samples. We call this  \emph{data usage efficiency} and illustrate the trade-offs between performance and data usage.
Finally, the inference speed of \sys enables interactive projection exploration far exceeding what is possible using existing implementations of t-SNE and UMAP.
We further discuss additional capabilities of \sys beyond its originally intended use of approximating projection methods.
Summarizing, we claim the following contributions:
\begin{itemize}[topsep=0pt, partopsep=0pt,itemsep=0pt,parsep=2pt]
    \item \sys, a \textbf{method} able to approximate projections of high-dimensional data across hyperparameter configurations at interactive speeds;
    \item a quantitative \textbf{evaluation} of our method that shows its data usage efficiency;
    \item \textbf{examples} of how our method can be applied within visual analytics systems to enable user-driven exploration of projection hyperparameters in a scalable framework.
\end{itemize}

\section{Related Work}
\label{sec:related_work}
We start with some notation. Let $D = \{\mathbf{x}_i\}$, $\mathbf{x}_i \in \mathbb{R}^n$, $1 \leq i \leq N$ be a $n$-dimensional dataset. 
Its $N$ points $\mathbf{x}_i$  (also called samples or observations) have each $n$ dimensions (also called variables or attributes).
\subsection{Overview of Projection Techniques}
\label{sec:related_work:projection_techniques}
Many techniques exist for visualizing multidimensional data\,\cite{hoffman:2002:survey_of_vis_for_high_d, oliveira:2003:viz_high_d_survey, 
kehrer13, liu:2017:viz_high_d_survey}.
Early, but still popular, solutions include parallel coordinate plots\,\cite{inselberg:1985:parallel_coords},
Andrews curves\,\cite{andrews:1972:andrews_curve}, glyphs\,\cite{chernoff:1973:chernoff_faces,pickett:1988:stick_figure_plots}, and small multiple designs (scatterplot matrices\,\cite{hartigan:1975:splom}, permutation matrices\,\cite{bertin:1983:semiology}, dimension stacking\cite{leblanc:1990:dim_stacking}), and pixel-based techniques\,\cite{keim:1994:visdb_pixel_based_md_vis}.
While some of these techniques scale well with the number of samples $N$, they typically cannot handle more than a few tens of dimensions.
Projection, or dimensionality reduction (DR) techniques, is a separate class of high-dimensional visualization methods.
A DR method can be seen as a function $P : \mathbb{R}^n \rightarrow \mathbb{R}^q$, where $q \ll n$ and, in practice $q=2$.
The projection $P(\mathbf{x}_i)$ of a sample $\mathbf{x}_i \in D$ is thus a 2D point.
Hence, the projection of $D$, denoted $P(D)$, is a 2D scatterplot. 
A projection $P$ aims to place points $\mathbf{x}_i \in D$ that are regarded to be similar from the perspective of a measure defined on the data space $D$ close to each other in the scatterplot $P(D)$.
Hence, users can reason about data patterns in $D$ by examining this scatterplot.
Projections scale well both in the number of samples $N$ and dimensions $n$, and as such have been widely used for visualizing high-dimensional data.
Several works examine projections and their added-value from several viewpoints: 
\vspace{4pt}
\noindent\textbf{Taxonomies:} Projection techniques can be grouped into several taxonomies, thereby providing ways for practitioners to compare and choose a technique based on specific requirements.
Van der Maaten \emph{et al.}\,\cite{maaten:2009:dim_reduction_survey} and Cunningham \emph{et al.}\,\cite{cunningham:2015:linear_dim_reduction_survey} show how $P$ can be computed by several types of non-linear, or respectively linear, optimization methods that offer different tradeoffs between computational scalability and projection quality.
Sorzano \emph{et al.}\,\cite{sorzano:2014:dim_reduction_survey} and Engel \emph{et al.}\,\cite{engel12} propose two taxonomies of projections based on their implementation aspects, in particular the cost functions they optimize to compute $P$, and the choices of available optimizers and their computational complexities. 
\vspace{4pt}
\noindent\textbf{Quality and perception:} A projection's effectiveness is a combination of how well $P(D)$ captures data patterns present in $D$ and how (easily) users actually perceive the patterns in $P(D)$.
Heulot \emph{et al.}\,\cite{heulot17} and Nonato and Aupetit\,\cite{nonato:2018:projection_for_visual_analytics} classify the different types of errors that projection techniques create and how these impact different classes of visual analytics tasks.
Separately, several authors propose perception models\,\cite{albuquerque11,tatu10,wang18} and visual quality metrics and visualizations thereof \,\cite{lee09,mokbel13,martins14,sedlmair15,sedlmair16,cutura18} to help users understand the quality of the projections they compute, as determined by the chosen algorithms.
\vspace{4pt}
\noindent\textbf{Quantitative analysis:} Several works propose benchmarks which measure the quality of projection techniques, gauged by means of several quality metrics, across a selection of datasets\,\cite{maaten:2009:dim_reduction_survey,espadoto:2019:quantiative_dr_survey}.
Such results provide guidelines that help practitioners to choose a suitable technique depending on the type of quality metrics they want to maximize for a given dataset type.
\subsection{Hyperparameters in Projection Techniques}
\label{sec:related_work:hyperparameter}
Formally put, the function $P$ does not depend only on $D$, but also on a set of \emph{hyperparameters} $\mathbf{h} = \{h_i\}$. These control several aspects of $P$ as follows.
Computing a \emph{global} mapping $P$ between $\mathbb{R}^n$ and $\mathbb{R}^2$ that has overall high quality is, in general, not possible.
As such, many projection techniques construct different mappings for different small-scale neighborhoods in $\mathbb{R}^n$.
For example, in a number of projection techniques (including UMAP, Isomap and others), a scalar parameter $h$ controls the neighborhood size.
Many projection techniques follow the above approach. Locally Linear Embedding (LLE)\,\cite{roweis:2000:lle} fits a hyperplane through each point and its nearest neighbors, keeping local relationships linear, but allowing the global structure to be nonlinear.
Isomap\,\cite{balasubramanian:2002:isomap} tackles the problem of projecting curved manifolds by estimating geodesic distances over neighborhoods and using these as a cost function to derive the projection.
Least Squares Projection\cite{paulovich:2008:lsp} projects a subset of landmarks and then uses a fast Laplacian-like operator to map the remaining samples.
Two Phase Projection\,\cite{paulovich:2010:two_phase_projections}, LAMP\cite{joia:2011:lamp}, and the Piece-wise Laplacian Projection\,\cite{paulovich:2011:piece_wise_lapacian_projection} use a similar approach. 
More recently, t-SNE\,\cite{maaten:2008:tsne} preserves neighbors during the projection by minimizing the Kullback-Leibler divergence between neighborhood probabilities in $\mathbb{R}^n$ and $\mathbb{R}^2$.
UMAP\,\cite{mcinnes:2018:umap} models the $\mathbb{R}^n$ manifold over small neighborhoods with a fuzzy topological structure and searches for a 2D representation that has the closest possible fuzzy topological structure.
All above methods, thus, depend on hyperparameters $\mathbf{h}$ that model neighborhood sizes or number (and selection) of landmark points.

All of the discussed work mention that finding good hyperparameter values is challenging.
Espadoto \emph{et al.}\,\cite{espadoto:2019:quantiative_dr_survey} recognize this problem and address it by proposing optimal and preset parameter values computed by grid search over the hyperparameter space.
However, this process is expensive. 
More importantly, certain projection methods do not have globally optimal presets.
Wattenberg \emph{et al.}\,\cite{wattenberg:2016:using-tsne-effectively} illustrate this for t-SNE's perplexity parameter whose variation can create projections which emphasize different aspects of the $n$-dimensional data.
However, experimenting with many hyperparameter values is a costly process, especially for projections which take long to compute.
Deep learning methods are particularly effective for computationally scalable DR as demonstrated originally by autoencoders\,\cite{hinton:2006:nn_dim_reduction,stuhlsatz:2012:generalized_discriminant_analysis}.
More recently, NNP\,\cite{espadoto:2020:dlmp, espadoto:2020:improving_dlmp} 
use deep learning to mimic any projection technique $P$ by training on $P(D')$ for a small subset $D' \subset D$.
In addition to providing fast computation, such approaches also have the ability to project data not used in the original mapping, or \emph{out-of-sample} data.
Furthermore, due to the neural-network based approach, inference computations are parameter free.
However, this is not always desirable. 
As observed by Wattenberg et al.\cite{wattenberg:2016:using-tsne-effectively}, users may want to control $\mathbf{h}$ to gain insight into their data or to illustrate local versus global patterns (e.g. by changing the neighborhood size).  
\sys addresses the challenges of hyperparameter exploration and out-of-sample projection in an interactive, naturally scalable framework.
\subsection{Projection-Based Visual Analytics Systems}
\label{sec:related_work:visual_analytics_systems}
Given their advantages mentioned in Sec.~\ref{sec:related_work:projection_techniques}, many visual analytics systems are built around projection techniques.
\emph{Interactivity} is key to all these systems\,\cite{sacha:2016:dim_reduction_interaction_survey} and involves two aspects, as users typically (1) aim to analyze subsets of $D$ and/or subsets of its $n$ dimensions to obtain local, more specific, insights; and (2) need to compute $P$ several times for different values of its hyperparameters $\mathbf{h}$ to obtain the desired quality.
As such, the ability to calculate $P$ at interactive rates and for large datasets is a much needed capability for a projection technique in a visual analytics setting.
We next discuss this by highlighting several visual analytics systems.  The works mentioned below are not exhaustive, but represent a cross-section of similar methods.
A number of techniques support a user's exploration of high-dimensional data. Explorer\,\cite{paulovich:2007:projection_explorer} is a general tool for exploring high-dimensional data through projections which can deal with various data types, such as structured tables and unstructured text. 
iPCA\,\cite{jeong:2009:ipca} enables users to better understand and utilize PCA in an interactive setting.
Multiple-coordinated views show the relationship between feature scaling and the resulting projection.
Projection Inspector\,\cite{pagliosa:2015:projection_inspector} is an interactive assessment tool to compare and contrast different projections, and also explores the space between different projections using interpolation.
Probing Projections\,\cite{stahnke:2015:probing_projections} provides interactive methods to probe the high-dimensional data as well as the projection to better examine multi-dimensional datasets.
Subspace Analysis\,\cite{liu:2015:dynamic_projection} is based on the assumption that a high-dimensional dataset is a mixture of low-dimensional linear subspaces with varied dimensions. It computes a dynamic projection to discover relationships between different projections and hidden patterns.
t-viSNE\,\cite{kerren:2020:tvisne} is an interactive tool for exploring different aspects of the t-SNE method, such as effects of hyperparameters, projection patterns, and accuracy.
Informative methods, such as point densities and highlighted areas, are proposed.
Alternatively, researchers have also proposed interactive techniques that allow a user to modify or make use of the projection for further tasks.
Dis-Function\,\cite{brown:2012:dis} allows users to interactively learn a distance function by moving points within an initial projection.
IVisClassifier\,\cite{choo:2010:ivisclassifier} uses LDA\,\cite{rao:1948:lda} with detail views to assist users in classification tasks. A follow-up system is described in\,\cite{sherkat:2018:interactive_doc_clustering_revisited}.
InterAxis\,\cite{kim:2016:interaxis} and AxisSketcher\,\cite{kwon:2017:axis_sketcher} allow users to interactively define or modify the axes of a projection by dragging samples to either side of the $x$ or $y$ axes.
Steerable t-SNE\,\cite{pezzotti:20017:steerable_tsne} is a user-steerable approximate t-SNE that updates distance calculations iteratively and interactively.
Praxis\,\cite{cavallo:2018:praxis} provides an interactive framework using both forward an backward projection to help users better understand projections, along with new visualization techniques.
DimReader\,\cite{faust:2018:dim_reader} studies the effect of perturbing dimensions of interest in a projection, also providing ways to discover good perturbations.
SIRIUS\,\cite{dowling:2019:sirius} treats samples and dimensions ``equally'' by creating a dual, symmetric, exploratory system for both.
This advocates a need to investigate the projection of both samples and dimensions and their inner relationship.

We find two key commonalities of all systems discussed above:  (1) They have to work at interactive rates, otherwise none of the exploration scenarios they propose would adequately function, even for moderately large datasets.
(2) Parameter values are not a main \emph{concern} in the exploration process. 
Most systems aim to support the user with the exploration of the data samples, dimensions, correlations, selections, etc.
Ideally, the user would not want to spend time on computing projections for different parameter values -- this should happen on-demand. 
These are the gaps that \sys is designed to address.
\section{Method}
\label{sec:method}
We present \sys, a technique that can approximate any projection of a high-dimensional dataset at interactive speeds.
This allows users to easily experiment with the hyperparameters of the approximated projection method.
Conceptually, similar to previous work on neural network projection (NNP)\,\cite{espadoto:2020:dlmp,espadoto:2020:improving_dlmp}, we use a neural network to learn the approximation of a projection.
As NNP showed, such networks can be highly accurate in their approximations and can infer a projection in a fraction of the time required by the original technique.
In addition to computing the projection $\mathbf{y}\in \mathbb{R}^2$ of a sample $\mathbf{x}\in \mathbb{R}^n$, and different from NNP, our trained network takes an additional argument $\mathbf{h}$ that represents the hyperparameters of the projection algorithm. For example, for UMAP, $\mathbf{h}$ is a single value that represents the number of nearest neighbors used in UMAP; for t-SNE, $\mathbf{h}$ is a single value that models perplexity.
Performing inference on a data point $\mathbf{x}$ can be written as
\begin{align}
\label{eq:inference}
\mathbf{y} = \hat{P}([\mathbf{x}; \mathbf{h}])
\end{align}
where $\hat{P}$ represents \sys's neural network, $\mathbf{y}$ is its projection of $\mathbf{x}$, and $[\mathbf{x}; \mathbf{h}]$ is the concatenated vector of the sample and hyperparameter values that serves as the input to $\hat{P}$.
\begin{figure}[t]
    \centering
    \includegraphics[width=\linewidth]{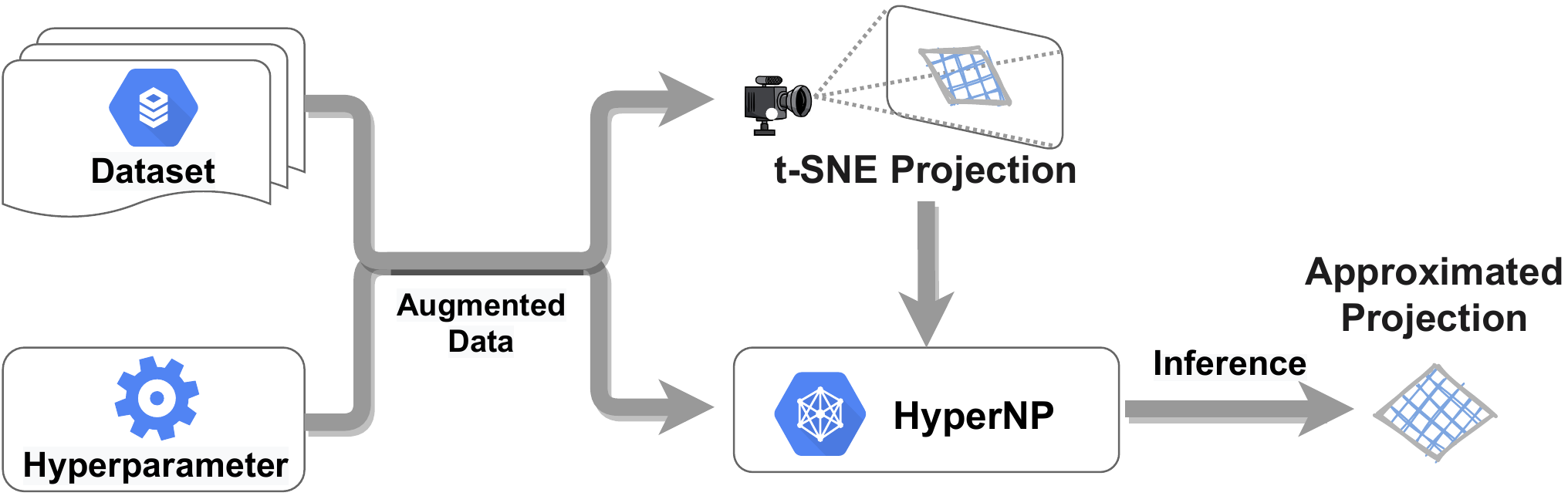}
    \caption{Architecture of \sys for approximating t-SNE: Training data and hyperparameters are used to generate a target t-SNE projection. \sys's neural network is trained and tuned using the combination of the training data, hyperparameters, and target projection. After training, \sys inferences to produce projections that approximate t-SNE's original ones, but at a fraction of the t-SNE computation cost. Note that t-SNE can be replaced with other projection methods such as UMAP or Isomap.}
    \label{fig:pipeline}
\end{figure}
\subsection{Model Definition}
\label{sec:method:model_definition}
As a neural network, $\hat{P}$ has $\lambda$ layers $[V_1, V_2 \ldots, V_\lambda]$, with $V_1$ being the input layer. A layer $V_\ell$ consists of a list of nodes $\mathbf{v}_\ell$, a weight matrix $W_\ell$, and a bias scalar $b_\ell$. 
We define nodes in each layer $\ell > 1$ as $\mathbf{v}_\ell = f_\ell(\mathbf{v}_{\ell-1}W_{\ell} + b_\ell)$ where $f_\ell$ is the activation function of layer $\ell$.
Overall, $\hat{P}$ has the parameters $\theta = [W_1, \ldots, W_\lambda, b_1, \ldots, b_\lambda]$ that are learned during training, as follows.
\subsection{Model Training}
\label{sec:method:model_training}
\begin{figure*}[ht!]
    \centering
    \begin{subfigure}[b]{0.19\linewidth}
        \centering
        \includegraphics[width=\linewidth]{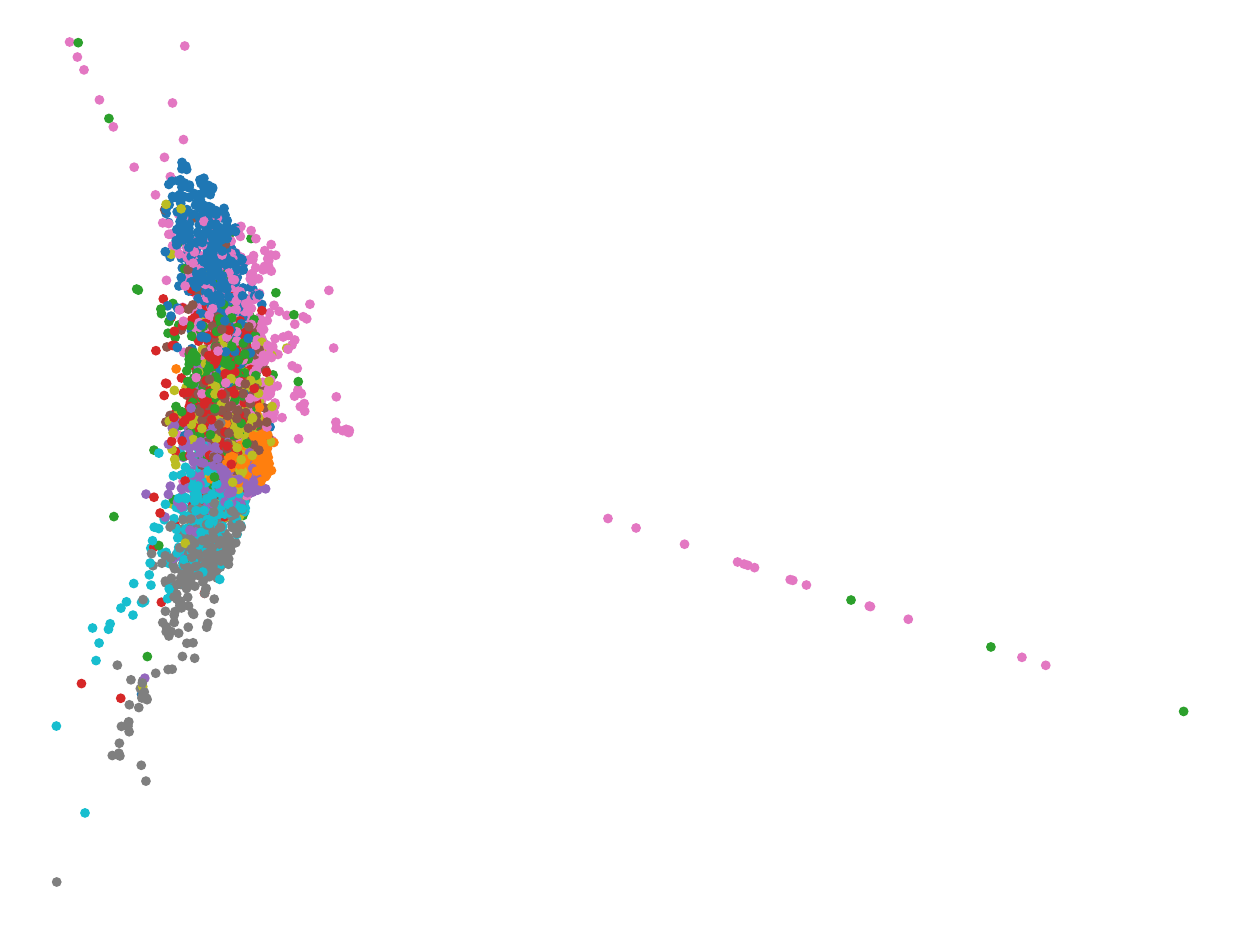}
        \caption{$k=3$}
        \label{fig:isomap_flipping:k3}
    \end{subfigure}
    \begin{subfigure}[b]{0.19\linewidth}
        \centering
        \includegraphics[width=\linewidth]{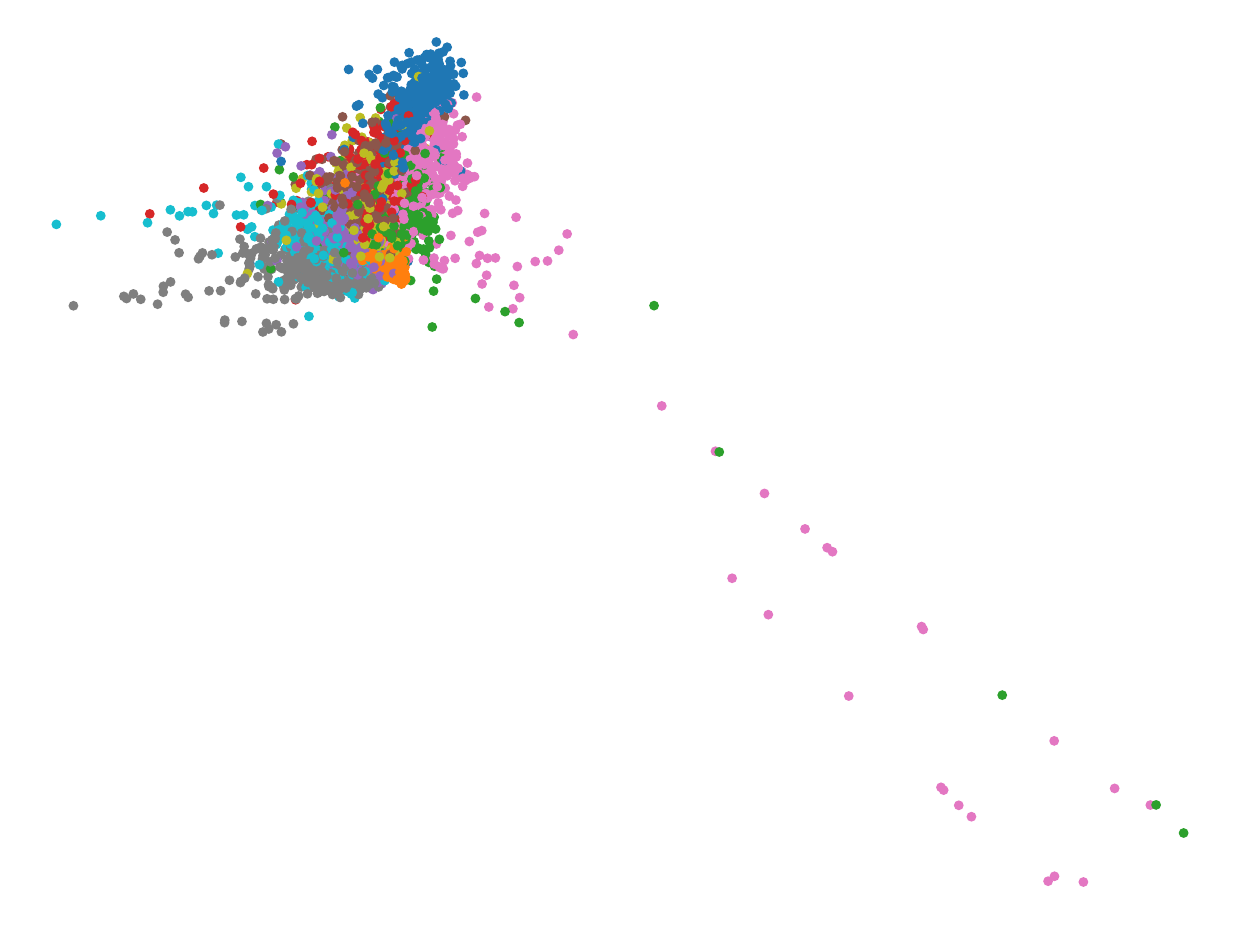}
        \caption{$k=4$, Possibility \#1}
        \label{fig:isomap_flipping:k4_1}
    \end{subfigure}
    \begin{subfigure}[b]{0.19\linewidth}
        \centering
        \includegraphics[width=\linewidth]{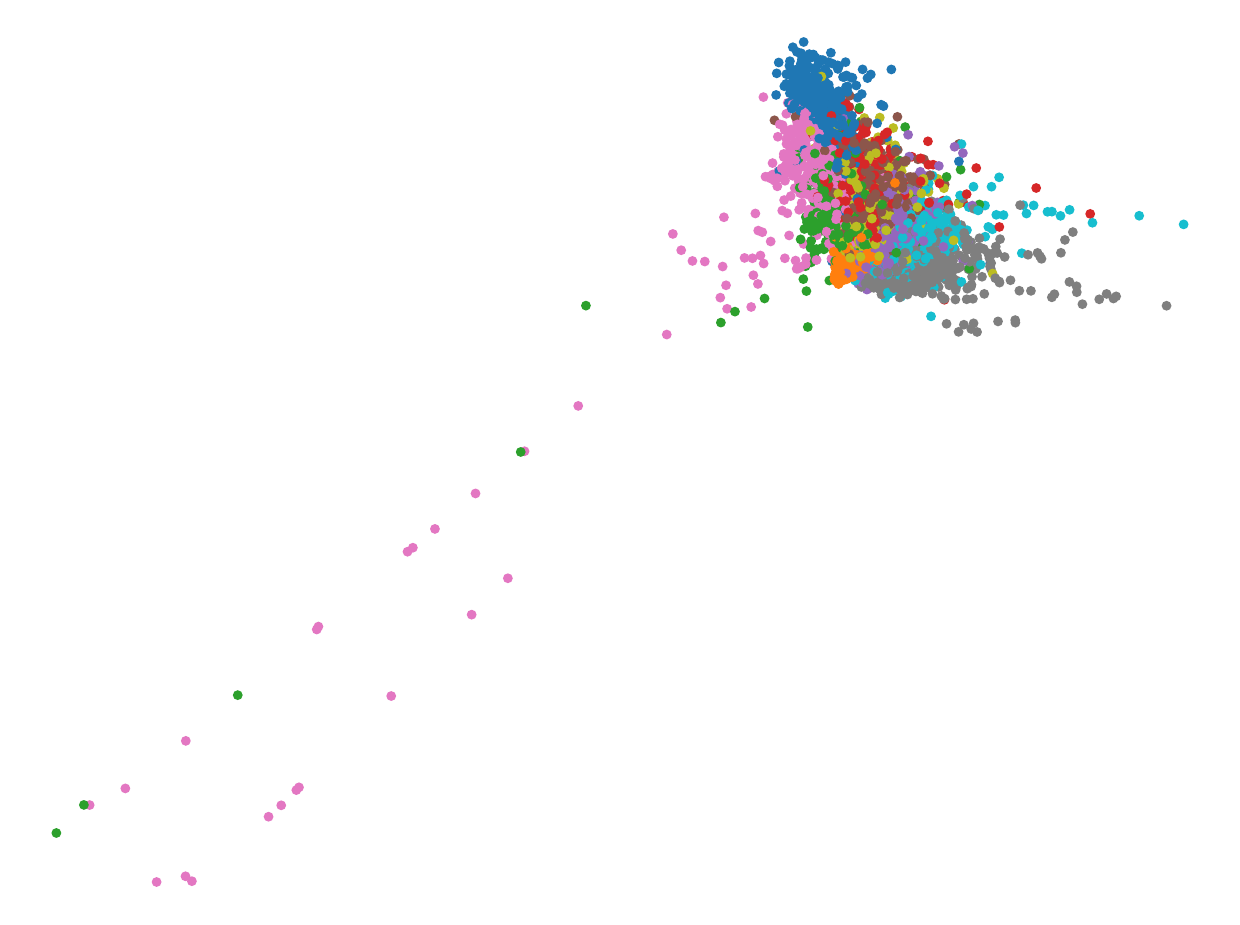}
        \caption{$k=4$, Possibility \#2}
        \label{fig:isomap_flipping:k4_2}
    \end{subfigure}
    \begin{subfigure}[b]{0.19\linewidth}
        \centering
        \includegraphics[width=\linewidth]{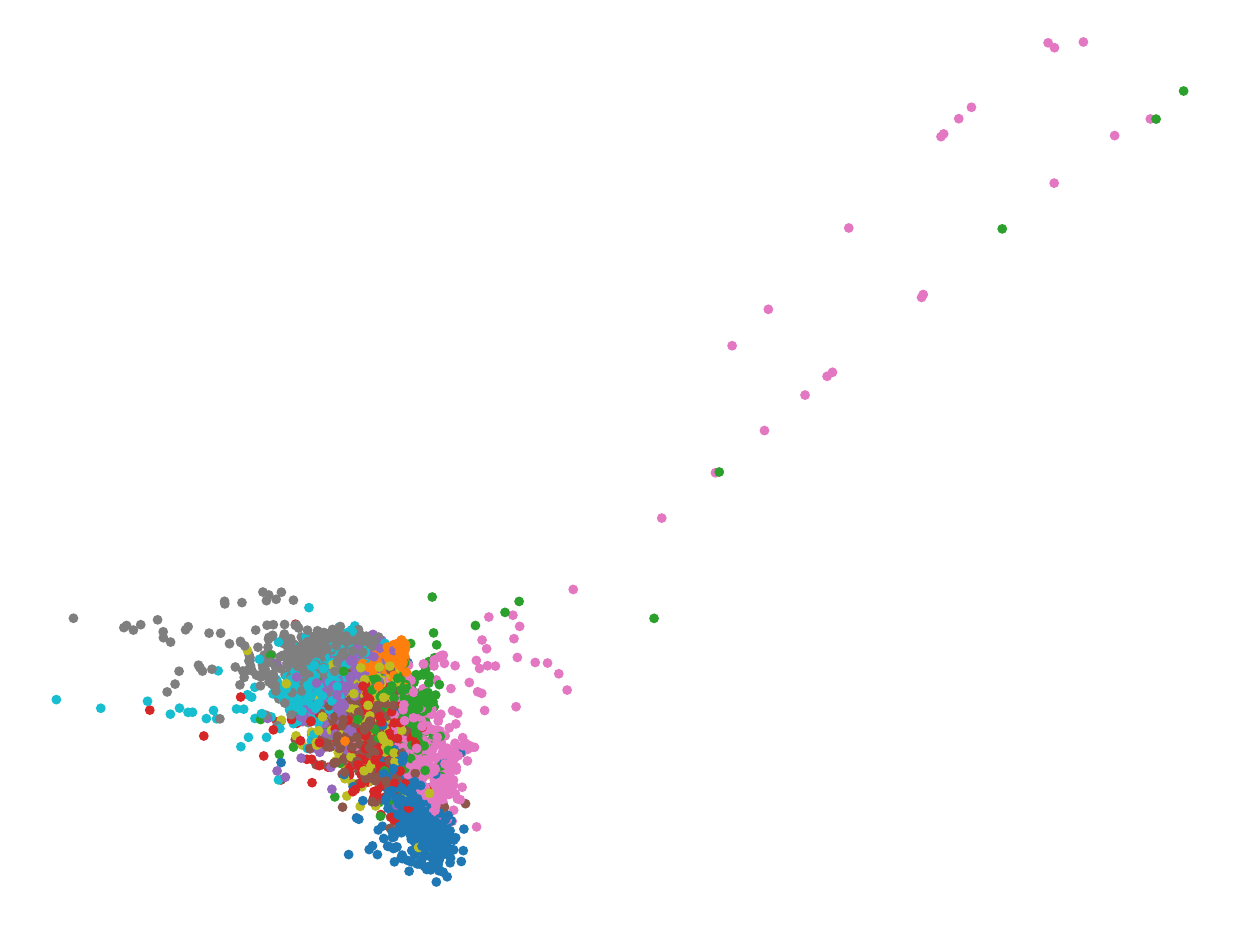}
        \caption{$k=4$, Possibility \#3}
        \label{fig:isomap_flipping:k4_3}
    \end{subfigure}
    \begin{subfigure}[b]{0.19\linewidth}
        \centering
        \includegraphics[width=\linewidth]{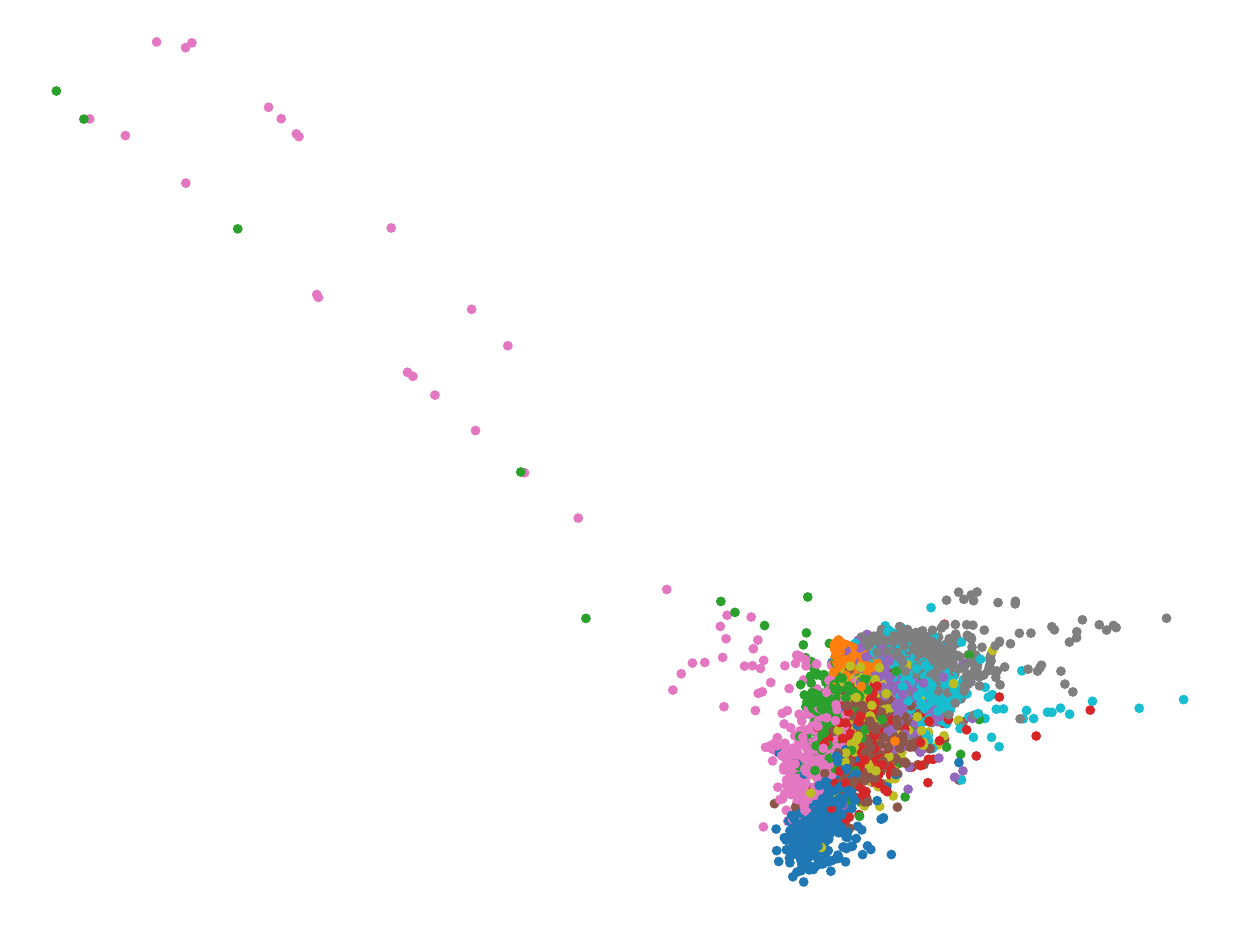}
        \caption{$k=4$, Possibility \#4}
        \label{fig:isomap_flipping:k4_4}
    \end{subfigure}
    \caption{Five Isomap projections of the MNIST dataset. (\subref{fig:isomap_flipping:k3}) uses $\mathbf{h}=3$ nearest neighbor, while (\subref{fig:isomap_flipping:k4_1}-\subref{fig:isomap_flipping:k4_4}) use $\mathbf{h}=4$. (\subref{fig:isomap_flipping:k4_1}-\subref{fig:isomap_flipping:k4_4}) show the four possible orientations due to Eigenanalysis instability (see Section~\ref{sec:visualization_considerations} for discussion) that can result from Isomap. Our technique produces image (b) -- the most similar of the four to the original image (a) in orientation.}
    \label{fig:isomap_flipping}
\end{figure*}
To train $\hat{P}$ to approximate any projection algorithm $P$, we seek to minimize the difference between the projections output by $\hat{P}$ and those of $P$. Given $\hat{P}$'s parameters $\theta$, this objective can be specified as: 
\begin{align}
\label{eq:full_objective}
\argmin_{\theta} &
\frac{\sum_{\mathbf{x}\in D} \sum_{\mathbf{h}\in H} \|\hat{P}_{\theta}([\mathbf{x}; \mathbf{h}]) - P(\mathbf{x}, \mathbf{h})\|^2}{|D| \cdot |H|}
\end{align}
where $D$ is the input dataset and $H$ is the set of all possible values of $\mathbf{h}$.
We further show empirically (Sec.~\ref{sec:validation}) that training $\hat{P}$ does not require minimizing Eqn.~\ref{eq:full_objective} for all data points and across all $\mathbf{h}$ values.
Depending on the algorithm $P$ that $\hat{P}$ aims to approximate, we can use as little as 20\% of all data points in $D$; for a hyperparameter $\mathbf{h}$, we can use sampling rates of 6 to 16 during training.
Hence, we rewrite Eqn.~\ref{eq:full_objective} as: 
\begin{align}
\label{eq:sampled_objective}
\argmin_{\theta} &
\frac{\sum_{\mathbf{x}\in D'} \sum_{\mathbf{h}\in H'} \|\hat{P}_{\theta}([\mathbf{x}; \mathbf{h}]) - P(\mathbf{x}, \mathbf{h})\|^2}{|D'| \cdot |H'|}
\end{align}
where $D' \subset D$ is the training set and $H' \subset H$ is the set of sampled values of the hyperparameters $\mathbf{h}$, respectively.
\subsection{Model Training in Practice}
\label{sec:method:training_in_practice}
Training $P$ follows the typical workflow for training a neural network, except the generation of the training data that includes both the selection of the data points and the choice of hyperparameter values.
To illustrate this, consider using \sys to explore UMAP's nearest-neighbor hyperparameter. 
In this example, we aim to train $\hat{P}$ using $20$\% of the data and with a gap size of $6$ for sampling the values of the hyperparameter that range from $2$ to $20$.
We begin generating the training data by randomly sampling $20$\% of $D$. 
We then project the data four times, at nearest neighbor values of $\{2, 8, 14, 20\}$.
Once we have all of our projections we stack the training data four times, once for each projection, and then augment it by concatenating the nearest neighbor value used in the projection.
Finally, we train the network following usual procedures, using the augmented stacked matrix as the input, and the UMAP projections as the desired output.
\subsection{Stability Considerations}
\label{sec:visualization_considerations}
\begin{figure}[h!]
    \centering
    \begin{subfigure}[b]{0.32\linewidth}
        \centering
        \includegraphics[width=\linewidth]{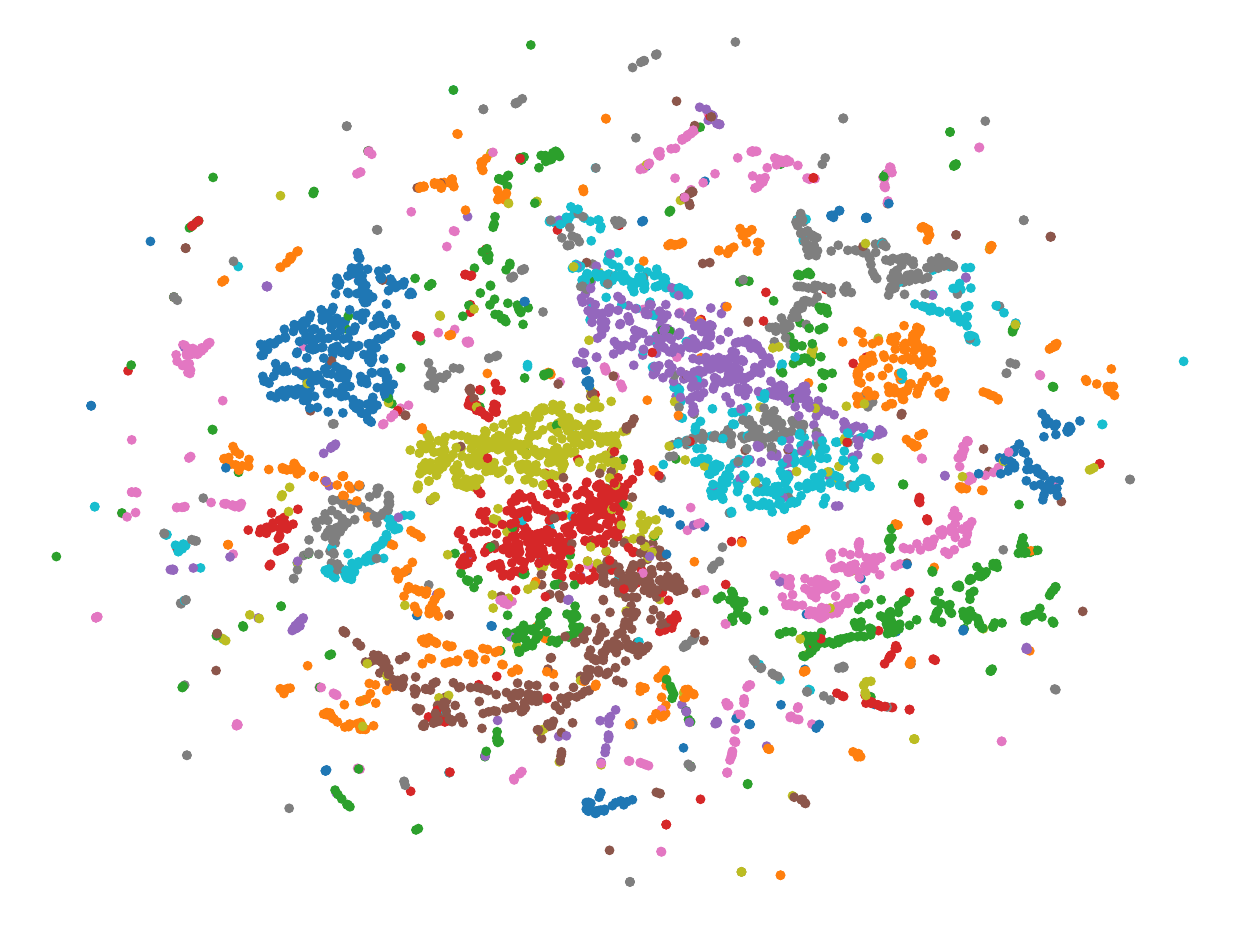}
        \caption{$k=3$, Random}
        \label{fig:umap_flipping:k3_random}
    \end{subfigure}
    \begin{subfigure}[b]{0.32\linewidth}
        \centering
        \includegraphics[width=\linewidth]{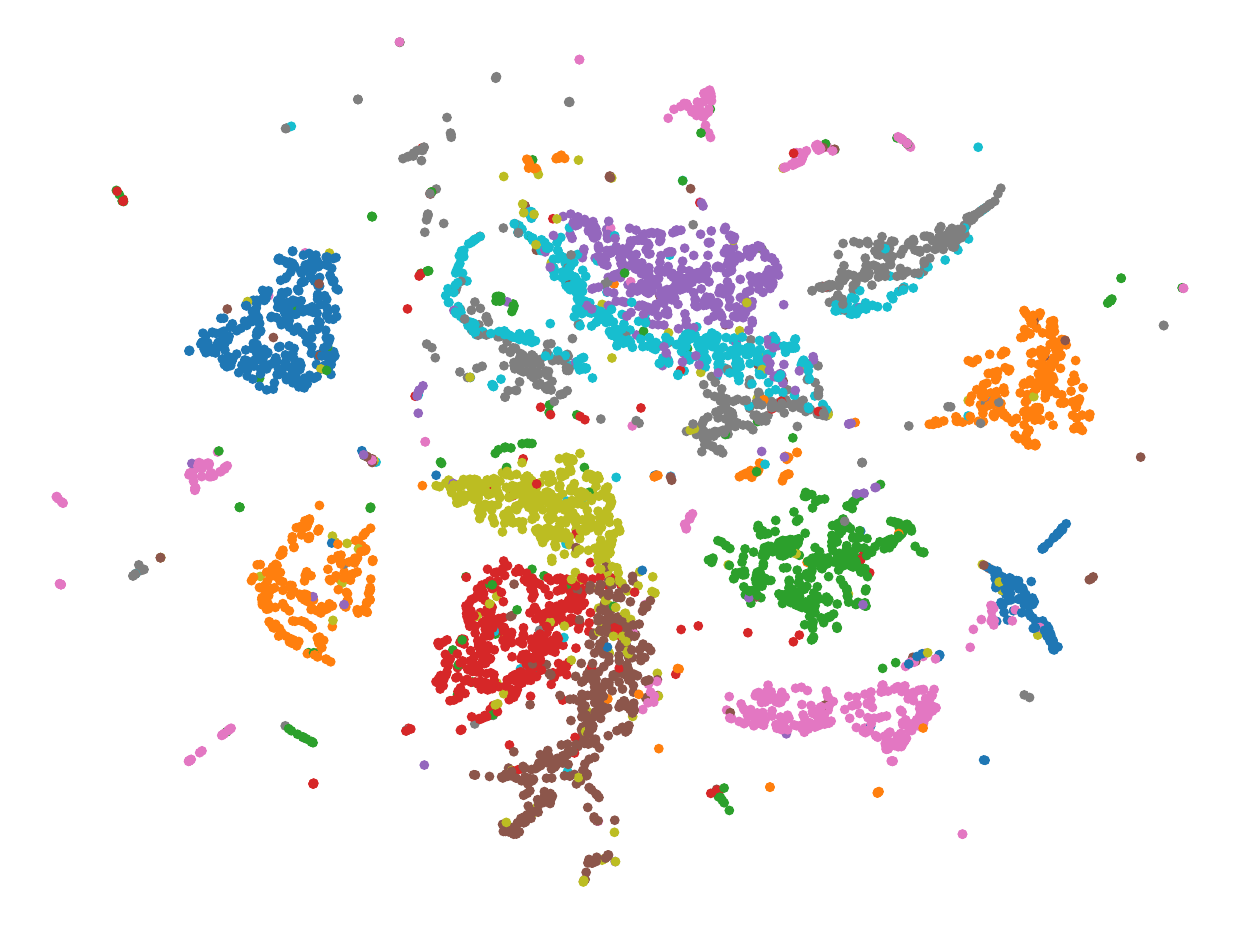}
        \caption{$k=4$, Previous}
        \label{fig:umap_flipping:k4_prev}
    \end{subfigure}
    \begin{subfigure}[b]{0.32\linewidth}
        \centering
        \includegraphics[width=\linewidth]{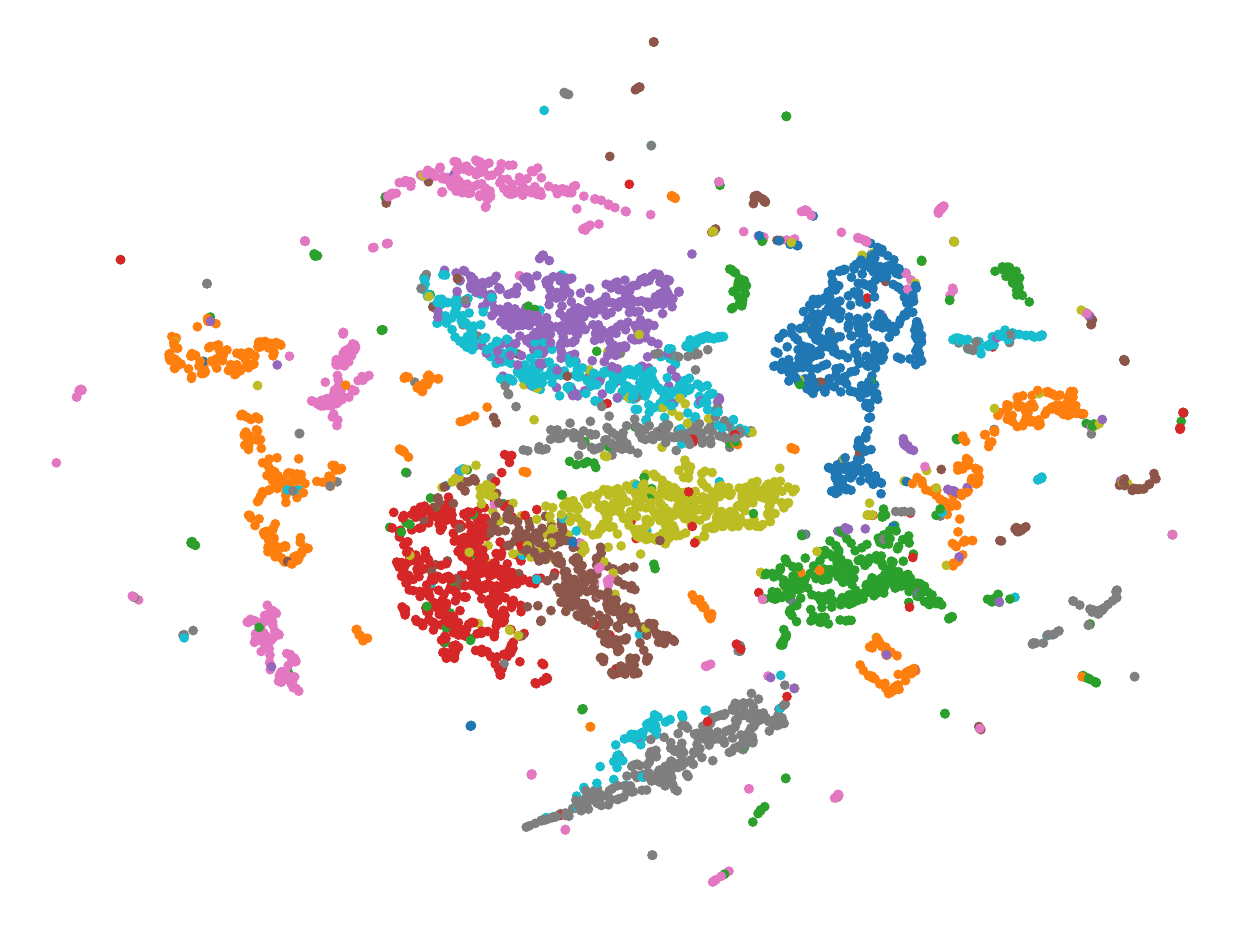}
        \caption{$k=4$, Random}
        \label{fig:umap_flipping:k4_random}
    \end{subfigure}
    \caption{Three UMAP projections of the MNIST dataset. (\subref{fig:umap_flipping:k3_random}) uses three nearest neighbors when constructing the kNN graph, while (\subref{fig:umap_flipping:k4_prev}) and (\subref{fig:umap_flipping:k4_random}) both use four nearest neighbors. (\subref{fig:umap_flipping:k4_prev}) initializes its embedding with positions from (\subref{fig:umap_flipping:k3_random}), while (\subref{fig:umap_flipping:k4_random}) is initialized as usual. Note that (a) and (c) appear differently due to the lack of projection stability, whereas (a) and (b) are more similar because of the initialization.}
    \label{fig:umap_flipping}
\end{figure}
Particular attention must be given to the issue of projection \emph{stability}.
Consider $P$ as a function of the hyperparameters $\mathbf{h}$ solely -- that is, we fix the dataset $D$.
Then, we argue that $P(\mathbf{h})$ should be a relatively smooth function -- small changes of $\mathbf{h}$ should yield only small changes of $P(\mathbf{h})$.
Indeed, if this were not the case, $P$ would confuse users by producing \emph{e.g.} sudden jumps in the scatterplot when varying $\mathbf{h}$.
This issue of stability \emph{vs} hyperparameters $\mathbf{h}$ is analogous to the stability of projections \emph{vs} changes in the data $D$, which was studied in detail for dynamic projections\,\cite{vernier:2020:quant_eval_time_dependent_projection}. 
As discussed in\,\cite{vernier:2020:quant_eval_time_dependent_projection}, not all projection techniques are stable. We next identify and address two types of instabilities.
\vspace{4pt}
\noindent\textbf{Seeding instabilities:} Many stochastic techniques start from a random 2D scatterplot which they iteratively update to minimize their cost\,\cite{maaten:2008:tsne,joia:2011:lamp}.
This makes $P$ not smooth, since even for \emph{no} parameter changes, $P$ depends on the random seed (see \emph{e.g.} the discussion in\,\cite{espadoto:2020:dlmp}, Fig. 11).
We address this as follows.
Let $\mathbf{h}_i$ be the samples of the hyperparameter $\mathbf{h}$ used during training, ordered ascendingly.
We construct the training projection $P_i = P(D,\mathbf{h}_i)$ as usual, \emph{i.e.}, running $P$ with random seeding.
Next, we construct $P_{i+1}$ similarly, but use $P_{i}$ to initialize the low-dimensional embedding of $P_{i+1}$.
A similar strategy was used in\,\cite{rauber16} to construct stable t-SNE projections for dynamic datasets.
Figure~\ref{fig:umap_flipping} illustrates this.
Image \ref{fig:umap_flipping}(a) shows a UMAP projection of the MNIST dataset using the hyperparameter $\mathbf{h}=3$ nearest neighbors.
Simply re-running UMAP for $\mathbf{h}=4$, the next possible hyperparameter value, yields image \ref{fig:umap_flipping}(c), in which many of the point clusters move significantly as compared to \ref{fig:umap_flipping}(a), as the color based on point labels shows -- see the light blue, dark purple, and green clusters.
In contrast, image \ref{fig:umap_flipping}(b) seeds UMAP for $\mathbf{h}=4$ by the 2D scatterplot in \ref{fig:umap_flipping}(a).
The resulting projection is now much closer to \ref{fig:umap_flipping}(a), which is desired.
\vspace{4pt}
\noindent\textbf{Eigenanalysis instabilities:} Many projection techniques, starting with PCA\,\cite{pearson:1901:pca}, project data along eigenvectors which have no \emph{sign} meaning.
As such, the resulting projections can easily `flip' along one of the 2D $x$ or $y$ axes upon slight changes in the input data or hyperparameters.
Such drastic changes convey no meaning and should be eliminated.
We address this problem by a similar solution to the pose-invariance fix used in multimedia retrieval\,\cite{blanken07} for comparing arbitrarily rotated shapes.
We compute the mean square error (MSE) between a training projection $P_i$ and the previous projection $P_{i-1}$ for all four possible mirrorings, \emph{i.e.}, multiplying the $x$ and $y$ projection axes by $-1$, and pick for $P_i$ the configuration with minimal MSE.
This minimizes undesired changes between $P_i$ and $P_{i-1}$ during training, thus makes $\hat{P}$ more stable.
Figure~\ref{fig:isomap_flipping} shows this.
Image \ref{fig:isomap_flipping}(a) shows Isomap run with $\mathbf{h}=3$ nearest neighbors on the MNIST dataset.
Images \ref{fig:isomap_flipping}(b-e) show Isomap on the same dataset for $\mathbf{h}=4$ where the projections result in different ``flipped'' orientations.
Our heuristic results in \ref{fig:isomap_flipping}(b) -- the most similar to the original \ref{fig:isomap_flipping}(a) in orientation.
\subsection{Implementation and Tuning}
\label{sec:implementation_and_tuning}
We implemented the neural network used in \sys using Tensorflow\,\cite{abadi:2015:tensorflow} and Keras\,\cite{chollet:2015:keras}.
We used Keras-Tuner's\cite{google:2019:kerastuner} Bayesian Optimization to find the appropriate number of layers, neurons per layer, and regularization.
The tuning consisted of 512 trials, with the first 40 using random search as initial points.
The search space included the number of layers (two to six), number of neurons per layer (32 to 512 in steps of 32), dropout probability\,\cite{srivastava:2014:dropout} (0, 0.25, or 0.5), and whether or not to use batch normalization\cite{ioffe:2015:batchnorm}.
The final configuration used was a network with layer sizes $|V_2|=320, |V_3| = 256, |V_3| = 352, |V_4|=2$. We used ReLU\cite{nair:2010:relu} activation for all layers except the final one ($V_4)$ where no activation function was used (linear). Between all layers we used batch normalization followed by a dropout probability of 0.25.
We chose mean absolute error as the loss function and the ADAM\,\cite{kingma:2014:adam}  optimizer, following NNP\,\cite{espadoto:2020:dlmp}.
\section{Applications}
\label{sec:applications}
We illustrate the value of \sys by showing how it approximates changes in the common hyperparameters provided by three popular projection methods: t-SNE, UMAP, and Isomap.
\subsection{Exploring Perplexity in t-SNE}
\label{sec:applications:tsne}
\begin{figure}[h!]
     \centering
     \begin{subfigure}[b]{0.32\linewidth}
         \centering
         \includegraphics[width=\linewidth]{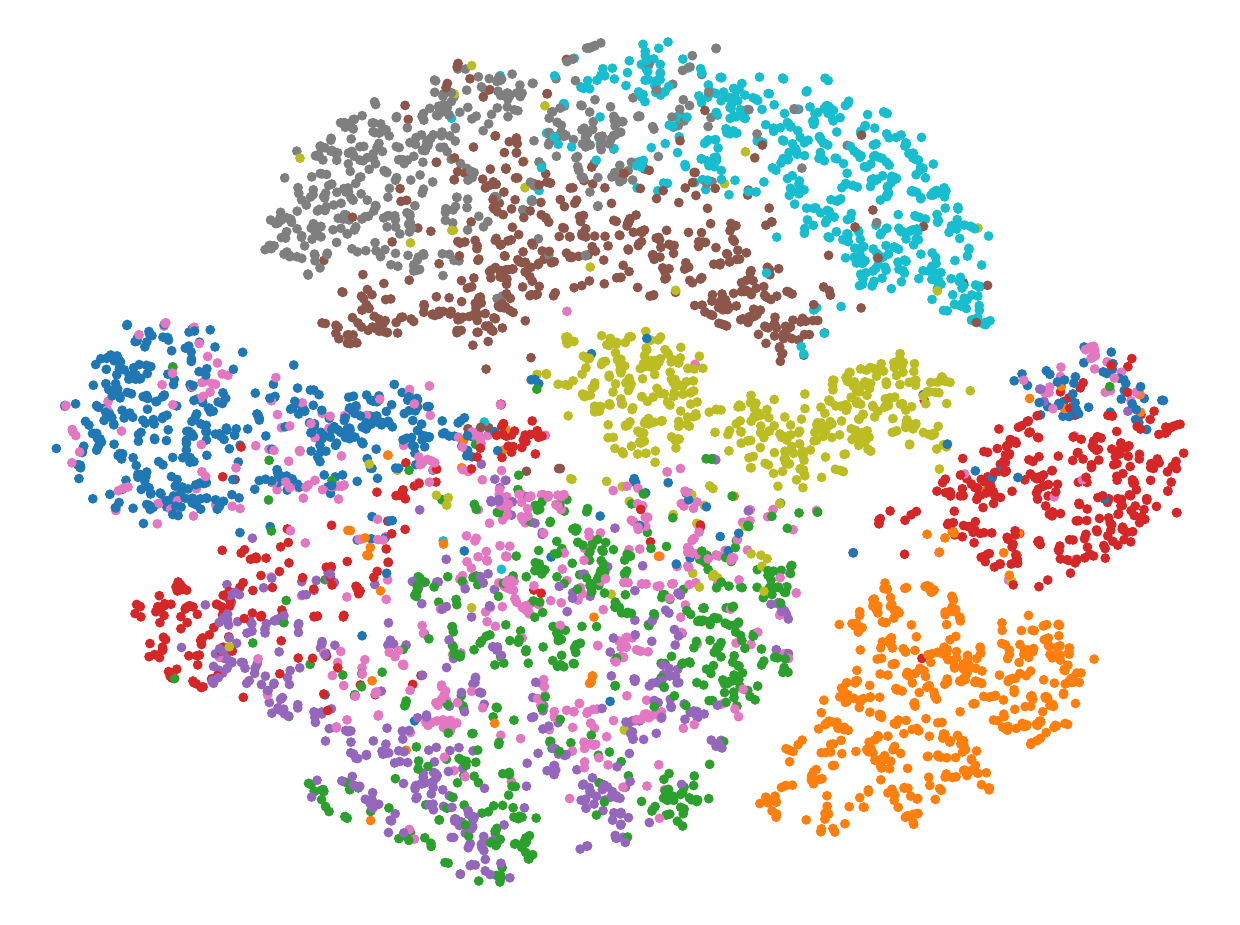}
         \caption{Ground Truth Train Data}
         \label{fig:tsne_train_test_both:train_gt}
     \end{subfigure}
     \begin{subfigure}[b]{0.32\linewidth}
         \centering
         \includegraphics[width=\linewidth]{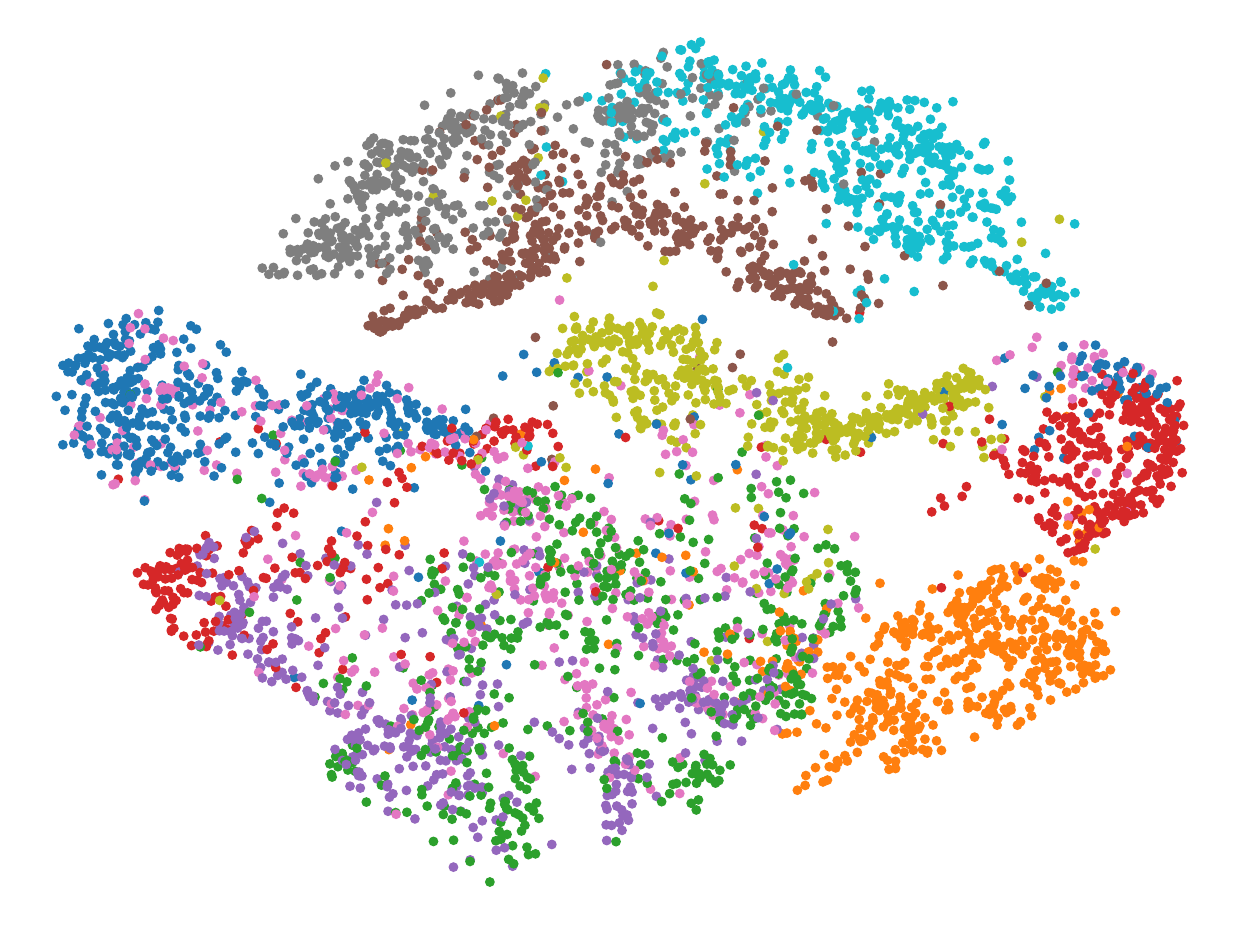}
         \caption{Train Predictions}
         \label{fig:tsne_train_test_both:train_pred}
     \end{subfigure}
     \begin{subfigure}[b]{0.32\linewidth}
         \centering
         \includegraphics[width=\linewidth]{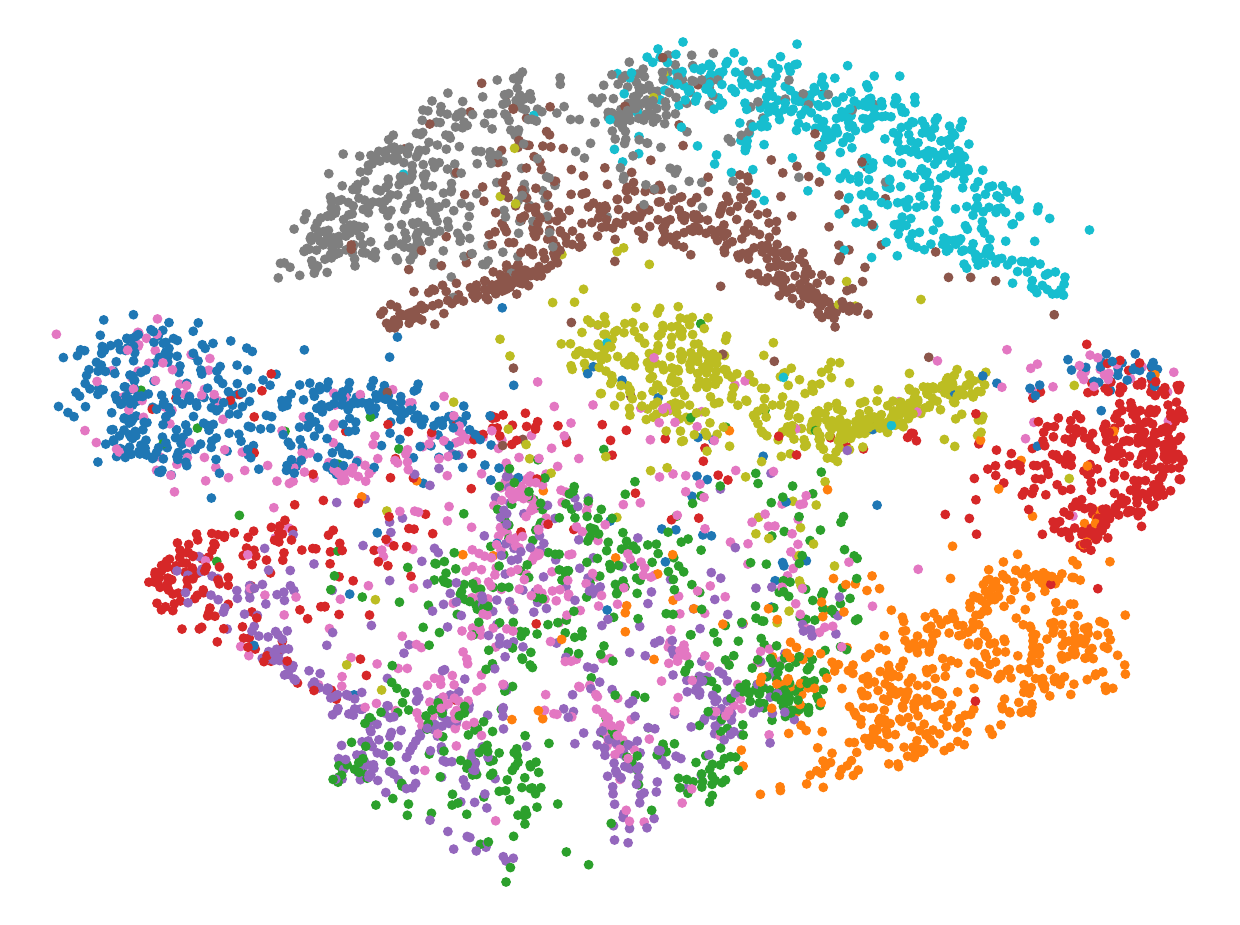}
         \caption{Test Predictions}
         \label{fig:tsne_train_test_both:test_pred}
     \end{subfigure}
    \caption{\sys approximation of the t-SNE projections of the FashionMNIST dataset.  From left to right:  (\subref{fig:tsne_train_test_both:train_gt}) t-SNE projection used for training with 20\% data, (\subref{fig:tsne_train_test_both:train_pred}) \sys projection of the same data used in (a), (\subref{fig:tsne_train_test_both:test_pred}) \sys projection of test data instances unseen during training. This result suggests that \sys is adequately learning the t-SNE projection using just 20\% of the data as these three images are visually similar.}
    \label{fig:tsne_train_test_both}
\end{figure}
\begin{figure}[b!]
    \centering
    \begin{subfigure}[b]{0.32\linewidth}
        \centering
        \includegraphics[width=\linewidth]{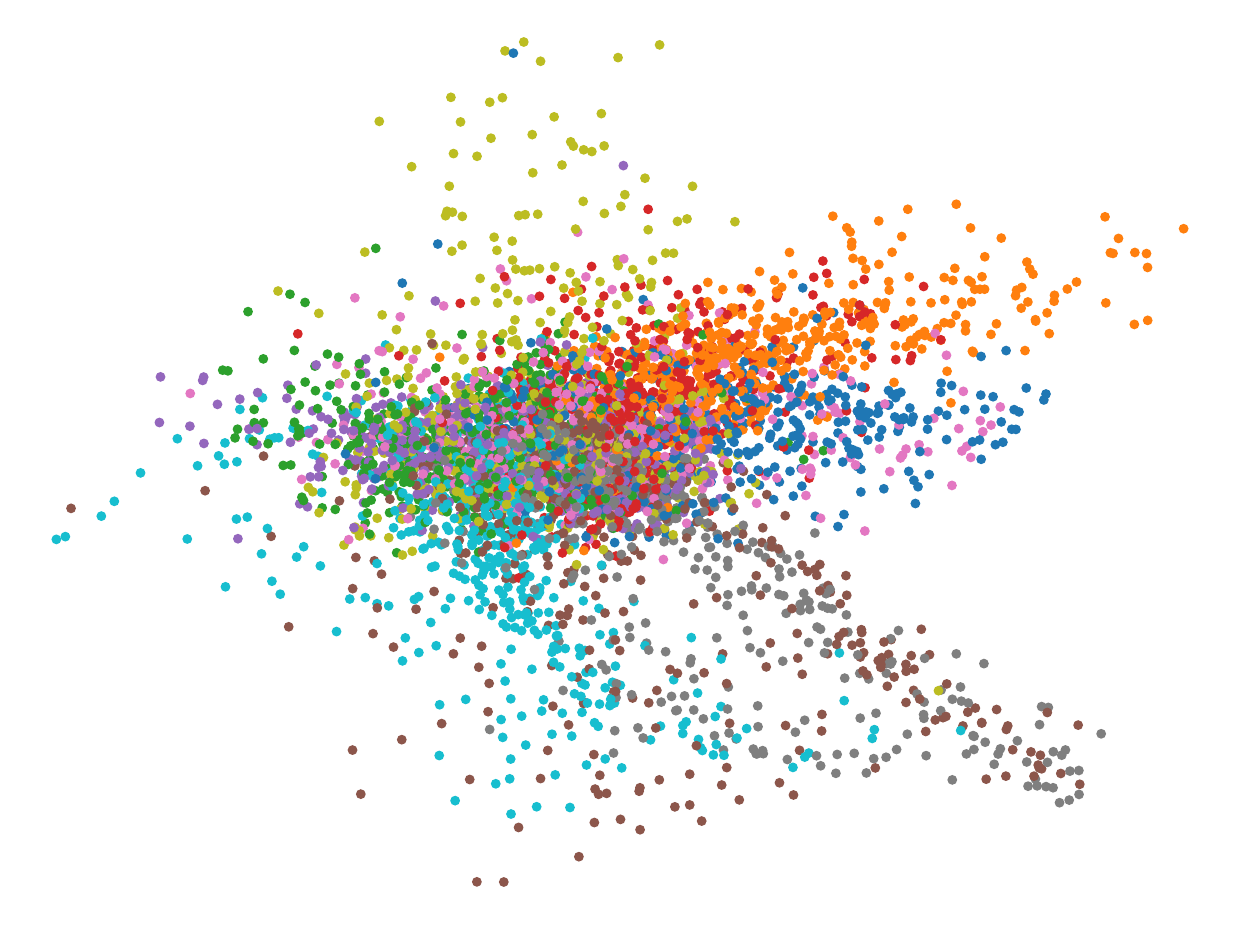}
        \caption{k=3}
        \label{fig:umap_noninteer_k:k3}
    \end{subfigure}
    \begin{subfigure}[b]{0.32\linewidth}
        \centering
        \includegraphics[width=\linewidth]{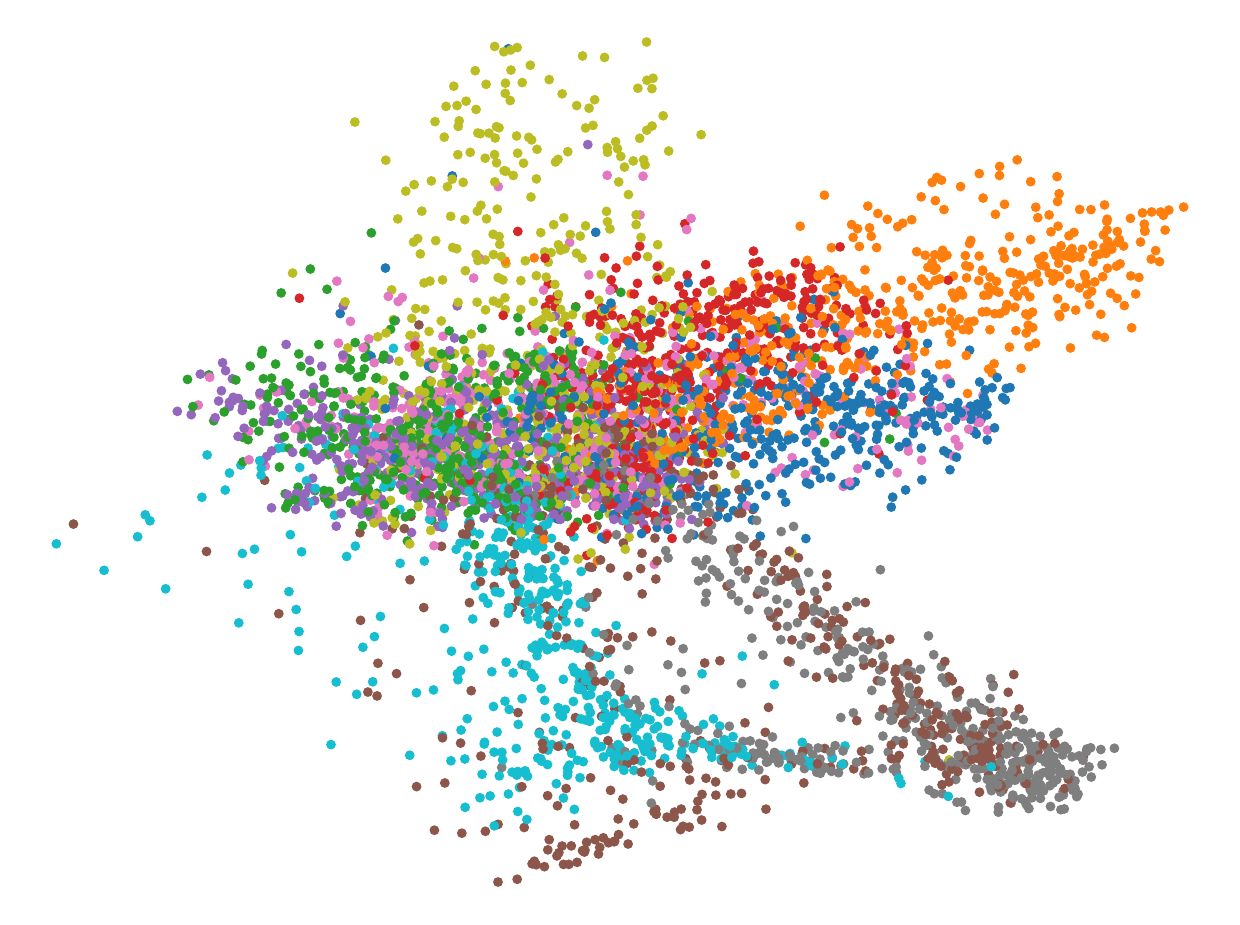}
        \caption{k=3.5}
        \label{fig:umap_noninteer_k:k3.5}
    \end{subfigure}
    \begin{subfigure}[b]{0.32\linewidth}
        \centering
        \includegraphics[width=\linewidth]{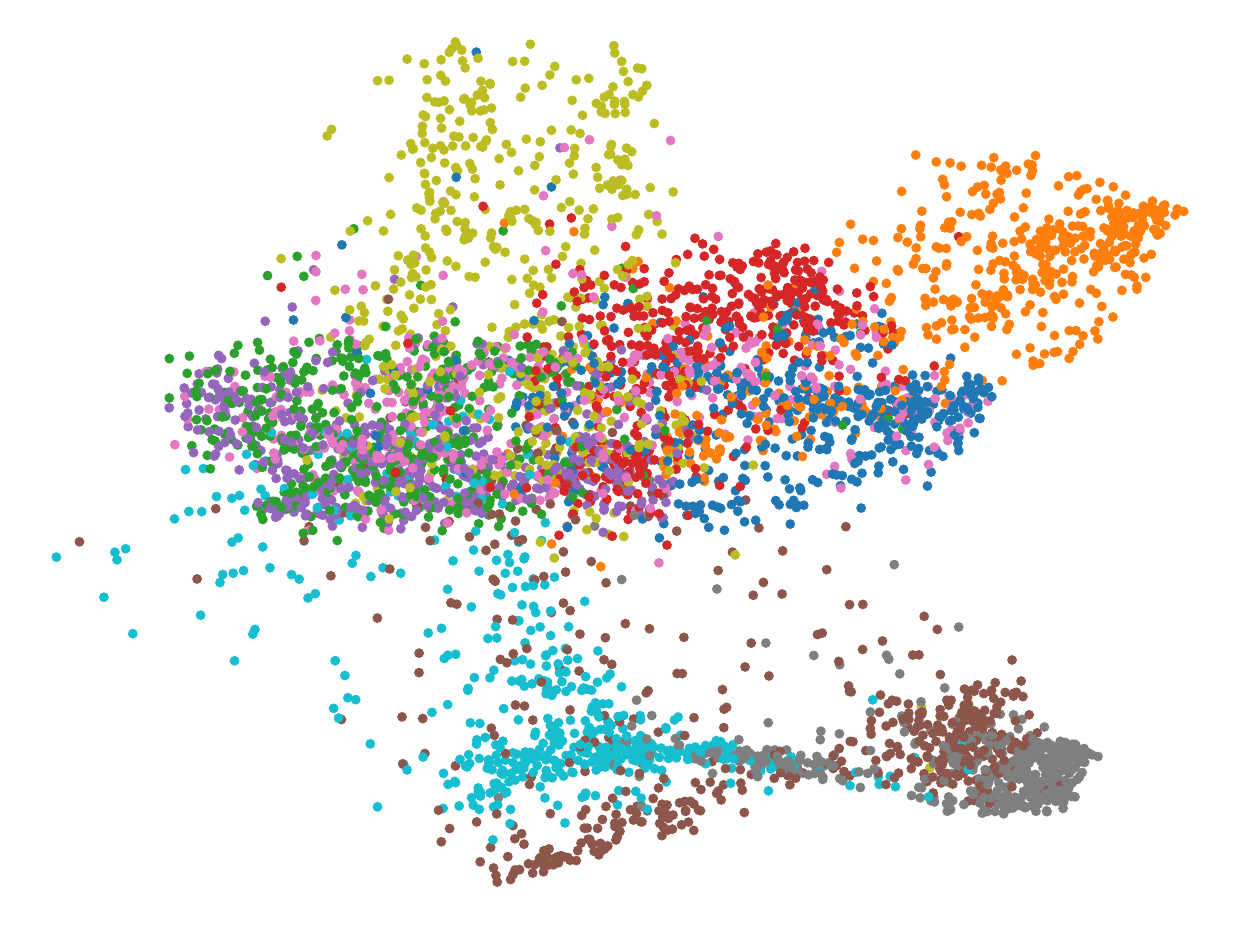}
        \caption{k=4}
        \label{fig:umap_noninteer_k:k4}
    \end{subfigure}
    \caption{Using \sys to explore values of the number of nearest-neighbor hyperparameter, $k$, in UMAP.  From left to right: (\subref{fig:umap_noninteer_k:k3}) $k=3.0$, (\subref{fig:umap_noninteer_k:k3.5}) $k=3.5$, (\subref{fig:umap_noninteer_k:k4}) $k=4.0$.  While non-natural values for the parameter $k$ are not valid, we show that \sys is able to smoothly interpolate between meaningful values without sacrificing projection quality.}
    \label{fig:umap_noninteger_k}
\end{figure}
t-SNE is a variation of Stochastic Neighbor Embedding\,\cite{hinton:2002:sne} which takes a probabilistic approach to preserving point-wise similarity when mapping high-dimensional data to a low-dimensional (typically 2D) space.
First, a Gaussian distribution is constructed over the data samples so that similar sample pairs have a higher value than dissimilar ones. Another distribution is then constructed for the 2D space. t-SNE then places each sample in 2D by minimizing the Kullback-Leibler divergence between the two Gaussian Distributions.
The bandwidth of the Gaussian kernels is chosen so that the perplexity of the high dimensional distribution is equal to a user-defined \emph{perplexity} hyperparameter. This adapts the size of the Gaussian distribution to the underlying density of the input data -- smaller kernels are used in dense areas.
As Wattenberg \emph{et al.}\,\cite{wattenberg:2016:using-tsne-effectively} point out, the value of the perplexity hyperparameter in t-SNE profoundly affects the final 2D embedding produced.
Simply put, as perplexity increases, the importance of global features in the data also increases.
However, the computational cost of robustly exploring the effects of many perplexity values is prohibitive for large data. 
Figure~\ref{fig:tsne_train_test_both} shows how \sys is able to appropriately project out-of-sample data instances. 
This minimizes training time for each realized projection configuration, $\hat{P}([\mathbf{x};\mathbf{h}])$, allowing \sys to scale to large data with only moderate training cost.
While speed of both training and inference are important to the overall performance of \sys, care must be taken to ensure that the projection $\hat{P}$ is adequately learned.
As shown in Fig.~\ref{fig:tsne_train_test_both}, \sys has learned the t-SNE projection of the MNIST dataset. 
Qualitatively, the image created using only 20\% of the data is practically indistinguishable from the others.
This provides strong evidence that using only a small data fraction for training is enough to faithfully represent a chosen projection configuration.
Besides out-of-sample inference for \emph{data}, \sys can perform out-of-sample inference for \emph{hyperparameter} values.  
Table~\ref{fig:tsne_application} shows how \sys explores different perplexity values for the MNIST dataset.
Looking across rows, Table~\ref{fig:tsne_application} shows that not only does \sys maintain similar clustering (columns 2-4) to the full projection (column 1) for hyperparameters not seen in training, but also that increasing the gap size has only a marginal impact on the overall projection quality.
%
%
%
This sampling strategy accelerates the initial training of the model without sacrificing quality.
%

%
Both Wattenberg \emph{et al.}\,\cite{wattenberg:2016:using-tsne-effectively} and Silva \emph{et al.}\,\cite{silva:2002:globalvlocal} argue strongly for a thorough inspection of hyperparameters when interpreting projected data. Yet, as said earlier, this critical task is often skipped due to its high cost.
Not only does \sys provide the means to explore hyperparameters, but it does this for large datasets and without visible quality loss. 
The interactivity thus afforded allows users to better understand how the perplexity hyperparameter influences the final rendering of their data.
\subsection{Exploring k Nearest Neighbors in UMAP}
\label{sec:applications:umap}
UMAP is a projection technique based on ideas from topological data analysis and manifold learning.
Being based on the notion of fuzzy simplicial sets, it is implemented using weighted graph techniques\,\cite{mcinnes:2018:umap}.
First, UMAP constructs $k$-nearest-neighbor (kNN) graph, whose edge weights model the probability that the edge exists. The graph describes the locally connected manifold that the data is assumed to be spread on, and to guide a force-directed graph layout in low dimensions.
%
%
%
After initialization, performed by well-established spectral methods, a low dimensional graph is constructed from the projected points.
The points are repositioned such that the low-dimensional graph approximates the original weighted graph in high dimension.
The objective function minimized by the force directed layout is the total cross entropy of all edge probabilities, ensuring that the two graphs have similar topologies.
As with other nearest-neighbor based projections, a key hyperparameter of UMAP is $k$, the number of considered neighbors. 
Varying $k$ changes the local scale that the high dimensional manifold considers planar. 
Features that exist at smaller scale than the neighborhood size are thus lost, with larger features being better preserved in the projection. 
Silva \emph{et al.}\,\cite{silva:2002:globalvlocal} provide a full discussion of the importance of $k$ with respect to feature size for the class of projection techniques using manifold learning approaches. 
Typically, methods that use kNN graphs require all data to be integrated into the graph prior to projection.
One challenge of manifold learning based projection techniques is the projection of out-of-sample data which does not exist in the kNN graph used to determine relationships across data points.
\sys is able to project out-of-sample data without first embedding  new points in the existing kNN graph.
Figure~\ref{fig:umap_application} shows how \sys maintains projection quality of out-of-sample data and also the importance of hyperparameter selection for different gap sizes: smaller neighborhoods lose projection quality faster as the gap size increases.

Care must be taken to ensure that the gap size used to train \sys does not overwhelm the smaller values for neighborhood size.
The tradeoff between gap size and parameter lower bound is not unique to UMAP, but must be kept in mind for any manifold learning technique.
As $k$ changes in the low end of its range, the resulting change in the neighborhood topology is large: changing $k$ from 3 to 4 yields a larger change than changing $k$ from 20 to 21.
Projection error increases as small values of $k$ move away from the training $k$ values, which is expected, as the projection change most rapidly for low $k$.

Unlike t-SNE's perplexity hyperparameter, $k \in \mathbb{N}$. Still, \sys is able to both learn and infer all $k \in \mathbb{R}$ values, as shown in Fig.~\ref{fig:umap_noninteger_k}. That is, \sys can smoothly interpolate between the strict $k \in \mathbb{N}$ values.
%
\subsection{Exploring Neighborhood Size in Isomap}
\label{sec:applications:isomap}
\begin{figure}[h!]
     \centering
     \begin{subfigure}[b]{0.32\linewidth}
         \centering
         \includegraphics[width=\linewidth]{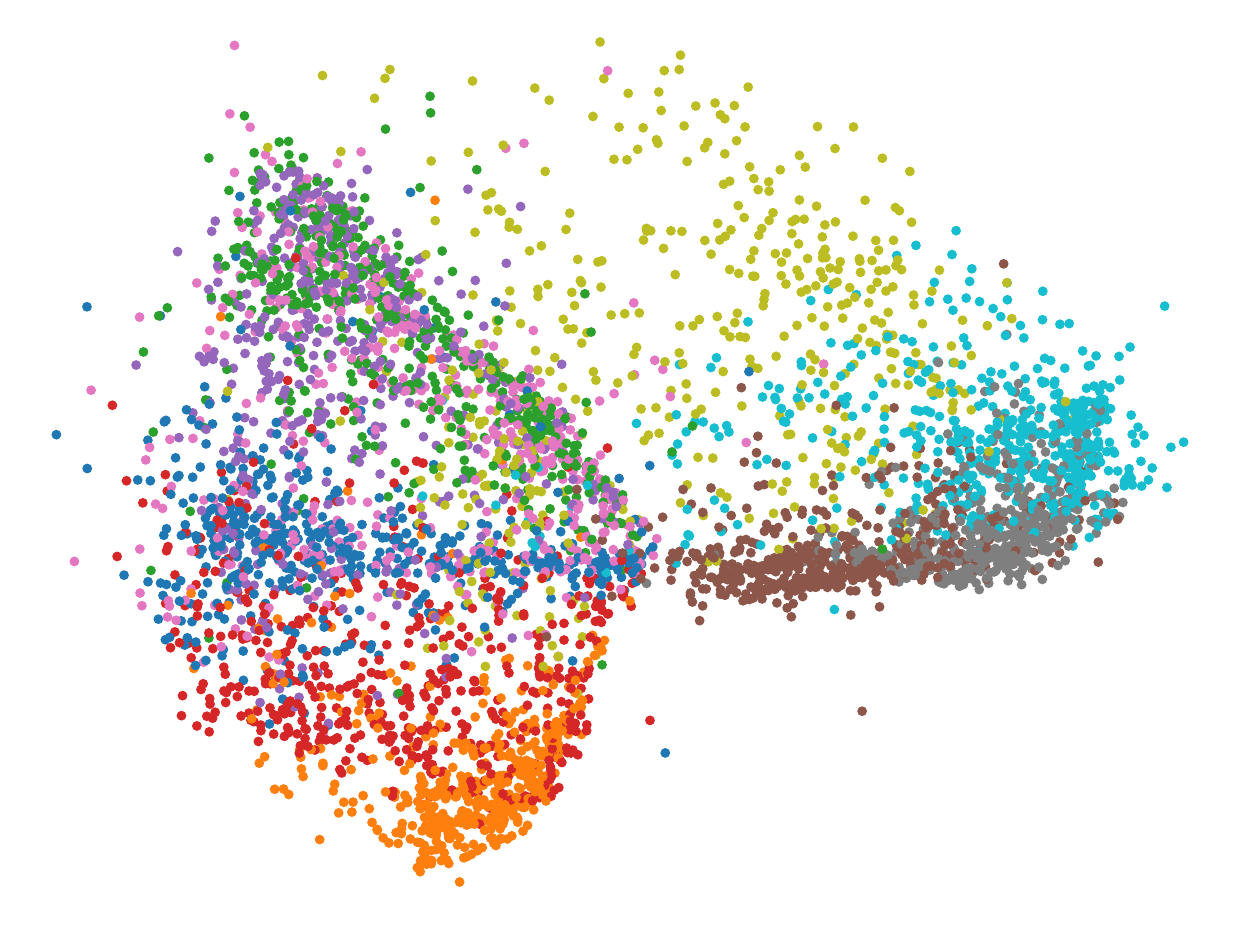}
         \caption{Ground Truth Train Data}
         \label{fig:isomap_train_test_both:train_gt}
     \end{subfigure}
     \begin{subfigure}[b]{0.32\linewidth}
         \centering
         \includegraphics[width=\linewidth]{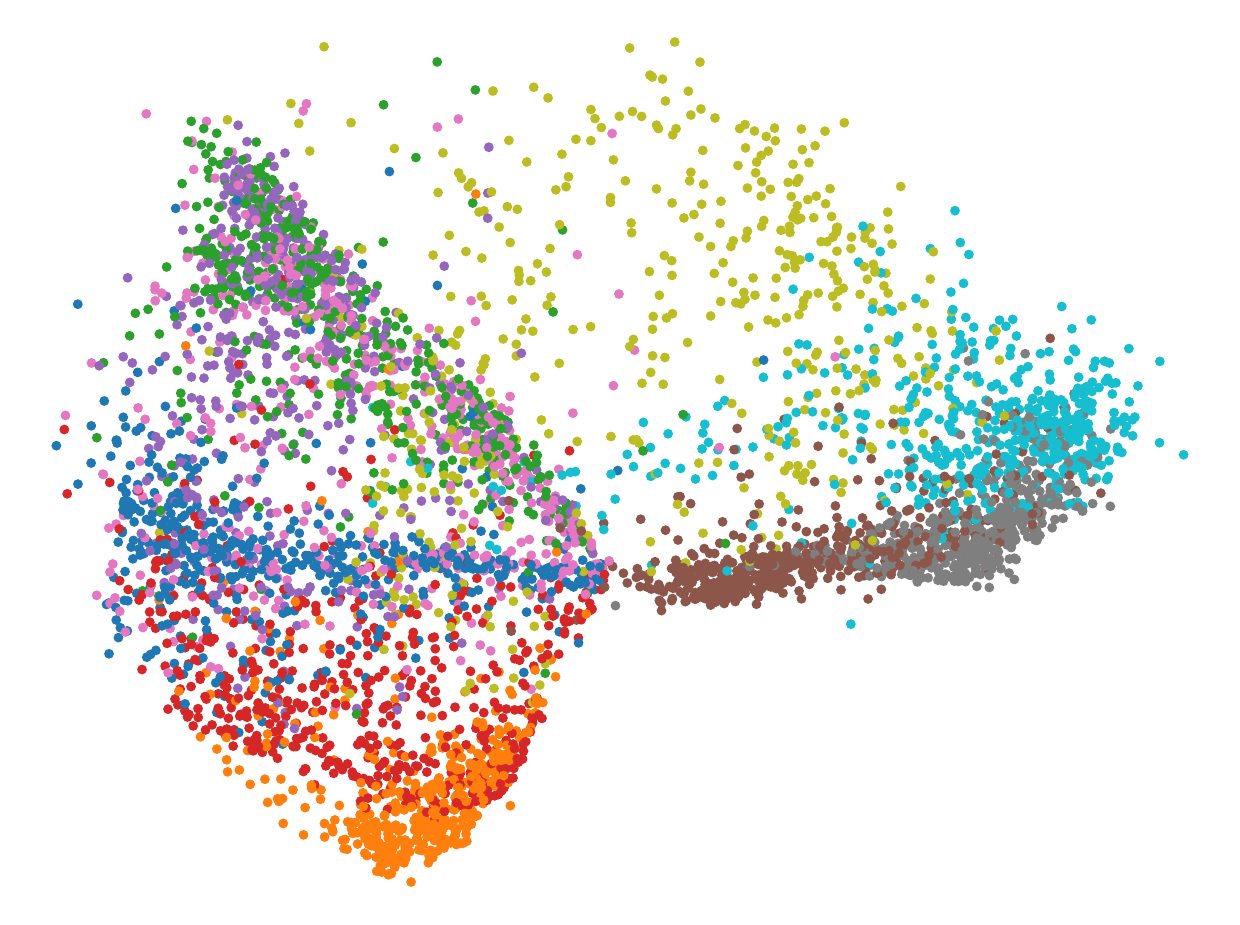}
         \caption{Train Predictions}
         \label{fig:isomap_train_test_both:train_pred}
     \end{subfigure}
     \begin{subfigure}[b]{0.32\linewidth}
         \centering
         \includegraphics[width=\linewidth]{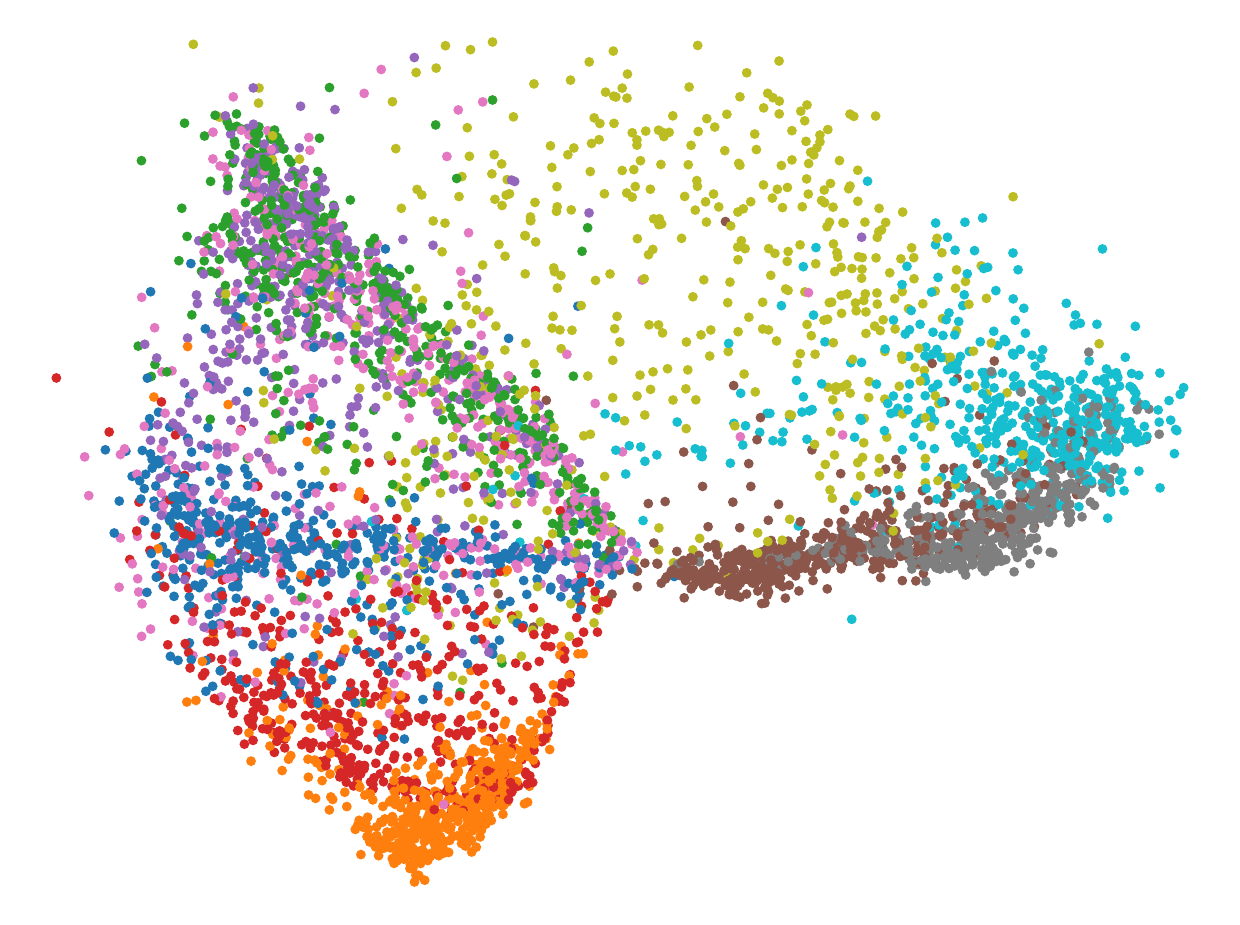}
         \caption{Test Predictions}
         \label{fig:isomap_train_test_both:test_pred}
     \end{subfigure}
    \caption{\sys approximation of the Isomap projections of the FashionMNIST dataset.  From left to right:  (\subref{fig:isomap_train_test_both:train_gt}) Isomap projection used for training with 20\% data, (\subref{fig:isomap_train_test_both:train_pred}) \sys projection of the same data used in (a), (\subref{fig:isomap_train_test_both:test_pred}) \sys projection of data instances unseen during training. This result suggests that \sys is adequately learning the Isomap projection using just 20\% of the data as these three images are visually similar.}
    \label{fig:isomap_train_test_both}
\end{figure}
\begin{table*}[htp!]
    \centering
    \begin{tabularx}{\textwidth}{c X X X X}
      \toprule
        \multirow{2}{*}{P} & \multirow{2}{*}{t-SNE} & \multicolumn{3}{c}{HyperNP}\\
        & & \multicolumn{1}{c}{Gap=2} & \multicolumn{1}{c}{Gap=4} & \multicolumn{1}{c}{Gap=8} \\
        \midrule
        \multicolumn{1}{c}{5} & \includegraphics[width=\linewidth]{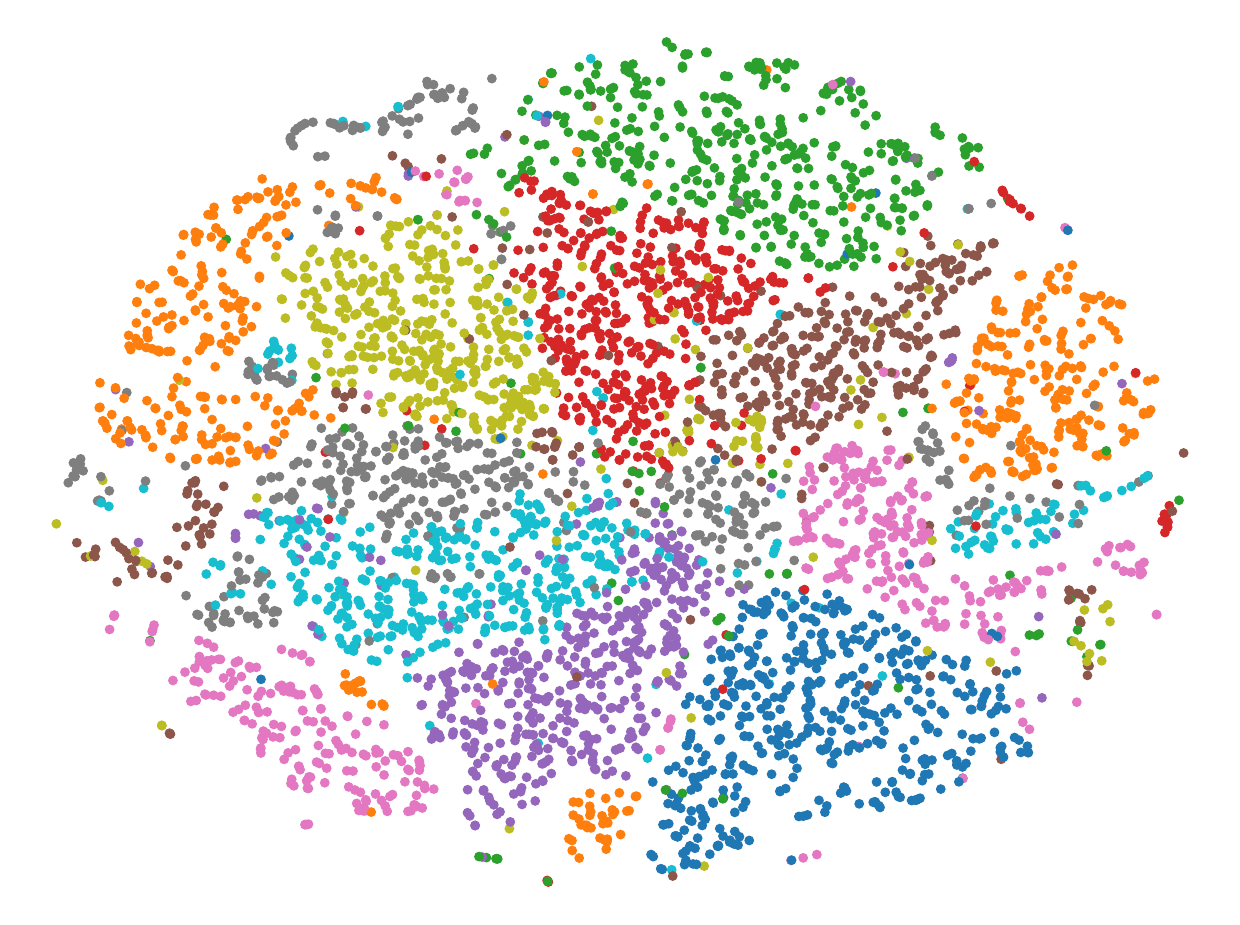} & \includegraphics[width=\linewidth]{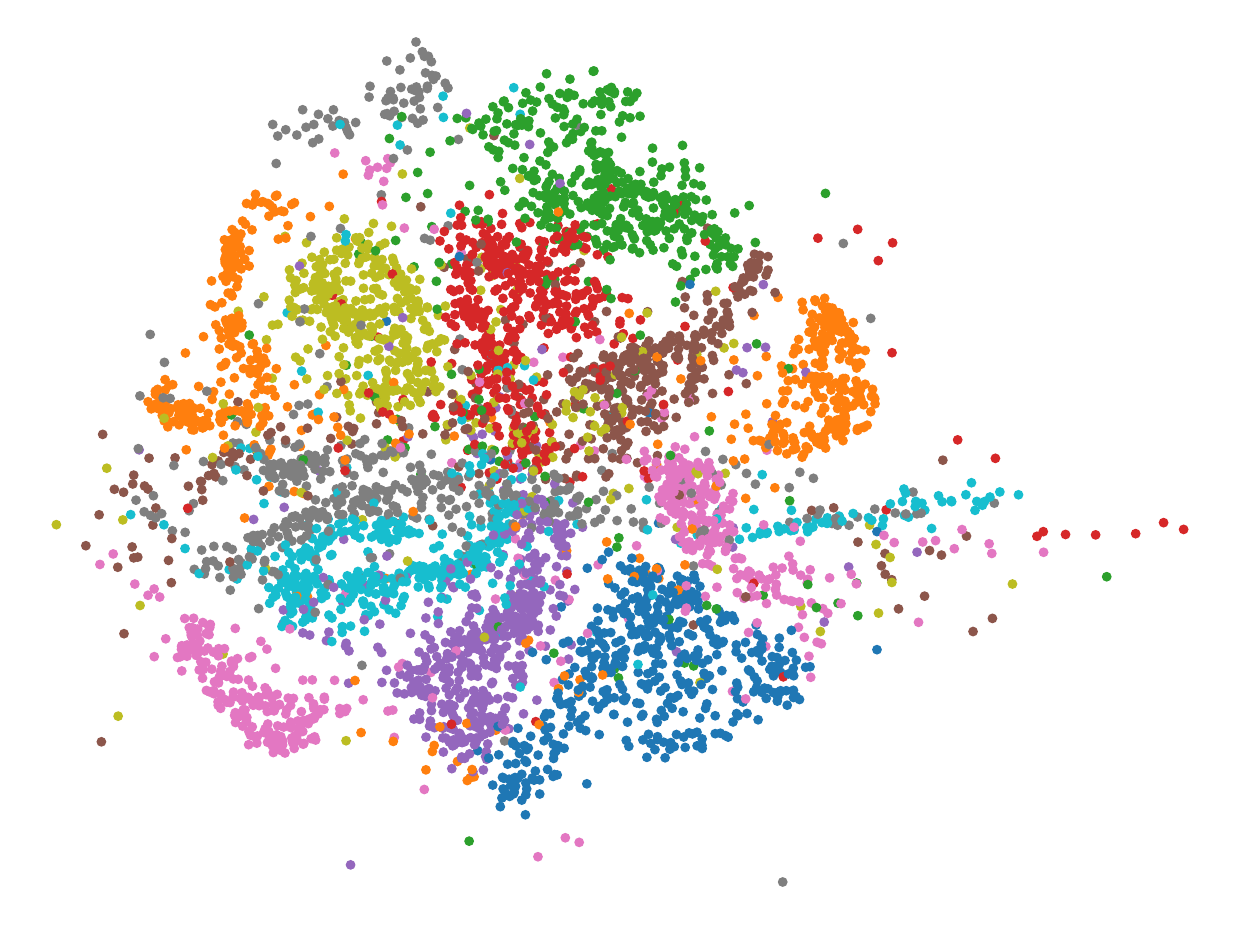} & \includegraphics[width=\linewidth]{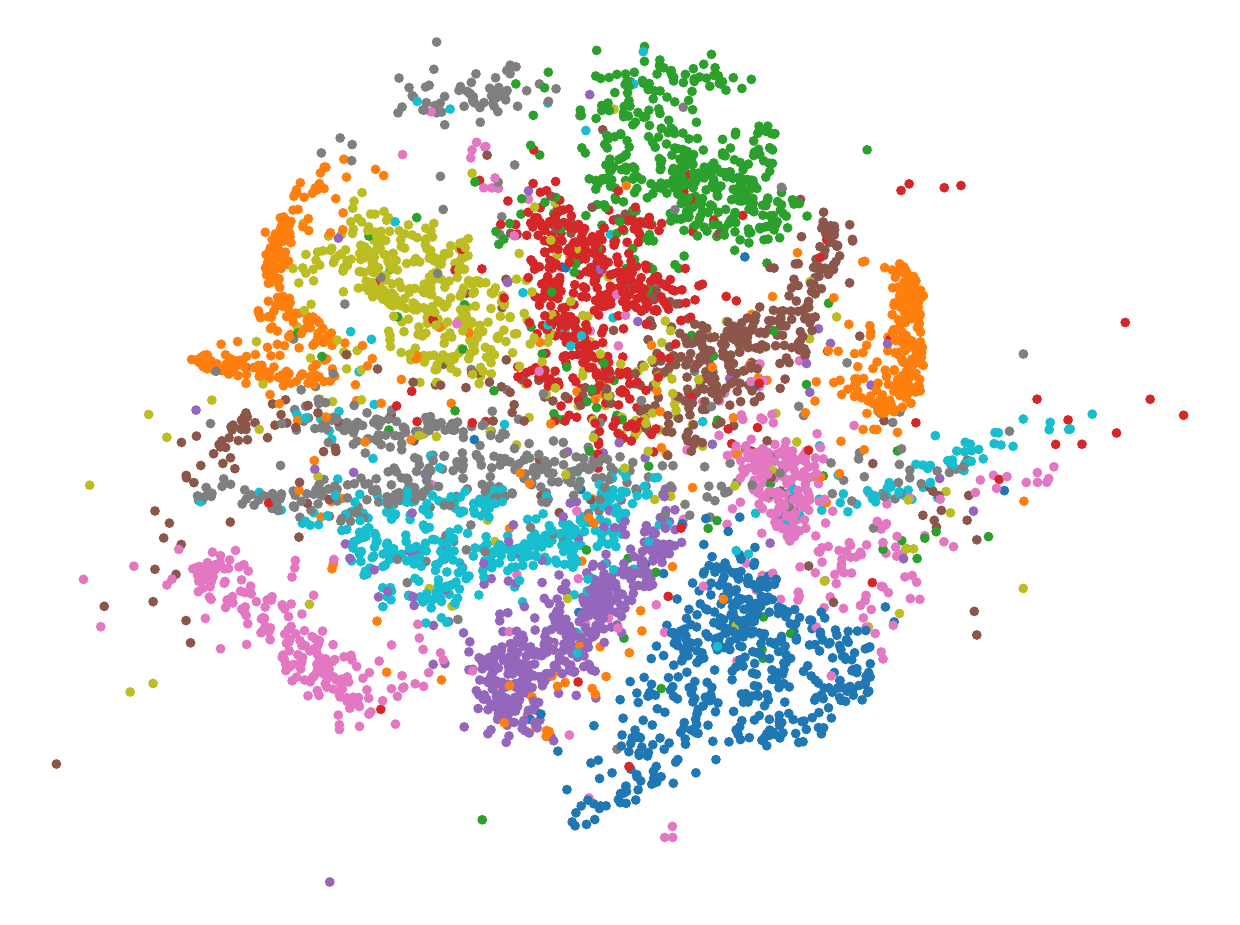} & \includegraphics[width=\linewidth]{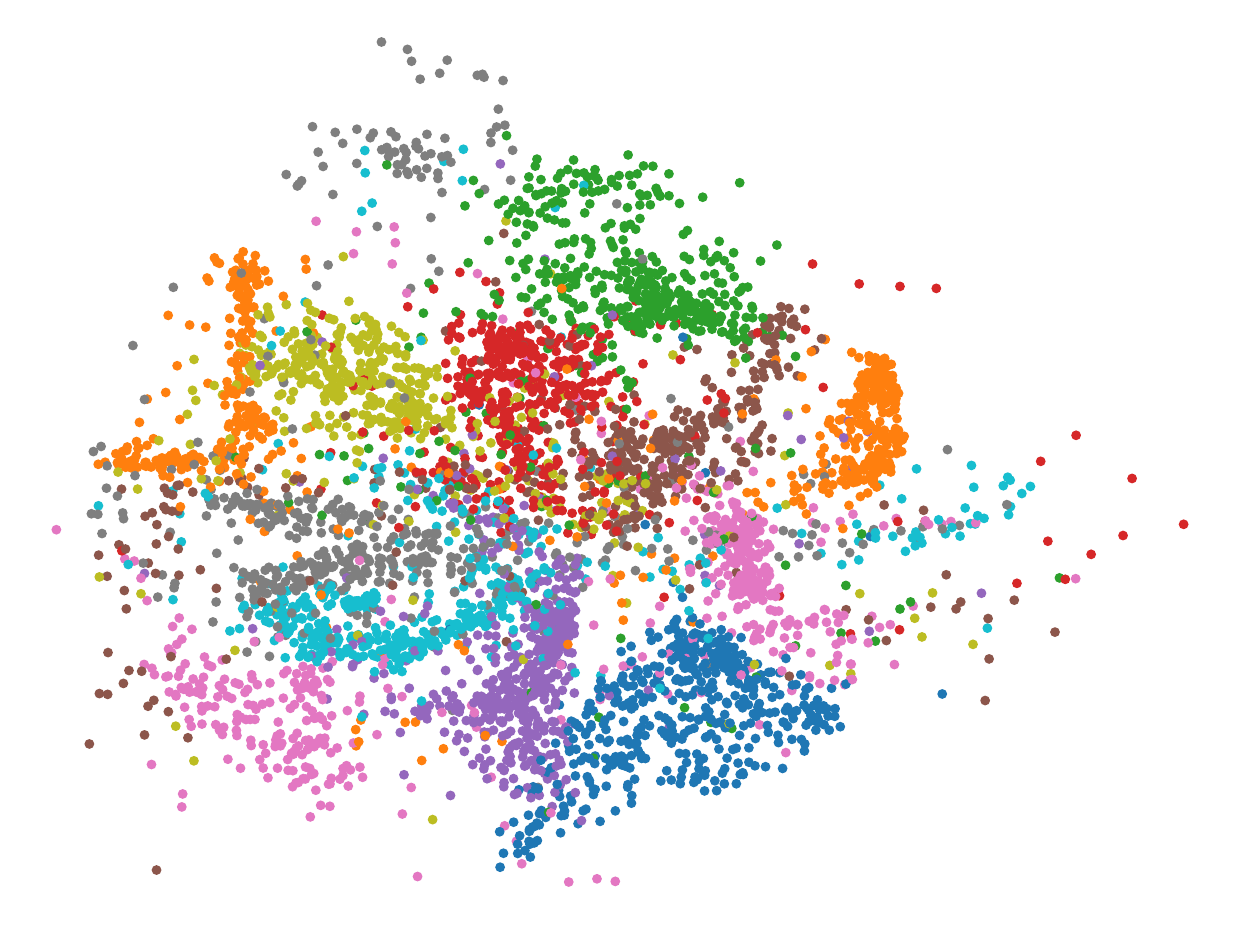}\\
        \multicolumn{1}{c}{15} & \includegraphics[width=\linewidth]{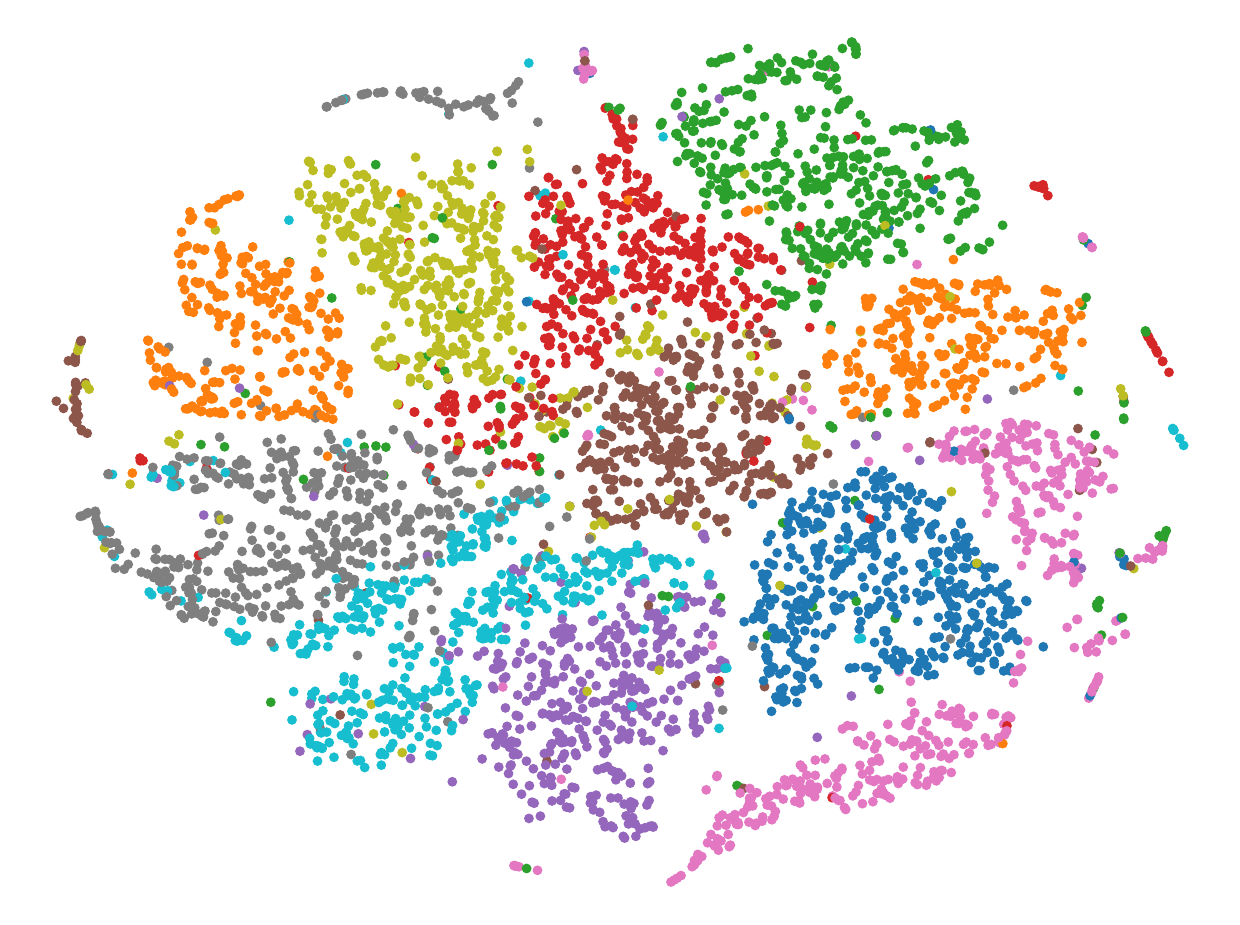} & \includegraphics[width=\linewidth]{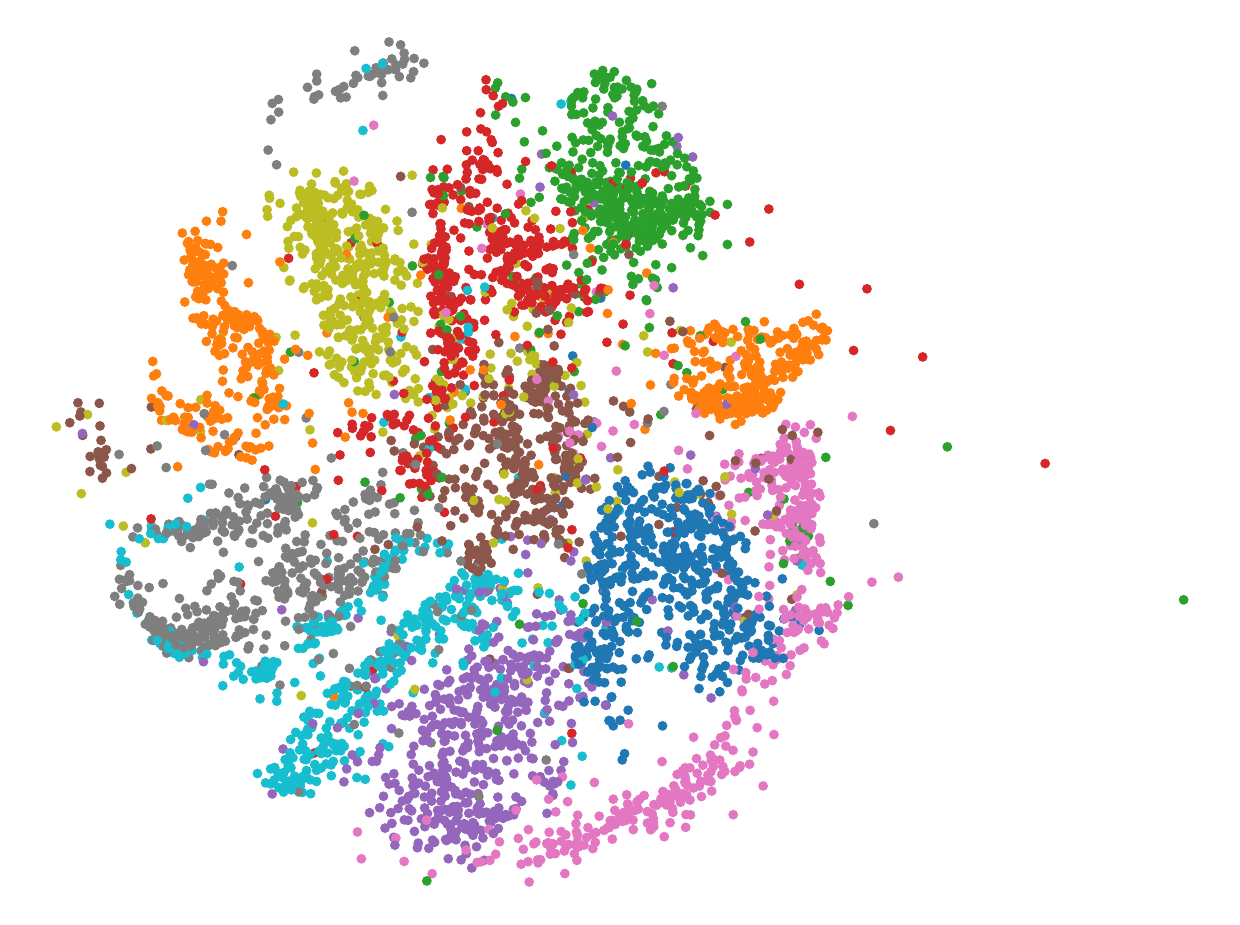} & \includegraphics[width=\linewidth]{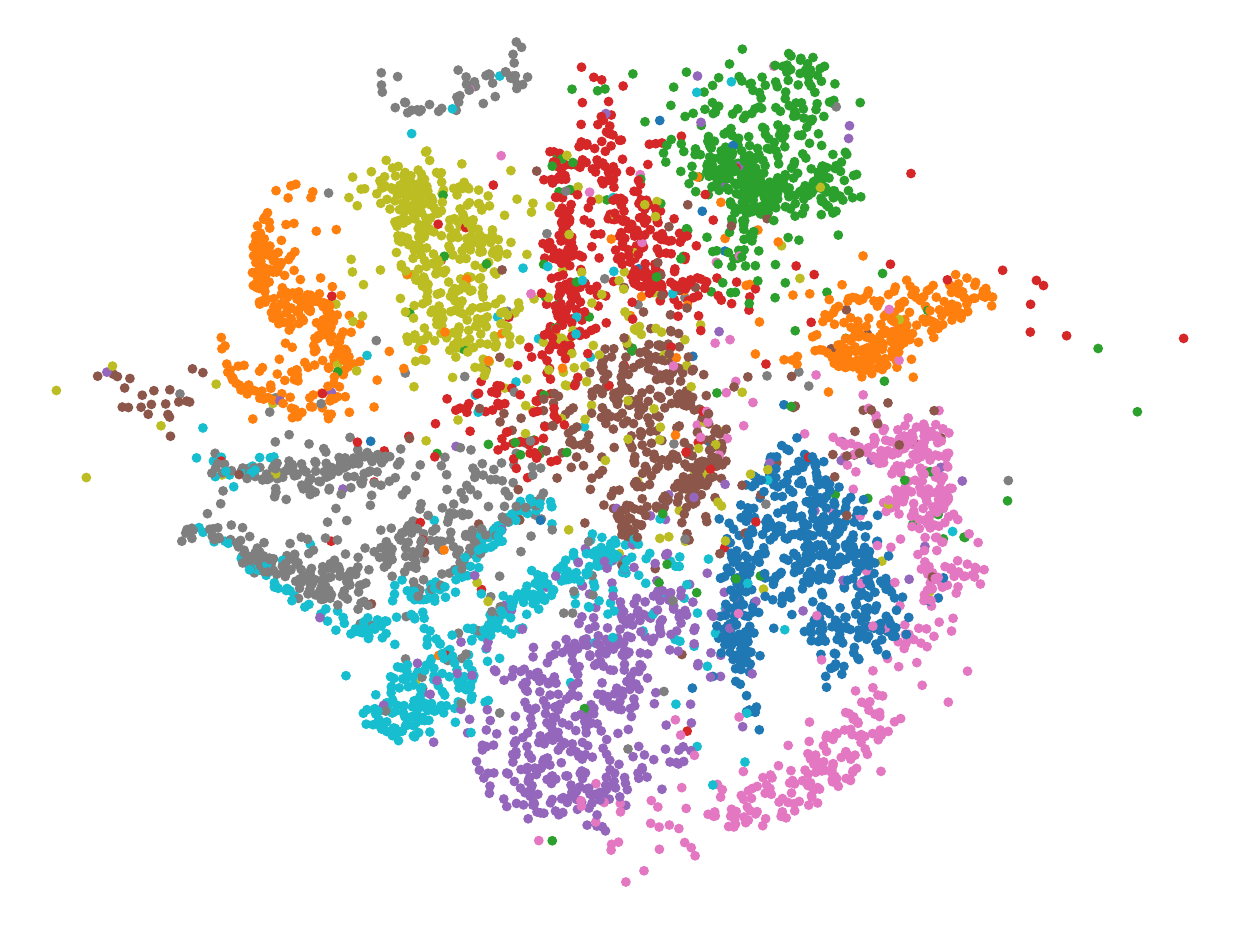} & \includegraphics[width=\linewidth]{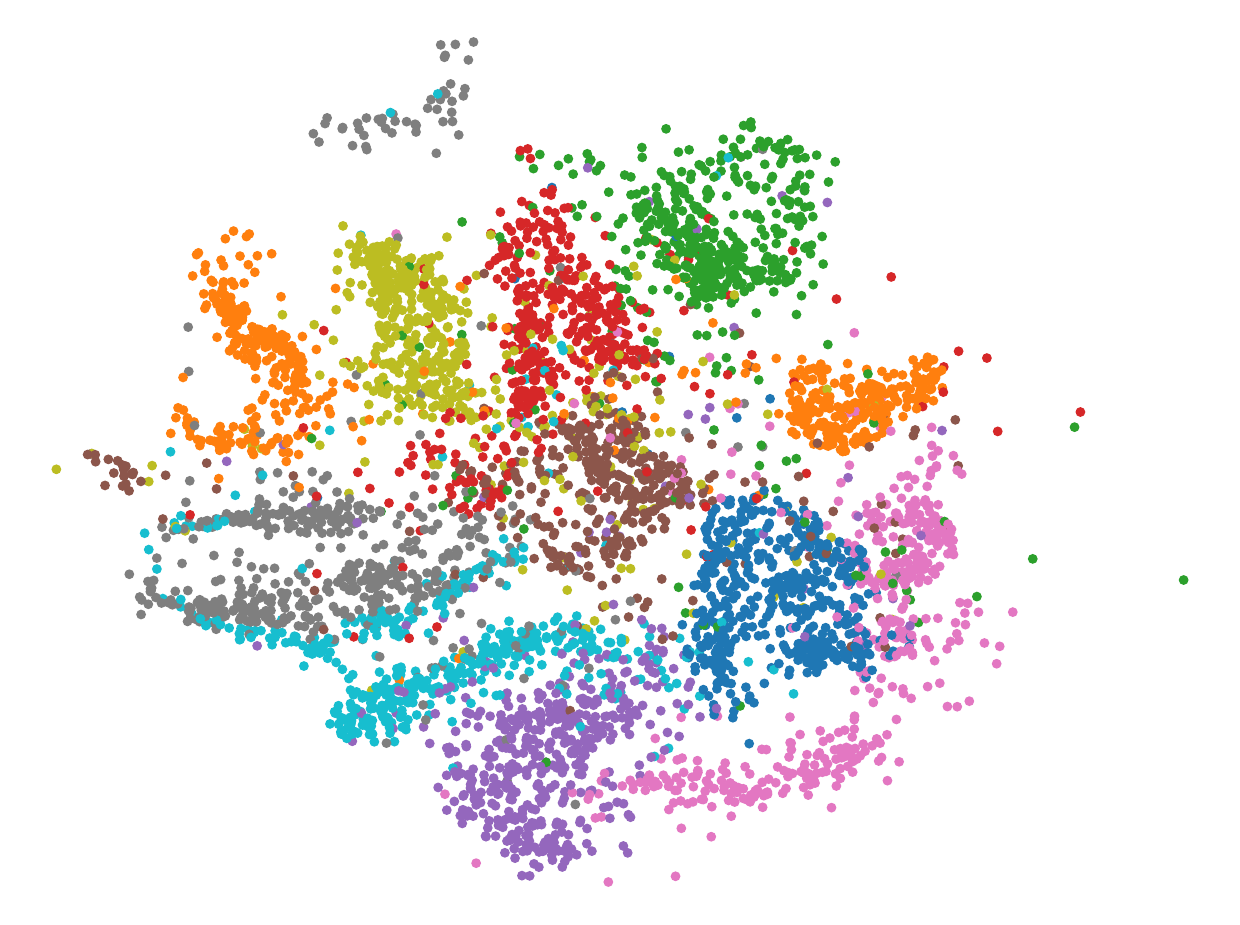}\\
        \multicolumn{1}{c}{25} & \includegraphics[width=\linewidth]{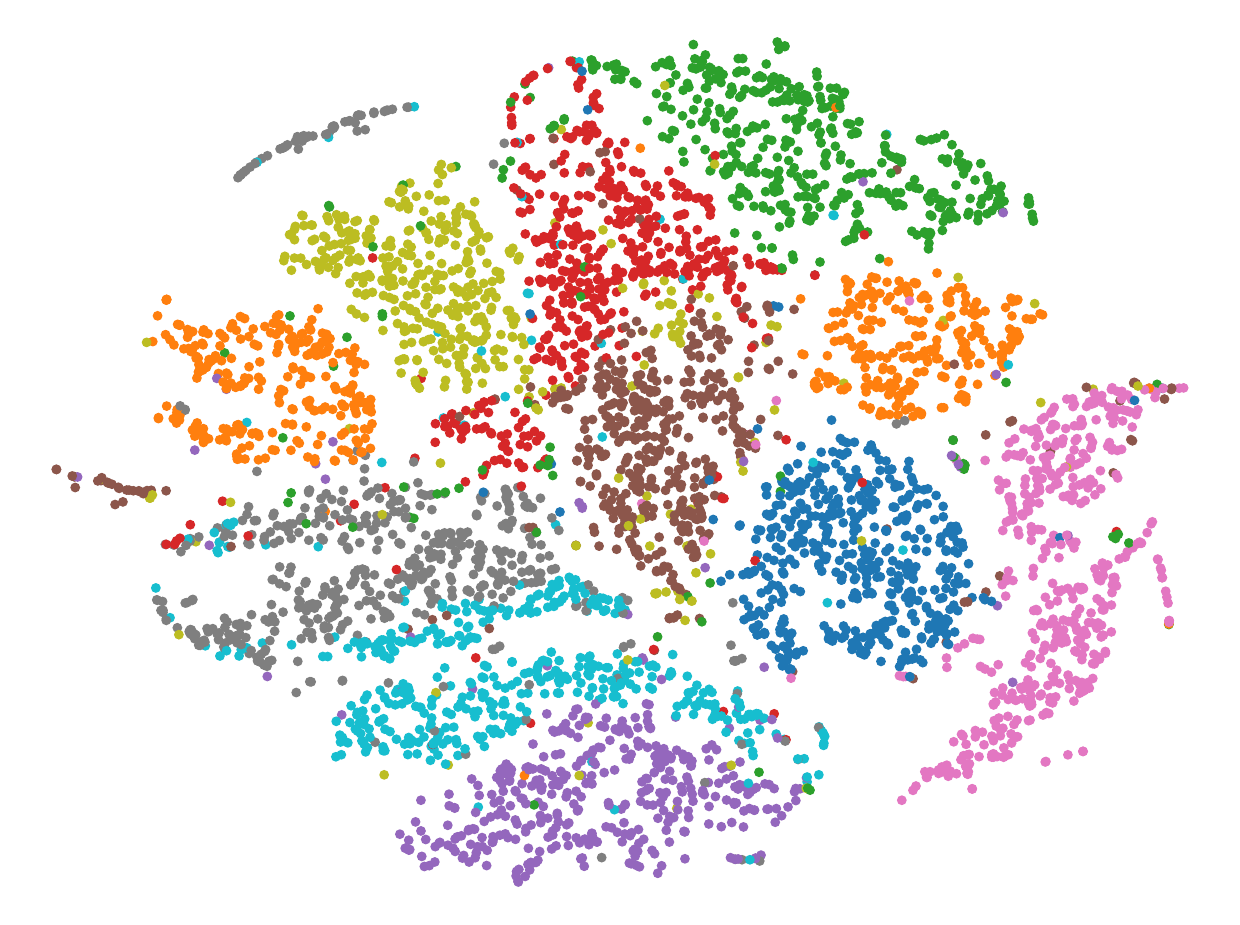} & \includegraphics[width=\linewidth]{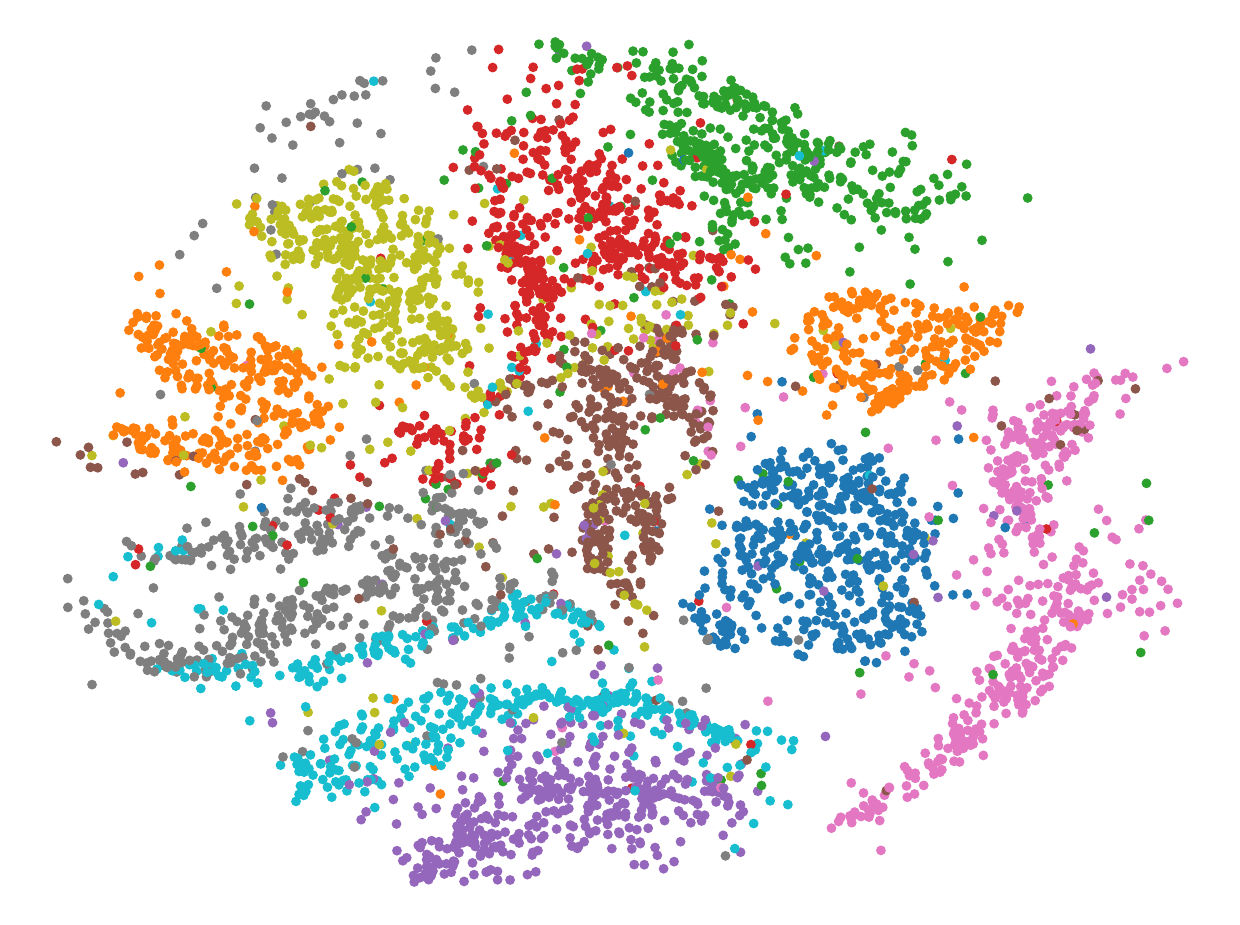} & \includegraphics[width=\linewidth]{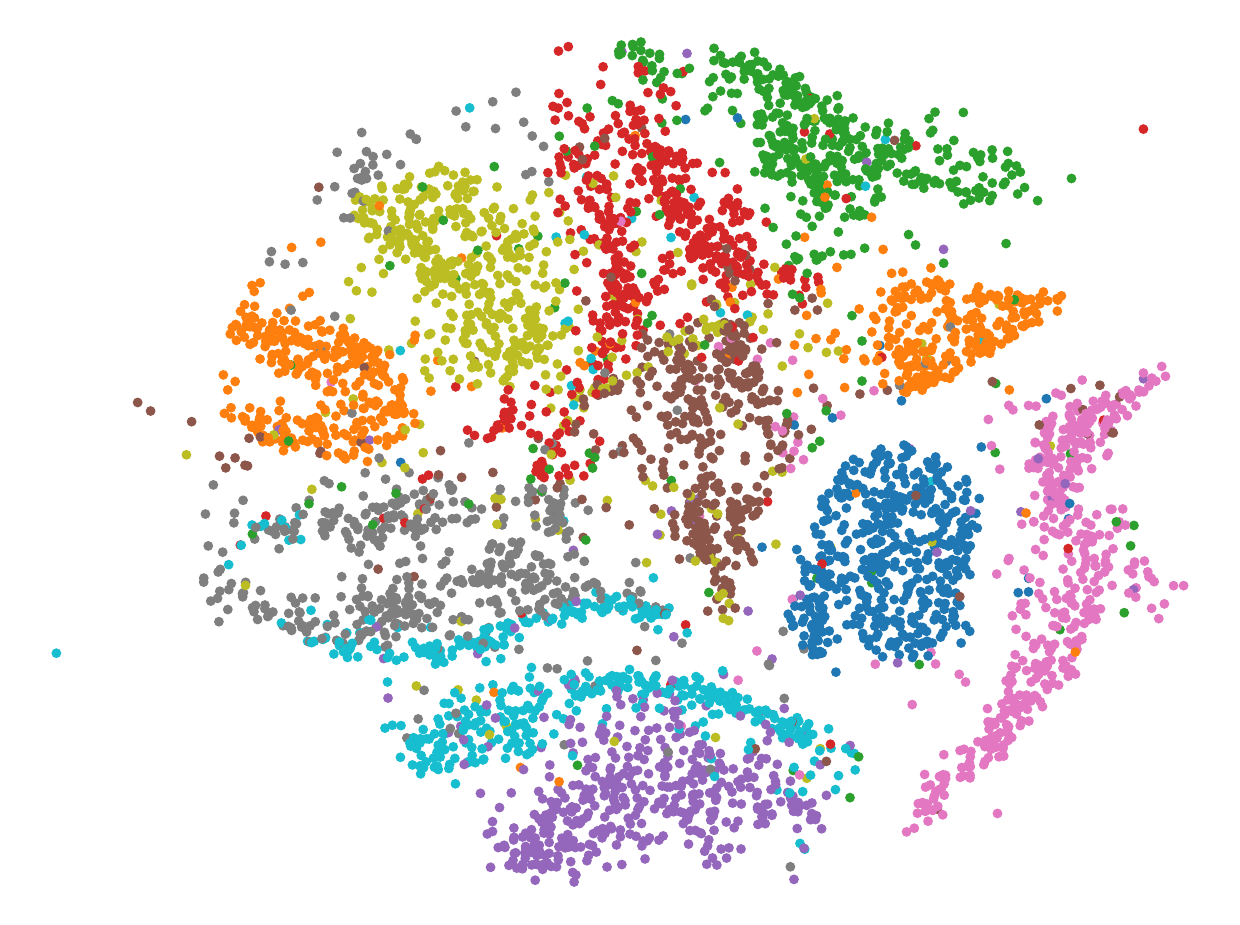} & \includegraphics[width=\linewidth]{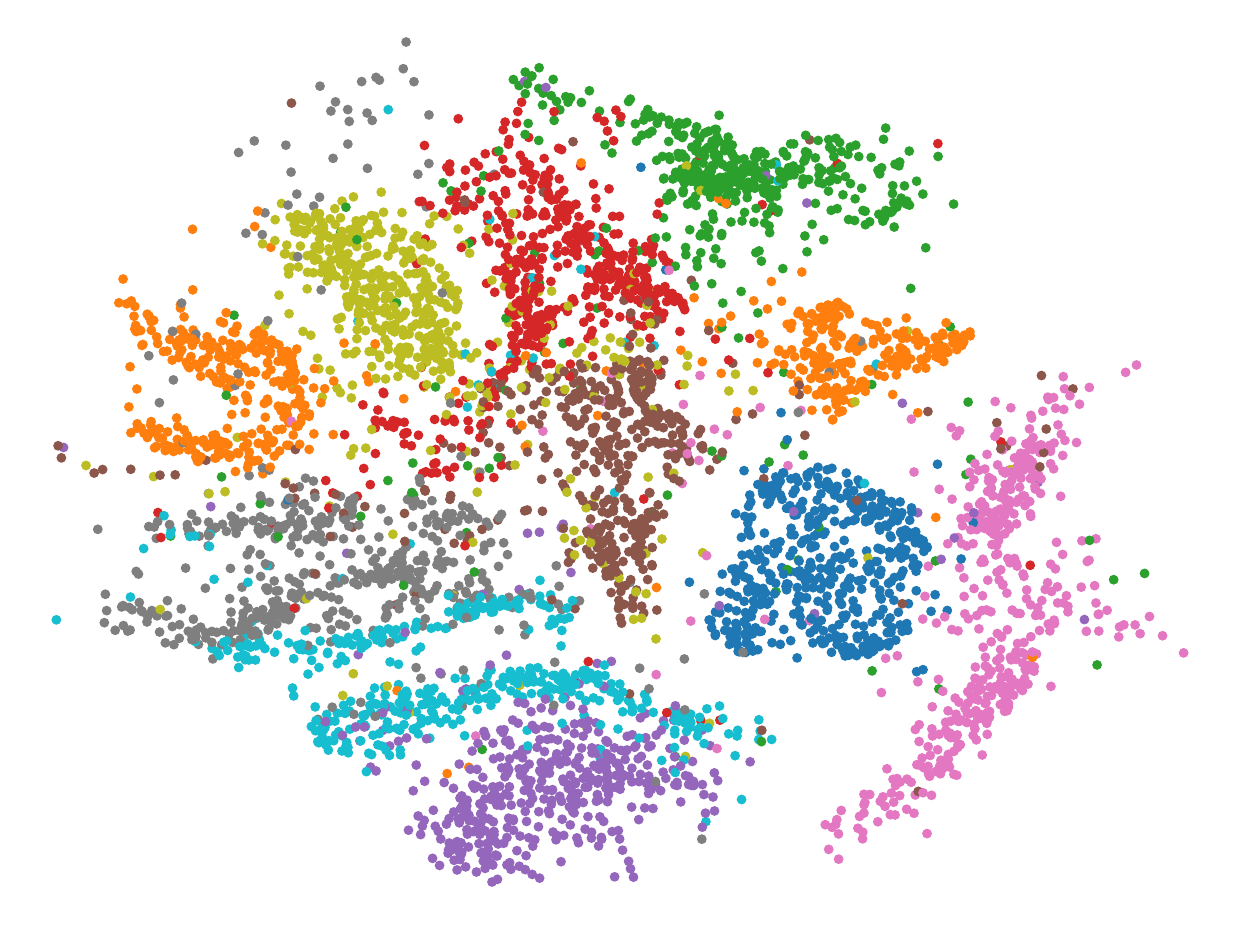}\\
        \bottomrule
    \end{tabularx}
    \caption{t-SNE and \sys projections for three different perplexity values. First column: MNIST projected with t-SNE projections at perplexities $p= \{5, 15, 25\}$. The following three columns: \sys results when trained with gap values of 2, 4, and 8.}
    \label{fig:tsne_application}
\end{table*}
\begin{table*}[htbp!]
    \centering
    \begin{tabularx}{\textwidth}{c X X X X}
      \toprule
        \multirow{2}{*}{k} & \multirow{2}{*}{UMAP} & \multicolumn{3}{c}{HyperNP}\\
        & & \multicolumn{1}{c}{Gap=2} & \multicolumn{1}{c}{Gap=4} & \multicolumn{1}{c}{Gap=8} \\
        \midrule
        \multicolumn{1}{c}{5} & \includegraphics[width=\linewidth]{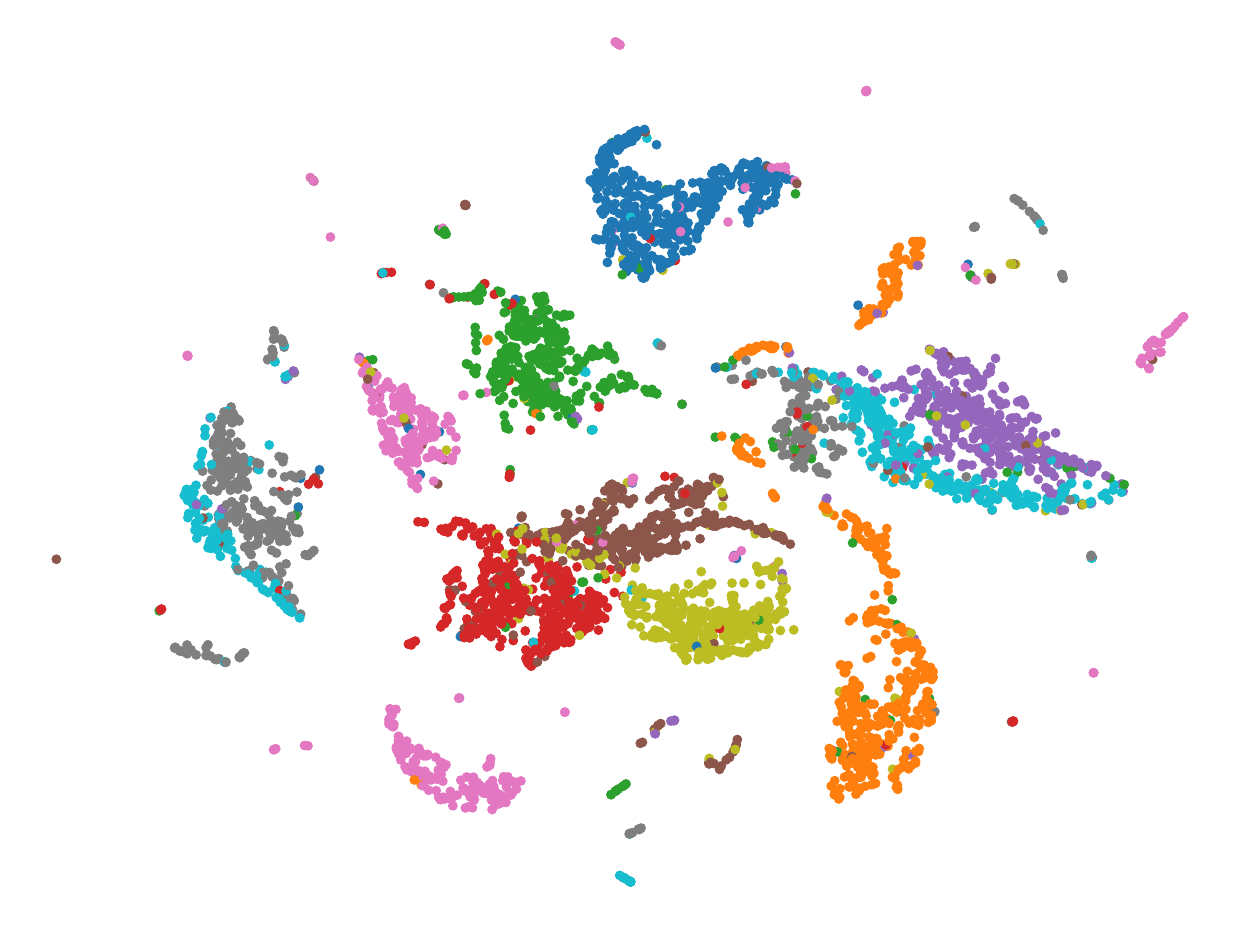} & \includegraphics[width=\linewidth]{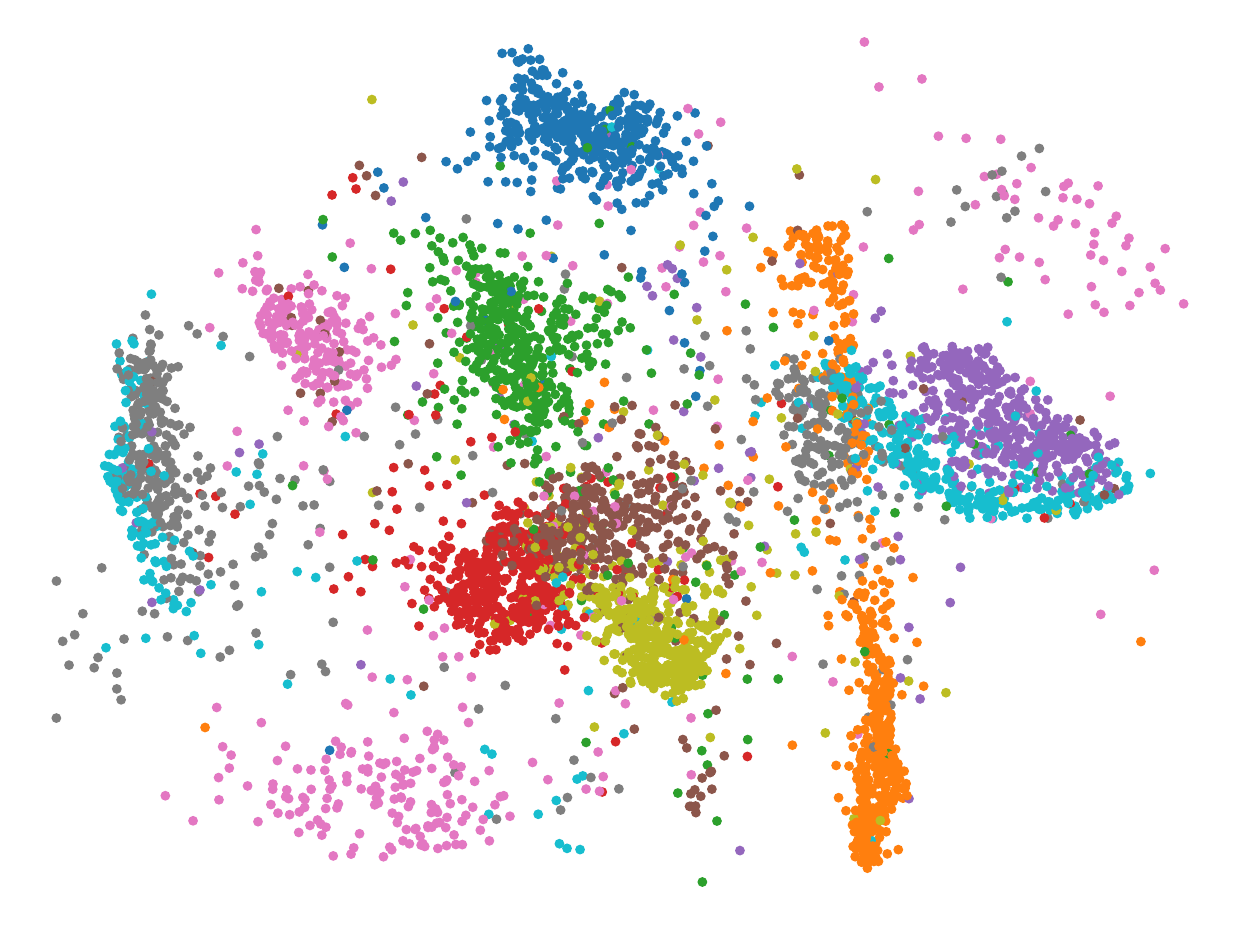} & \includegraphics[width=\linewidth]{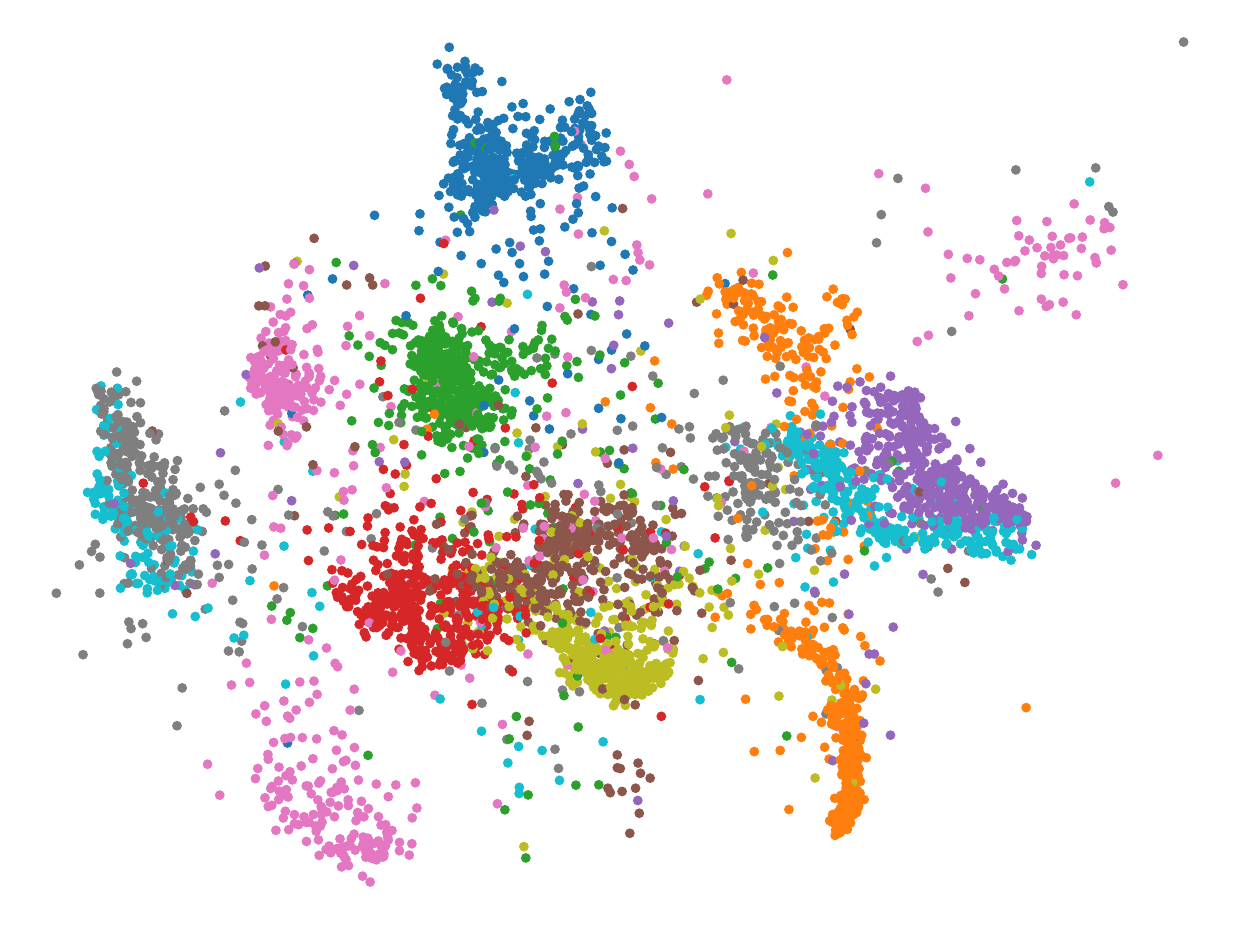} & \includegraphics[width=\linewidth]{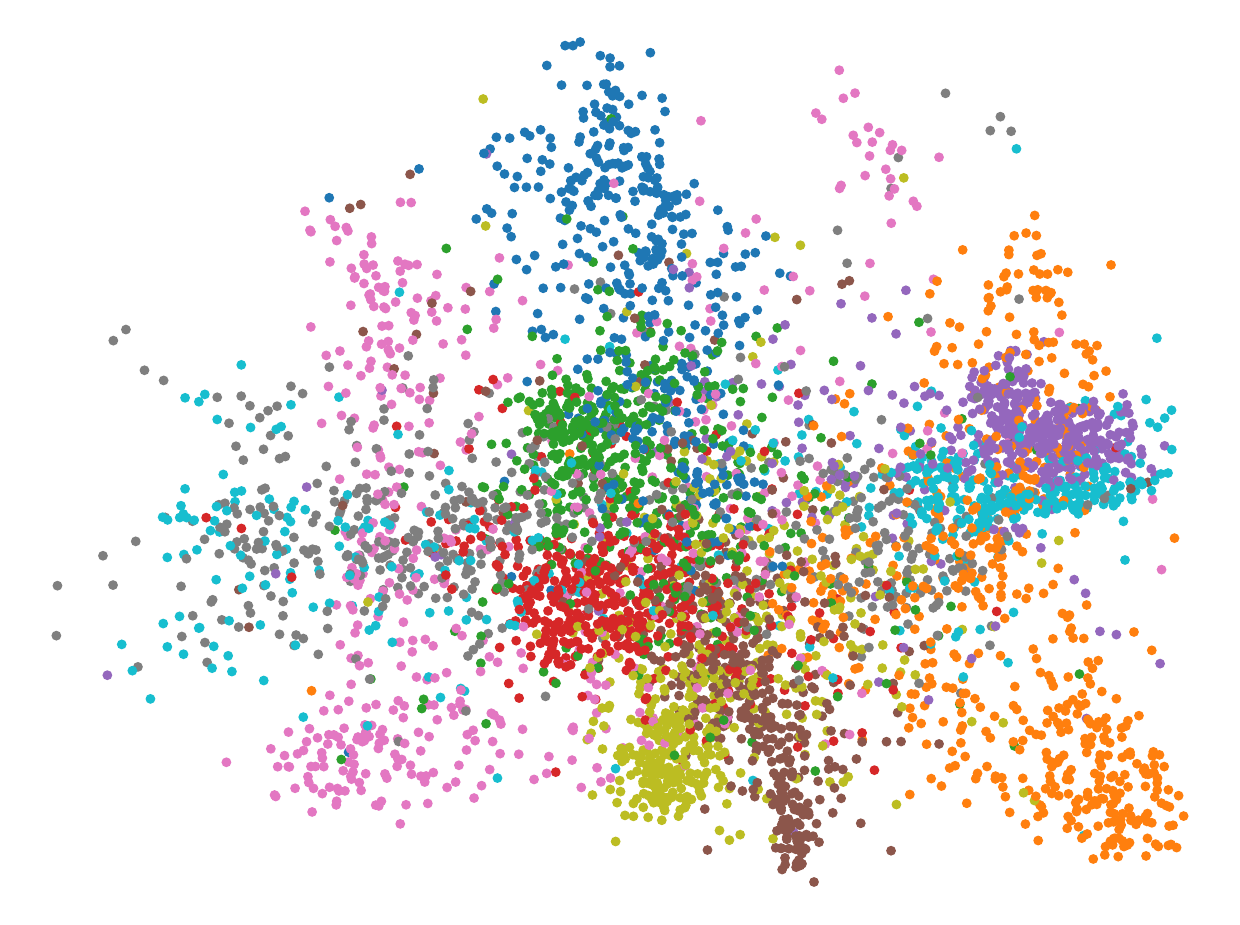}\\
        \multicolumn{1}{c}{15} & \includegraphics[width=\linewidth]{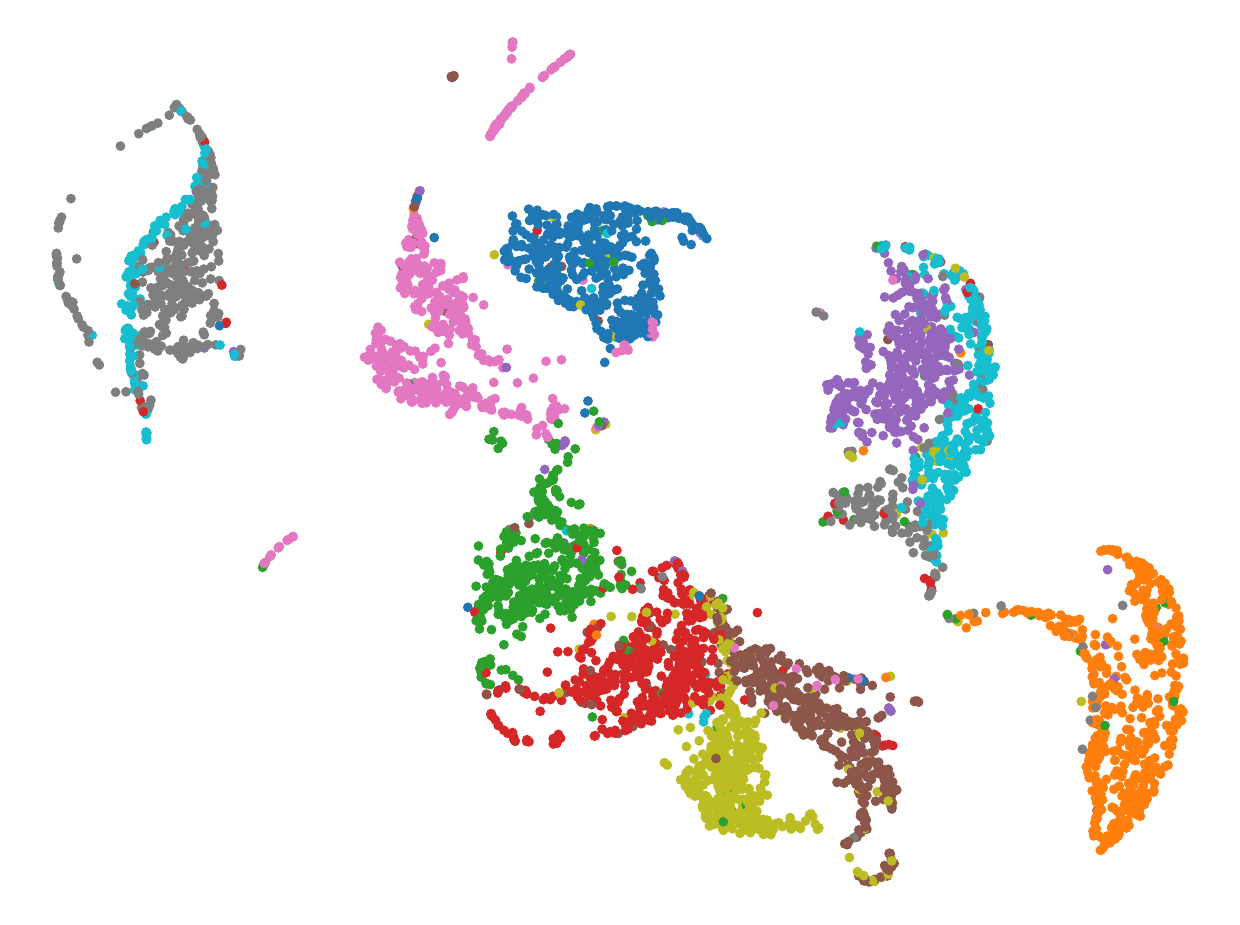} & \includegraphics[width=\linewidth]{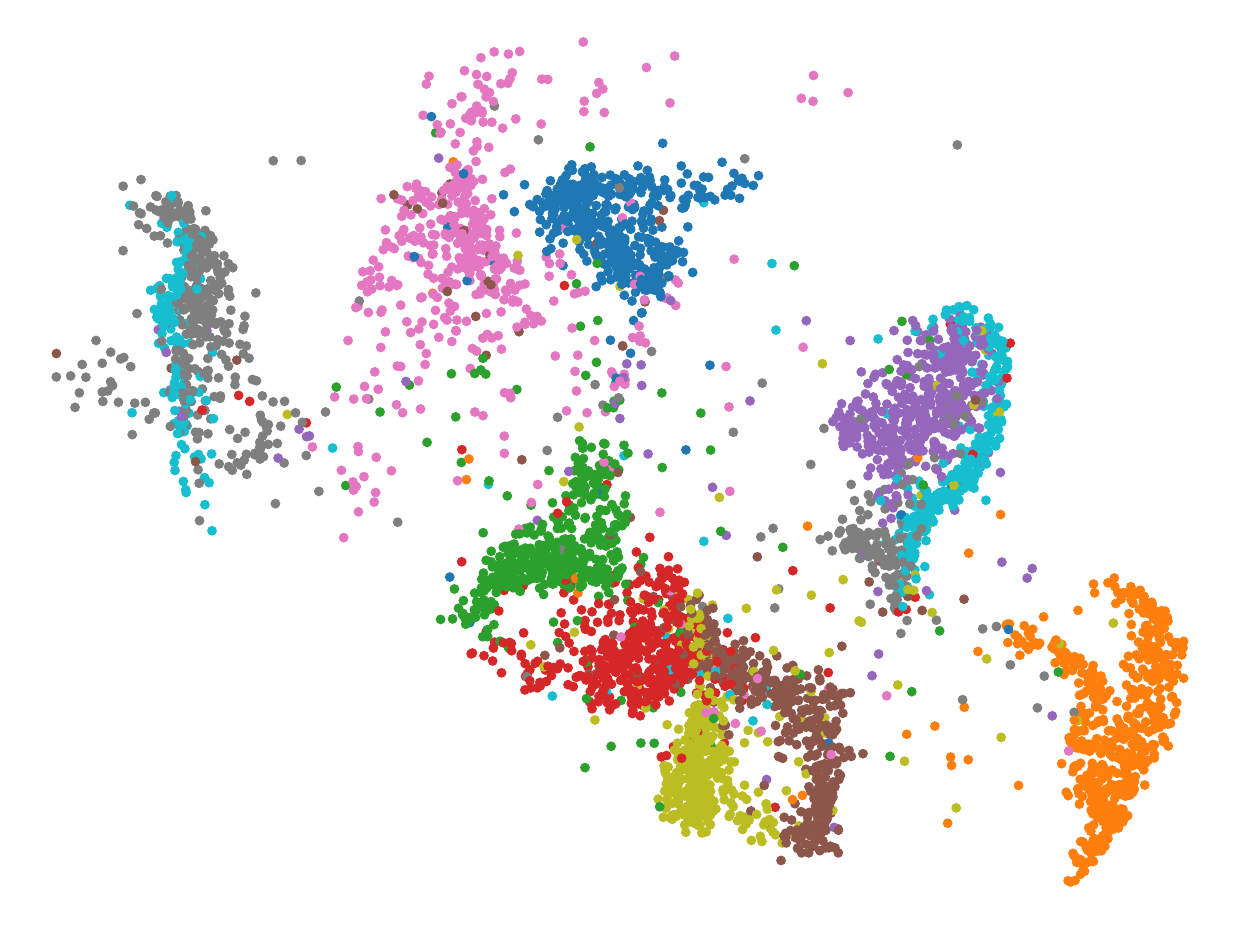} & \includegraphics[width=\linewidth]{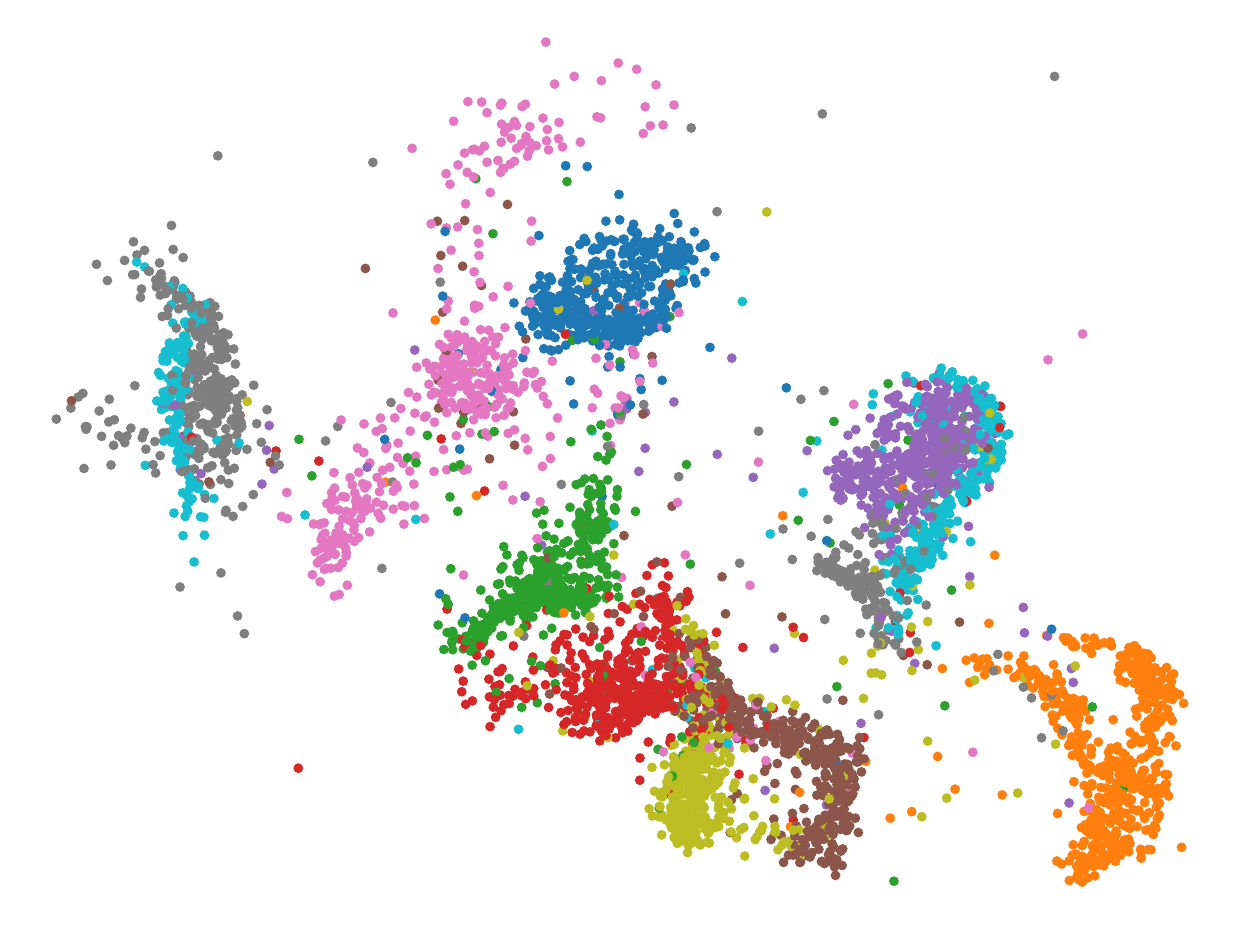} & \includegraphics[width=\linewidth]{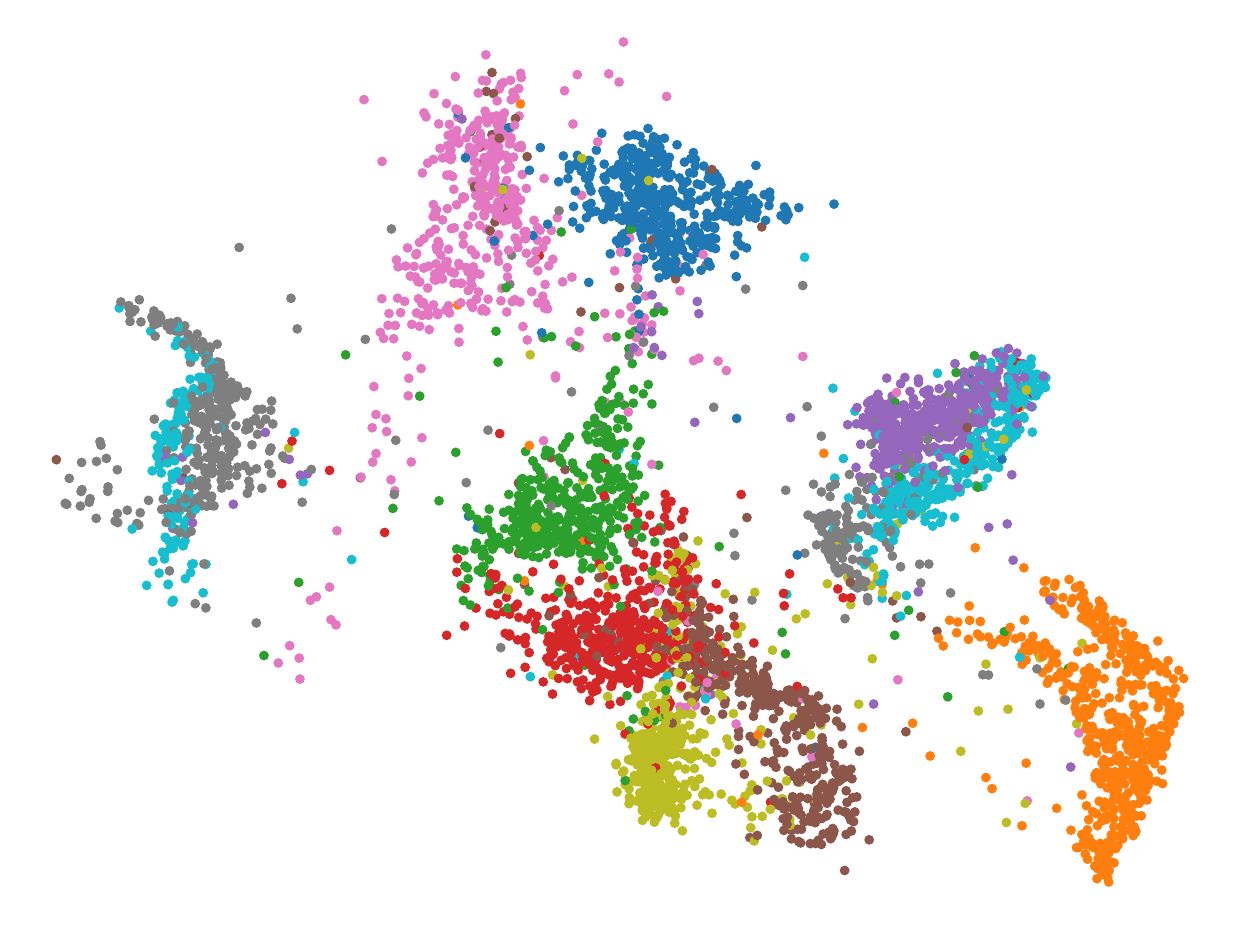}\\
        \multicolumn{1}{c}{25} & \includegraphics[width=\linewidth]{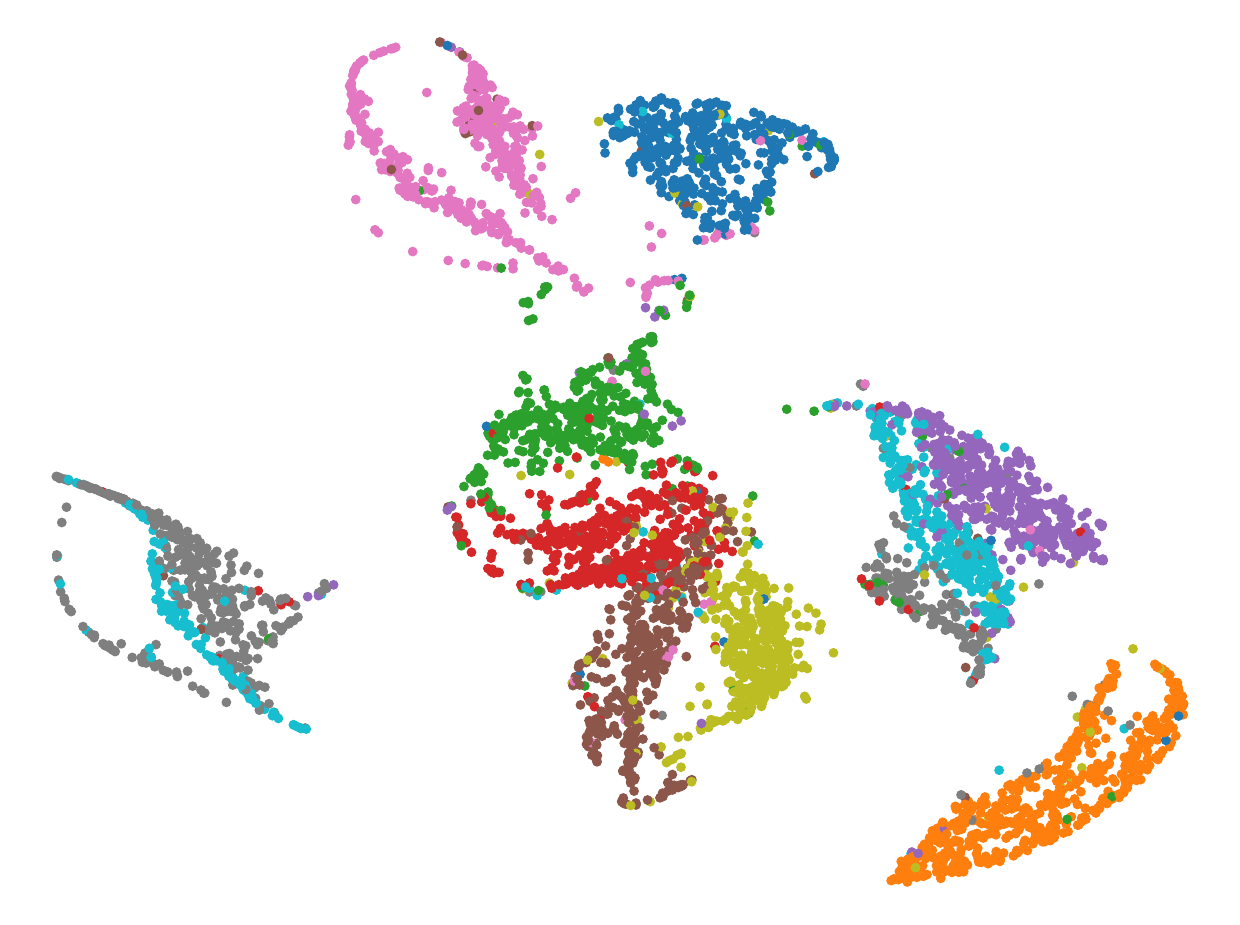} & \includegraphics[width=\linewidth]{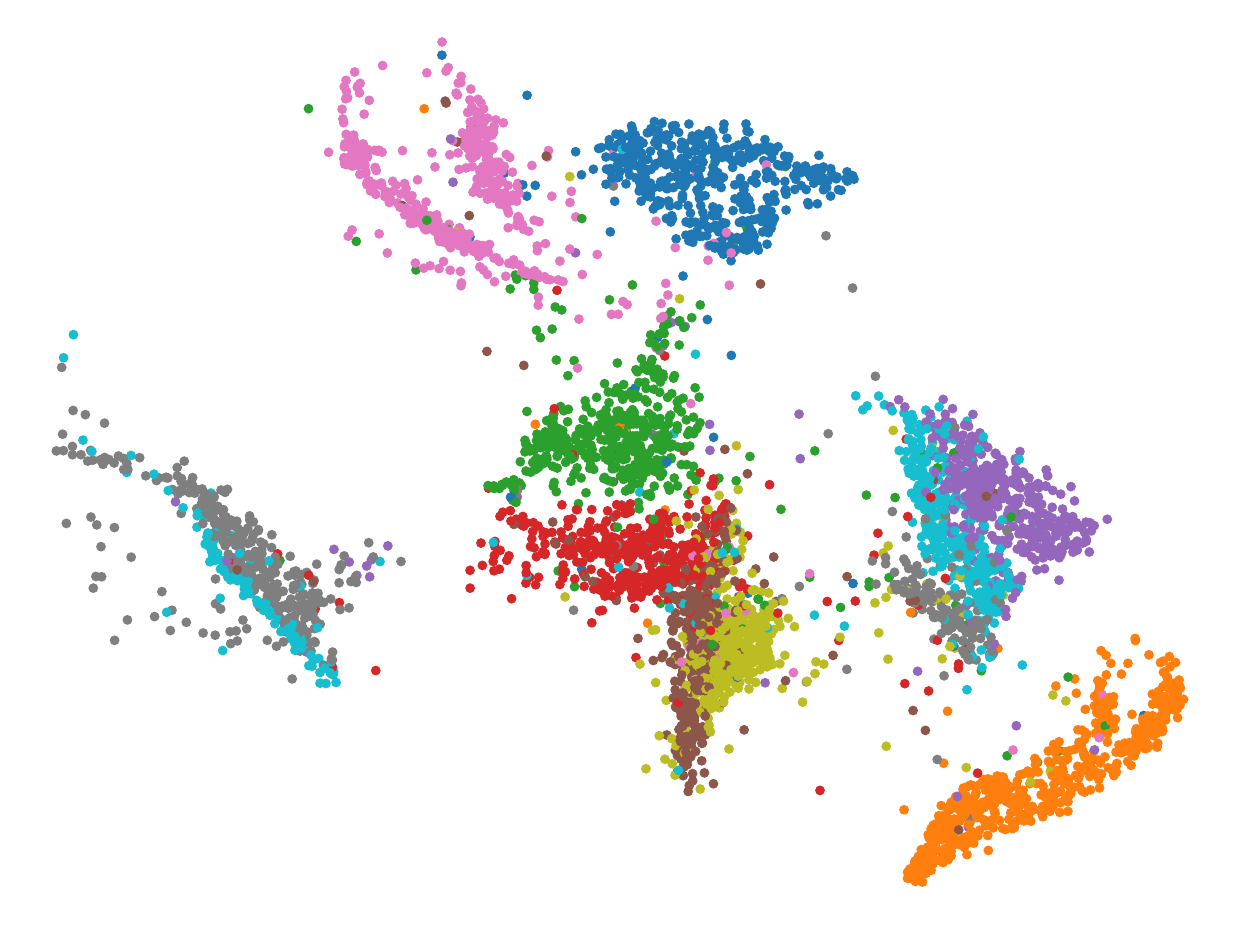} & \includegraphics[width=\linewidth]{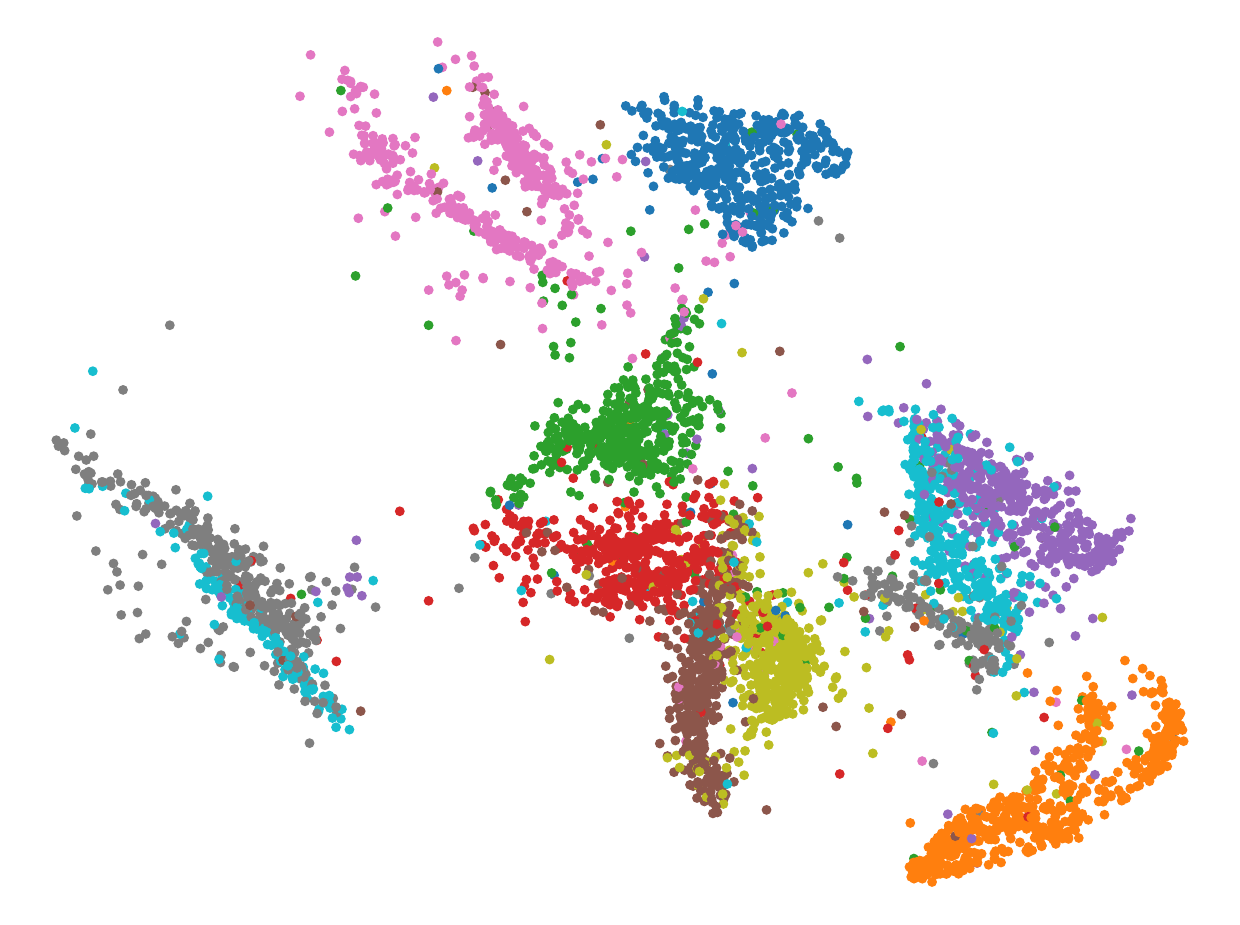} & \includegraphics[width=\linewidth]{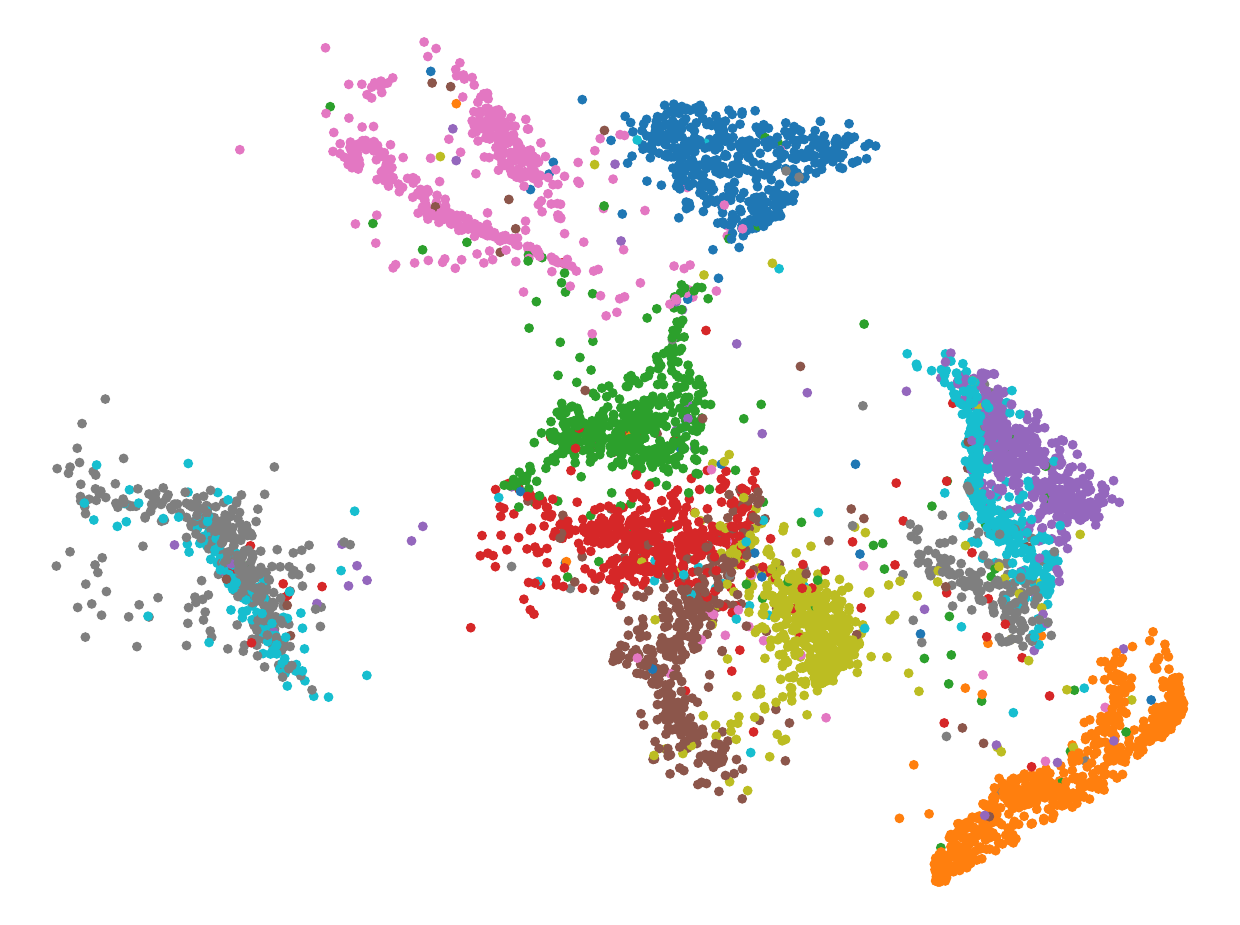}\\
        \bottomrule
    \end{tabularx}
    \caption{UMAP and \sys projections for three different perplexity values. First column: MNIST projected with UMAP projections at nearest neighbor values of $k= \{5, 15, 25\}$. The following three columns: \sys results when trained with gap values of 2, 4, and 8.}
    \label{fig:umap_application}
\end{table*}
%
%

%
Isomap is also a technique that projects data points by first learning the manifold structure in high dimensions via kNN graphs~\cite{balasubramanian:2002:isomap}.
As with UMAP, the most often changed hyperparameter of Isomap is the number of nearest neighbors to consider while building the kNN graph, $k$.  
However, unlike UMAP's notion of probabilistic edge weights (See Section~\ref{sec:applications:umap}), Isomap uses distances as edge weights.
These weights enable methods like Djikstra's Algorithm\cite{dijkstra:1959:dijkstra_sssp} to approximate a geodesic distance along the induced manifold for each pair of points.
After pair-wise geodesic distances are approximated, classical MDS methods are employed to construct the final low-dimensional embedding.
As both UMAP and Isomap make use of manifold learning and represent the manifold structure through kNN graphs, both methods rely on the value of the hyperparameter $k$ to set the importance of global and local feature importance.
Silva et al. provide a full discussion of the importance of this hyperparameter with respect to feature size for the class of projection techniques using manifold learning approaches~\cite{silva:2002:globalvlocal}. 
As the underlying functionality of our method does not depend on any computational or theoretical framework unique to a given projection method, new methods are easily adapted to \sys's model. 
As shown in Figure~\ref{fig:isomap_train_test_both}, \sys can approximate Isomap with high accuracy.
These results are comparable to the same experiment performed using UMAP (See Fig.~\ref{fig:tsne_train_test_both}).
Additional attention must be paid to the value of $k$ with respect to gap size in order to ensure projection quality remains high across all projection configurations for both UMAP and Isomap.
%
%
%
%
\section{Evaluation}
\label{sec:validation}
We evaluate HyperNP's performance through several experiments.
We use three publicly available high-dimensional datasets that are reasonably large (thousands of samples) and have non-trivial data structure.

\vspace{4pt}
\noindent\textbf{MNIST}\,\cite{lecun:2010:mnist}: 70K images of handwritten digits from 0 to 9, rendered as 28x28-pixel grayscale images, flattened to 784-element vectors;

\vspace{4pt}
\noindent\textbf{FashionMNIST}\,\cite{xiao:2017:fashion_mnist}: 70K images of 10 clothing piece types, rendered and flattened as for MNIST;

\vspace{4pt}
\noindent\textbf{GloVe}\,\cite{pennington:2014:glove}: 70K GloVe (Global Vectors for Word Representation) vectors sampled from the 400k 300-dimensional vectors trained on Wikipedia 2014 + Gigaword 5.
%

\subsection{Performance: Quantitative Metrics}
\label{sec:validation:quality}
We measure the quality of \sys by the following metrics commonly used in the projection literature\,\cite{espadoto:2019:quantiative_dr_survey}.

\vspace{4pt}
\noindent\textbf{Trustworthiness}\,\cite{venna_visualizing_2006}: Measures the fraction of points in $D$ that are also close in $P(D)$.
It tells us how much we can trust the local patterns in a projection (\emph{e.g.} clusters) to represent actual data patterns.
In the definition (Tab.~\ref{tab:metrics}), $U^{(K)}_i$ is the set of points that are among the $K$ nearest neighbors of point $i$ in 2D but not among the $K$ nearest neighbors of point $i$ in $\mathbb{R}^n$; and $r(i, j)$ is the rank of point $j$ in the ordered-set of nearest neighbors of $i$ in 2D.
We choose $K=7$ in our experiments, in line with\,\cite{maaten:2009:dim_reduction_survey,martins_explaining_2015,espadoto:2019:quantiative_dr_survey}.

\vspace{4pt}
\noindent\textbf{Continuity}\,\cite{venna_visualizing_2006}: Measures the fraction of close points in $P(D)$ that are also close in $D$.
In the definition (See Table~\ref{tab:metrics}), $V^{(K)}_i$ is the set of points that are among the $K$ nearest neighbors of point $i$ in $\mathbb{R}^n$ but not among the $K$ nearest neighbors in 2D; and $\hat{r}(i, j)$ is the rank of point $j$ in the ordered set of nearest neighbors of $i$ in $\mathbb{R}^n$. 
Similar to trustworthiness we choose $K=7$ following convention.
\begin{table}[H]
\vspace{-0.15cm}
    \centering
    \scriptsize
    \begin{tabular}{ | l | c |  c | c |}
    \hline
\textbf{Metric}    & \textbf{Definition} & \textbf{Range} \\ \hline
Trustworthiness & $ 1 - \frac{2}{NK(2n-3K-1)}\sum_{i=1}^{N}{\sum_{j \in U^{(K)}_i}{(r(i,j) - K)}}$ & $[0,\mathbf{1}]$ \\ \hline
Continuity & $ 1 - \frac{2}{NK(2n-3K-1)}\sum_{i=1}^{N}{\sum_{j \in V^{(K)}_i}{(\hat{r}(i,j) - K)}}$  & $[0,\mathbf{1}]$ \\ \hline
\end{tabular}
\vspace{-0.15cm}
\caption{Projection quality metrics used in our evaluation with optimal values in bold in the Range column.}
\vspace{-0.15cm}
\label{tab:metrics}
\end{table}
\begin{figure*}[!ht]
     \centering
     \begin{subfigure}[b]{0.49\linewidth}
         \centering
         \includegraphics[width=\linewidth]{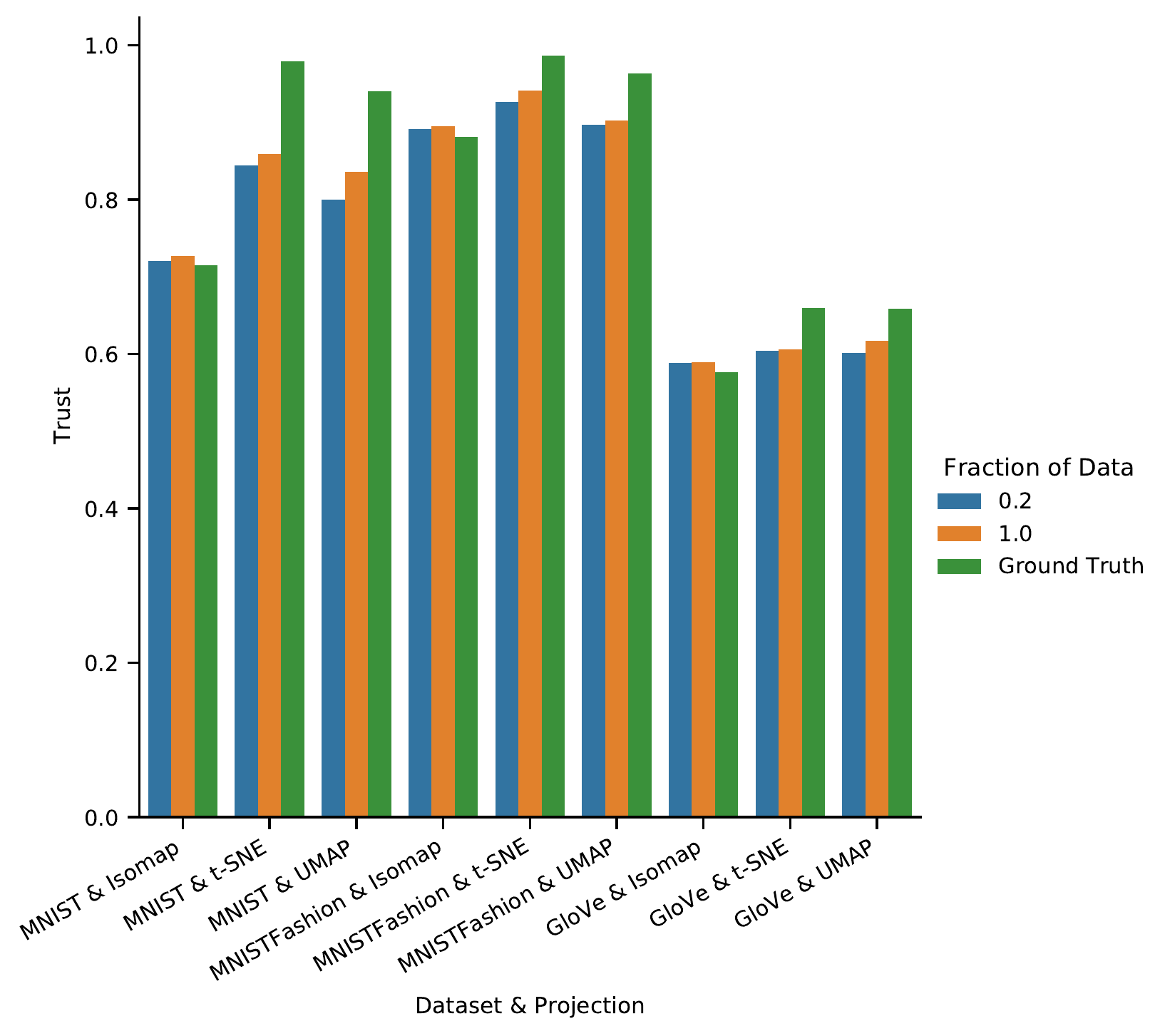}
         \caption{Trust}
         \label{fig:aggregated_trust_and_continuity:trust}
     \end{subfigure}
     \begin{subfigure}[b]{0.49\linewidth}
         \centering
         \includegraphics[width=\linewidth]{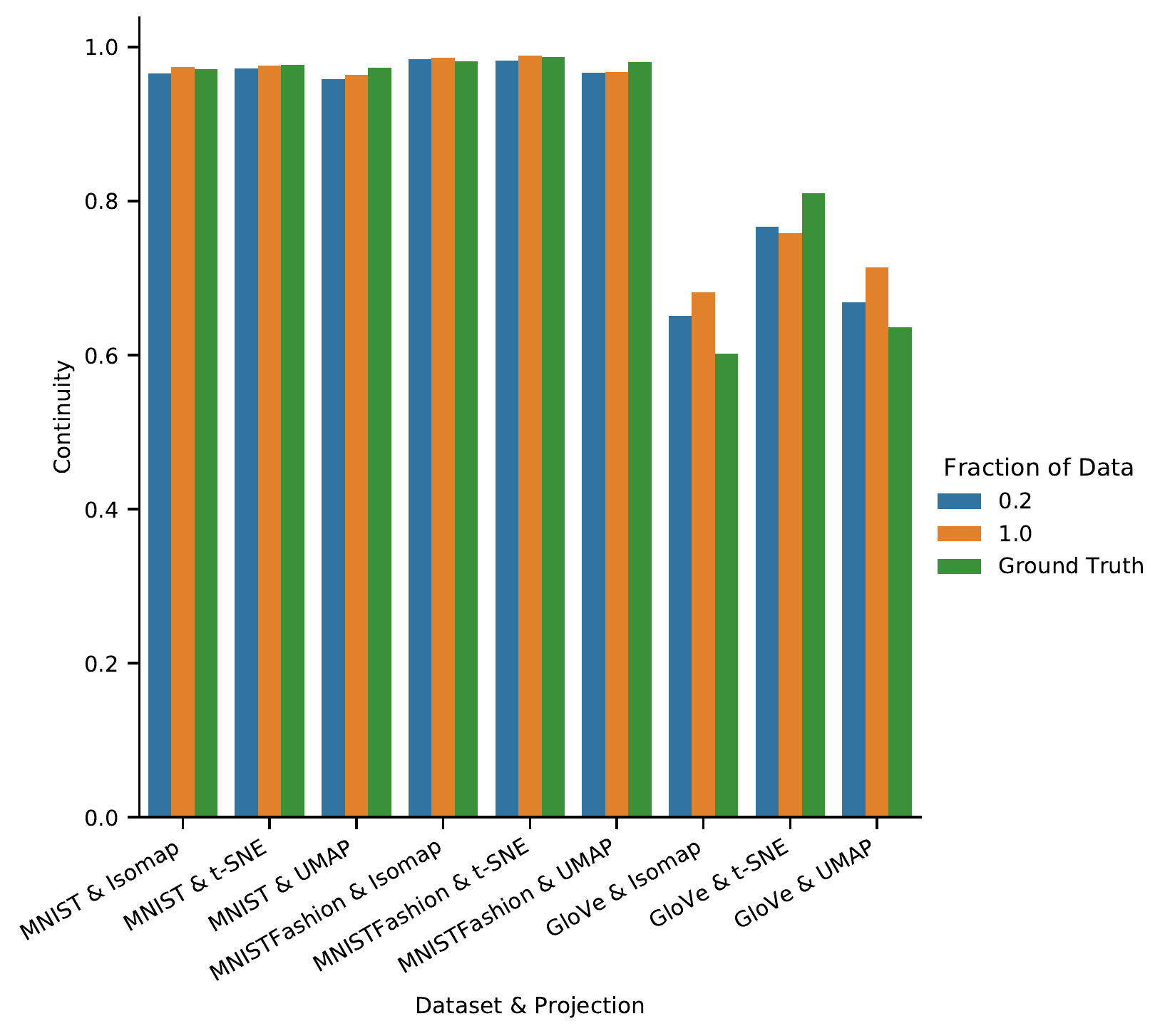}
         \caption{Continuity}
         \label{fig:aggregated_trust_and_continuity:continuity}
     \end{subfigure}
    \caption{In this figure we present the Trust and Continuity scores associated with the ground truth projections as well as \sys trained used a gap size of $16$, and data fractions of $.2$ and $1.0$. We have averaged the scores over all of the hyperparameter values from $2$ to $50$.}
    \label{fig:aggregated_trust_and_continuity}
\end{figure*}
We next explore different hyperparameter settings to determine their effect on \sys's quality, namely the gap size and the data fraction used for training  (see Sec.~\ref{sec:method}).
We first examine the metrics in aggregate, by taking their averages across hyperparamter values.
For each dataset and projection combination, we fix the gap size and training fraction and compute the mean trust and continuity values across either perplexity or $k$, from two to fifty.
We present these results in Figure~\ref{fig:aggregated_trust_and_continuity}.
Notice that across all dataset and projection combinations within the figure, our method produces trust and continuity scores within $0.14$ of the ground truth projection method.
Increasing the data fraction to 1.0 from 0.2, a 5-fold increase, shows only a minimal improvement at the cost of additional training time.
These results inspire confidence; they show that \sys is able to infer projection locations of out-of-sample data, generalizing from a small subset of the entire dataset.
There is a difference in performance across both datasets and projection methods.
The ground truth projections of GloVe score particularly lower in our metrics than the MNIST and FashionMNIST datasets.
While \sys faithfully reproduces these mappings, the resulting approximations will also suffer from the poor quality of the ground truth projections.
Similarly, Isomap performs worse than UMAP and t-SNE on the three datasets when measuring trust, but performs comparably when measuring continuity.
Figure~\ref{fig:aggregated_trust_and_continuity} shows how closely \sys replicates projections (trained using a gap size of 16).
In Figure~\ref{fig:digits_umap_trust_and_continuity}, we plot trust and continuity for UMAP, varying the value of the hyperparameter $k$, and data fraction and gap size used to train.
In this experiment, \sys performs similarly to ground truth for continuity, and with a trust metric approximately 80\% that of ground truth.
\begin{figure*}[hb!]
     \vspace{-0.15cm}
     \centering
     \begin{subfigure}[b]{0.44\linewidth}
         \centering
         \includegraphics[width=\linewidth]{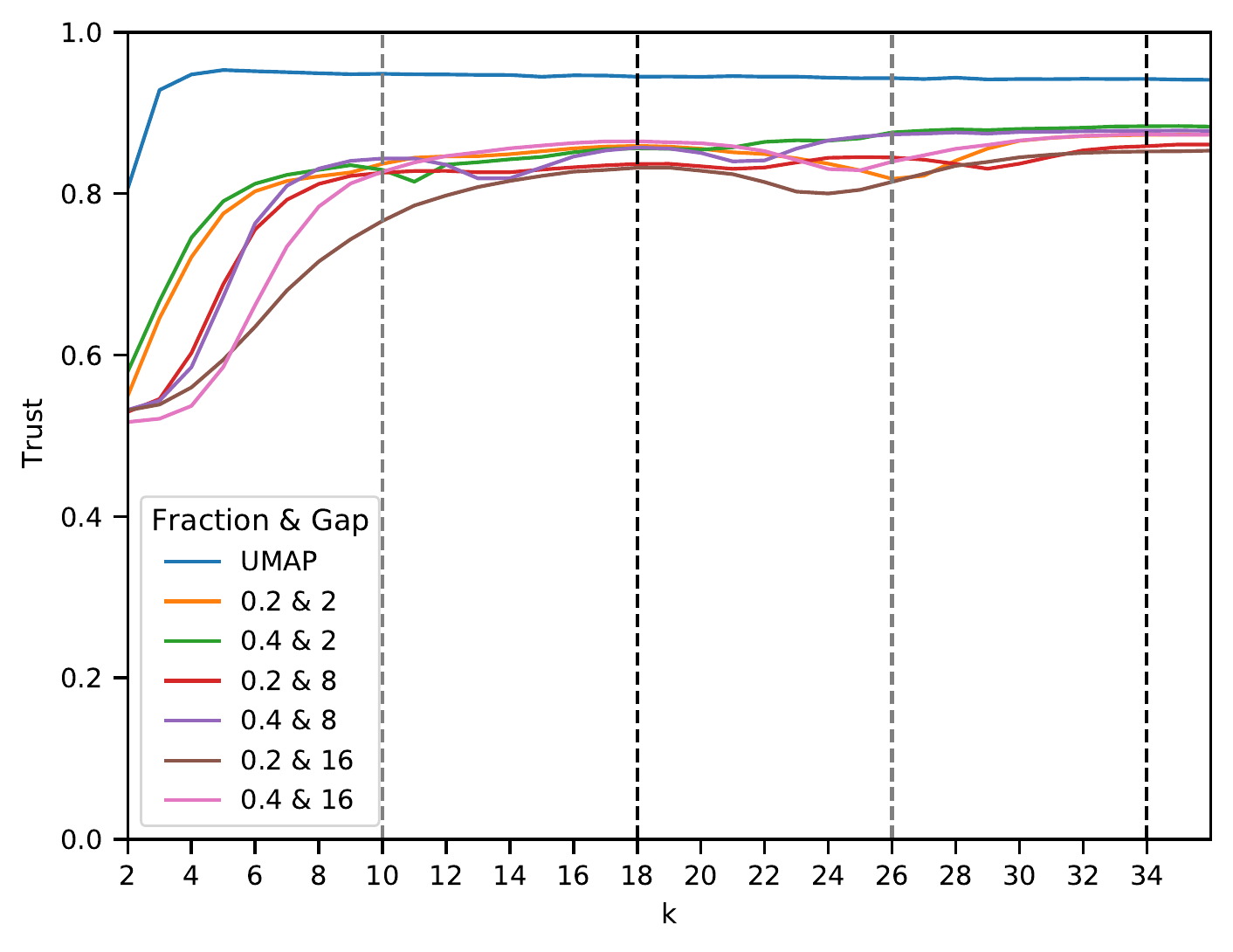}
         \caption{Trust}
         \label{fig:digits_umap_trust_and_continuity:trust}
     \end{subfigure}
     \begin{subfigure}[b]{0.44\linewidth}
         \centering
         \includegraphics[width=\linewidth]{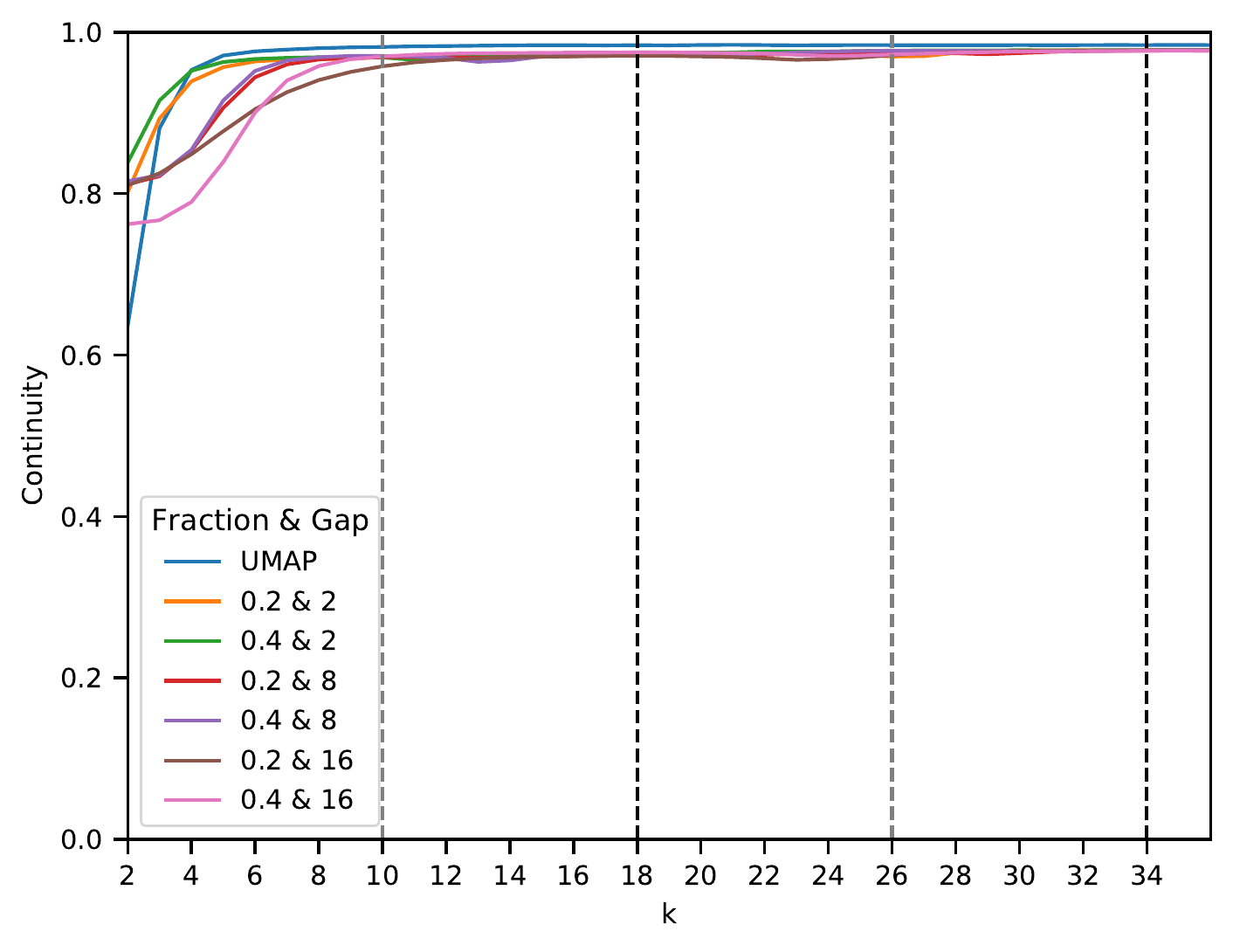}
         \caption{Continuity}
         \label{fig:digits_umap_trust_and_continuity:continuity}
     \end{subfigure}
    \caption{Continuity and trustworthiness for \sys trained with UMAP, on the MNIST dataset, for different hyperparameter values. Light gray vertical lines show parameter values used for training with gap $g=8$; dark gray ones show parameter values used for training with $g=16$.}
    \label{fig:digits_umap_trust_and_continuity}
     \vspace{-0.15cm}
\end{figure*}
\subsection{Performance: Speed}
\label{sec:validation:speed}
%



\begin{figure}[htbp!]
    \centering
    \includegraphics[width=\linewidth]{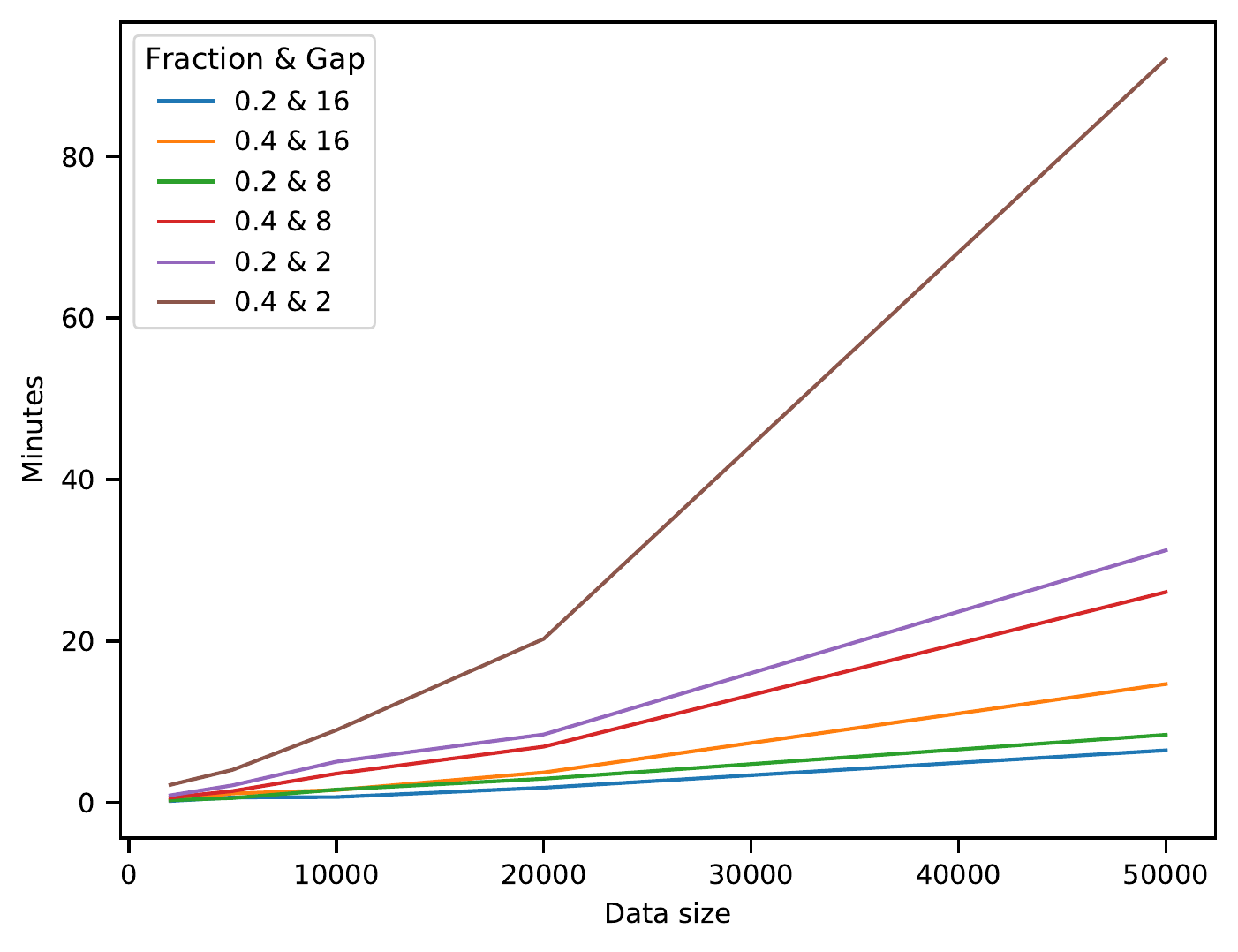} 
    \caption{\sys training speed, MNIST dataset and t-SNE training projection, different training set fractions $f$ and gap sizes $g$. Smaller gap sizes result in larger $|\mathbf{h}|$ and thus longer training times. For smaller datasets (less than 20k), \sys can be fully trained within 10 to 20 minutes}.
    \label{fig:training_speed}
\end{figure}

Figure~\ref{fig:training_speed} shows training times for \sys for different gap and data fraction values.
The figure illustrates the trade offs in training speed for different data fractions and gap sizes.
As detailed in Section~\ref{sec:method}, the neural network will be trained on the number of sampled instances times the number of hyperparameters.
For our experiments we chose what we expect an average case would be, examining hyperparameter values from $2$ to $50$.
It should be noted that Figure~\ref{fig:training_speed} shows overall training time for \sys as a function of dataset size; the number of training instances is determined not just by this value, but the number of hyperparameter values and fraction of this number used.
The \sys model is trained with a batch size of $32$ in Figure~\ref{fig:training_speed}.
It is worth noting that the batch size will often have a considerable effect on not only the wall time it will take a network to converge, but also the final performance of the model.
A more thorough discussion of this topic can be found in a number of papers~\cite{keskar:2017:on_large_batchsize_deeplearning, masters:2018:revisiting_small_batchsize, he:2019:control_batchsize_generalize}.

We evaluate the inferencing speed of \sys. Table~\ref{tab:inference_speed} shows the \sys's run-time performance in inferencing using a batch size of $20k$ instances.
As indicated in the table, \sys remains highly interactive with $500k$ data points, performing inferencing within less than 500ms\cite{liu2014effects}. 
Overall, the speed of \sys scales linearly with the number of data points projected.
\begin{table}[H]
\centering
\small
\begin{tabular}{lrr}
\toprule
Number of Instances &   Seconds \\
\midrule
56,000 &  0.192241 \\
168,000 &  0.264158 \\
336,000 &  0.370889 \\
504,000 &  0.479788 \\
616,000 &  0.551125 \\
728,000 &  0.624992 \\
840,000 &  0.695622 \\
952,000 &  0.766805 \\
1,064,000 &  0.839778 \\
\bottomrule
\end{tabular}
\caption{\sys inference speed on the MNIST dataset oversampled up to roughly 1M samples. Note that at 500k data points, \sys can still achieve interactivity (sub 500ms response rate) as suggested by Liu and Heer\,\cite{liu2014effects}.}
\label{tab:inference_speed}
\end{table}

The training speed tests were executed in a 2-way, 16-core Intel Xeon Silver 4216@2.1 GHz with 256 GB of RAM, and an NVidia GeForce 1080 Ti GPU, with 11 GB of VRAM. The inferencing speed tests were executed on a 8-core Intel i7-6700k@4 GHz with 64 GB of RAM, and an NVidia GeForce 1080 GPU, with 8GB of VRAM.

\section{Discussion}
\label{sec:discussion}
Our evaluations suggest that \sys accurately approximates projection techniques such as t-SNE, UMAP, and Isomap, and scales with data size so as to provide an analyst an interactive experience when exploring the effects of projection hyperparameters. 

Beyond approximating projection methods, \sys can have additional applications in visual analytics given its use of a neural network. We next describe these potential applications, their limitations, and the possibility of future research in developing neural networks as approximations of machine learning functions.

\subsection{Case Study: Using \sys to Approximate iPCA}
\label{sec:discussion:ipca}
\begin{figure}[htbp!]
    \centering
    \includegraphics[width=\linewidth]{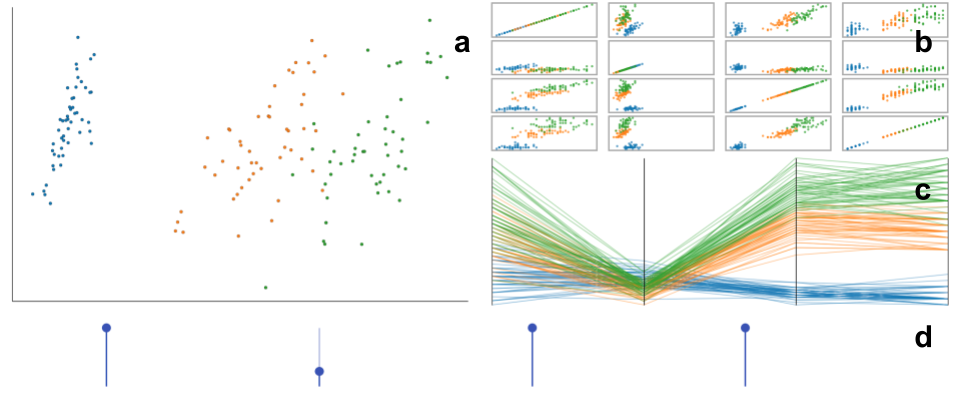} 
    \caption{A \sys implementation of iPCA. Each slider (d) corresponds to  a dimension of the input dataset (Iris)\cite{anderson:1935:iris}. Adjusting a slider scales a feature from 100\% to 0\%. This is reflected in the parallel coordinates plot (c) and scatterplot matrix (b). \sys re-projects (a) the dataset in real time as the sliders change given its speed (Sec.~\ref{sec:validation:speed}).}
    \vspace{-0.15cm}
    \label{fig:ipca_demo}
\end{figure}
iPCA\,\cite{jeong:2009:ipca} allows users to interactively adjust the weighted contributions of data dimensions and observe the effects in a PCA projection. 
Similar to \sys, iPCA allows a user to interactively manipulate projections using sliders in real time. However, unlike the use shown so far of \sys for exploring a \emph{projection's} hyperparameters, iPCA allows users to modify the hyperparameters of a \emph{data transformation} function to alter the weights (or scaling factors) of the input data.

We next show how \sys can approximate iPCA. The training process is the same as described in Sec.~\ref{sec:method}. In Eqn.~\ref{eq:inference}, $\mathbf{h}$ now is a $n$-dimensional vector that represents the scaling factors in iPCA where $n$ is the number of data dimensions, each controlled by a slider in iPCA.

Figure~\ref{fig:ipca_demo} shows \sys's approximation of iPCA.
We see that \sys is successful in emulating iPCA: interactions with the sliders in \sys produces the same projections as iPCA while retaining the interactive speed. 
This is important in terms of genericity of \sys. 
Indeed, iPCA is designed specifically around PCA, using  Online SVD\,\cite{brand:2003:online_svd} to reduce the computation cost of singular value decomposition from $O(N^2)$ to $O(n^2)$ where $N$ and $n$ are the data size and dimensionality, respectively. Replacing iPCA with another projection while maintaining interactivity is thus hard. In contrast, doing so with \sys is simple -- replacing PCA in iPCA requires just changing the function $P$ in Eqn.~\ref{eq:sampled_objective} from PCA to another projection technique. This gives credence to the use of \sys to generalize high-dimensional data exploration techniques (like iPCA) beyond their associated projection methods. However, a thorough exploration of the potential use of \sys in generalizing visual analytics methods based on projections (to use other projection methods) is beyond the scope of this paper.
%

%

%
%
\subsection{Other Considerations}
\label{sec:discussion_considerations}
Using deep learning to approximate projections raises some important considerations.
Since a neural network does not consider the semantics of a hyperparameter, it can generate projections using invalid configurations, see \emph{e.g.} the use of non-integral $k$ values for UMAP (Sec.~\ref{sec:applications:umap}). In this case, \sys infers the projection by interpolating the hyperparameter setting between integral values that it saw during training.
While, formally speaking, this goes beyond the assumptions of UMAP (\emph{i.e.}, $k$ is an integer), this interpolation has yielded reasonable results for all $k$ values, with no negative consequences noticed (see Fig.~\ref{fig:umap_noninteger_k}). 
%

%
The hyperparameter sampling strategy is also an important factor for \sys.
While t-SNE's perplexity and UMAP's $k$-nearest neighbors are both hyperparameters, they respond differently to sampling.
Figure~\ref{fig:digits_umap_trust_and_continuity} show how quality changes when hyperparameters move farther from training values. For t-SNE, neither continuity nor trustworthiness change much. For UMAP, both metrics decrease as we move $k$ away from training values. Still, while these metrics show a dip in the middle of large gaps (farthest from $k$ training values), we still get high quality (continuity $>$ 0.96, trustworthiness $>$ 0.80). This tells us that the gap size used to train \sys has greater impact for UMAP than for t-SNE; yet, even a gap size of 16 only marginally changes the projection quality. Summarizing, choosing the sample step (gap size) of hyperparameters needs to be studied individually for different projection techniques.

\subsection{Limitations and Future Work}
In this section we present potential limitations of \sys and discuss avenues to further develop this work.

\vspace{6pt}\noindent\textbf{Training Time:} Section~\ref{sec:validation}  discusses the training time for \sys for a single hyperparameter. This time can grow exponentially as the number of hyperparameters $|\mathbf{h}|$ increases. If the complexity of training a traditional network is $O(\text{K})$, the complexity of training \sys with one hyperparameter is $O(\text{K}\times|H'|)$ with $|H'|$ as described in Eqn.~\ref{eq:sampled_objective}. For $|\mathbf{h}|$ hyperparameters, this  becomes $O(\text{K}\times|H'|^{|\mathbf{h}|})$.
Further work is needed to develop heuristics to reduce the training cost of \sys for projection algorithms with many hyperparameters.

\vspace{6pt}\noindent\textbf{Training Data:} Like any deep learning technique, \sys can be influenced by the quality of training data. Our results do not show obvious problems in this sense. For example,  Fig.~\ref{fig:tsne_train_test_both} uses only 20\% data for training, but we have observed that the sampling strategy can introduce biases. Unintentional bias can be introduced by any sampling strategy (to create training sets) that does not capture the `essence' of the distribution that the data comes from. We note that the problem of training with sampling data is not unique to our method. We leave as future work strategies to mitigate its effects on approximating projections.

\vspace{6pt}\noindent\textbf{Smoothness Perception:} The technique described in Sec.~\ref{sec:visualization_considerations} maintains projection stability and visual coherence between frames as users interactively change a hyperparameter value. When combined with the high interactivity of \sys, this may induce a false sense of continuity concerning the \emph{properties of the underlying learned projection method}. For example, consider approximating UMAP with \sys (Sec.~\ref{sec:applications:umap}). When the hyperparameter $k$ changes from $k$ to $k+1$, there is no guarantee that the two learned manifolds share correspondences, so the perception of smoothness created by \sys may be misleading. Conversely, for iPCA (Sec.~\ref{sec:discussion:ipca}), weight changes should result in a smooth animation as PCA rotates the underlying coordinate system to maximize the projected data variance. Summarizing, the built-in perception of smoothness of \sys should not always be interpreted as reflecting mathematical smoothness properties of the learned projections. 


\section{Conclusion}
\label{sec:conclusion}
We proposed \sys, a deep learning approach to approximating projections that enables real-time interactive hyperparameter exploration.
We showed that \sys can learn to infer projections of several techniques, for a wide range of parameter values, and different real-world datasets, using only a small fraction of the data and hyperparameter values. \sys performs real-time for reasonably large datasets, can be generalized to any projection technique (and hyperparameters), and produces stable animations of the resulting projections upon parameter variations. We demonstrated \sys to the interactive exploration of the key hyperparameters of t-SNE, UMAP, and Isomap.

Future work can target accelerating of \sys for projections with many parameters; a deeper analysis of how finely sampled the input data and parameter space is needed for high-quality inference; and most importantly, deploying \sys in production visual analytics systems where projections need to be explored interactively upon hyperparameter change.
%


\bibliographystyle{abbrv-doi}
\clearpage
\bibliography{bibliography}
\end{document}